\documentclass{article}

\newif\ifpreprint
\preprinttrue

\newif\ificlrstyle
\IfFileExists{iclr2027_conference.sty}{%
  \iclrstyletrue
  \usepackage{iclr2027_conference}
}{%
  \iclrstylefalse
  \usepackage[margin=1in]{geometry}
}

\usepackage[T1]{fontenc}
\usepackage{times}
\usepackage{microtype}
\usepackage{amsmath,amssymb,amsfonts}
\usepackage{mathtools}
\usepackage{booktabs}
\usepackage{multirow}
\usepackage{array}
\usepackage{graphicx}
\usepackage{adjustbox}
\usepackage[table]{xcolor}
\usepackage{enumitem}
\usepackage{natbib}
\usepackage{url}
\usepackage{placeins}
\usepackage{float}
\usepackage[hidelinks]{hyperref}

\definecolor{evepink}{HTML}{ED028C}

\floatstyle{ruled}
\newfloat{algorithm}{tbp}{loa}
\floatname{algorithm}{Algorithm}
\floatstyle{plain}


\newcommand{\method}{\textsc{EVEWorld}}
\newcommand{\Lcic}{\mathcal{L}_{\mathrm{TIA}}}
\newcommand{\Lpipe}{\mathcal{L}_{\mathrm{IGR}}}
\newcommand{\safeincludegraphics}[2][]{%
  \includegraphics[#1]{#2}%
}

\title{EVEWorld: Physical Evolution Supervision for Embodied World Models}
\date{}

\hypersetup{%
  pdftitle={EVEWorld: Physical Evolution Supervision for Embodied World Models},
  pdfauthor={Kaiqi Wang, Songxin Zhang, Zejian Xie, Xiao Xiong, Zhuoyang Song, Ziwei Wu, Jun Yu Lu, Yitan Teng, Ziying Song, Jiaxing Zhang},
  pdfsubject={World Models; Embodied AI; Video Generation},
}

\author{\normalfont%
Kaiqi Wang\textsuperscript{1},\;
Songxin Zhang\textsuperscript{1},\;
Zejian Xie\textsuperscript{1},\;
Xiao Xiong\textsuperscript{1},\;
Zhuoyang Song\textsuperscript{1},\\
Ziwei Wu\textsuperscript{1},\;
Jun Yu Lu\textsuperscript{1},\;
Yitan Teng\textsuperscript{1},\;
Ziying Song\textsuperscript{2,\dag},\;
Jiaxing Zhang\textsuperscript{1,\dag}\\[1.0ex]
{\small\textsuperscript{1}China Merchants Group, \textsuperscript{2}Nanyang Technological University}\\[0.6ex]
{\small\textsuperscript{\dag}Corresponding authors: Ziying Song (ziying.song@ntu.edu.sg), Jiaxing Zhang (zhangjiaxing@cmhk.com)\hspace*{-\tabcolsep}}
}

\ificlrstyle
  \ifpreprint\iclrfinalcopy\fi   % real author block, no submission ruler
\fi

\begin{document}
\maketitle
\ificlrstyle\ifpreprint\lhead{Preprint}\fi\fi

% Project page, centred under the author block (arXiv / preprint build only).
\ifpreprint
\vspace{-1.2em}
\begin{center}
\href{https://behappya.github.io/EVEWorld/}{\textcolor{evepink}{\texttt{https://behappya.github.io/EVEWorld/}}}
\end{center}
\vspace{-0.6em}
\fi

\begin{figure}[H]
  \centering
  \includegraphics[width=\textwidth]{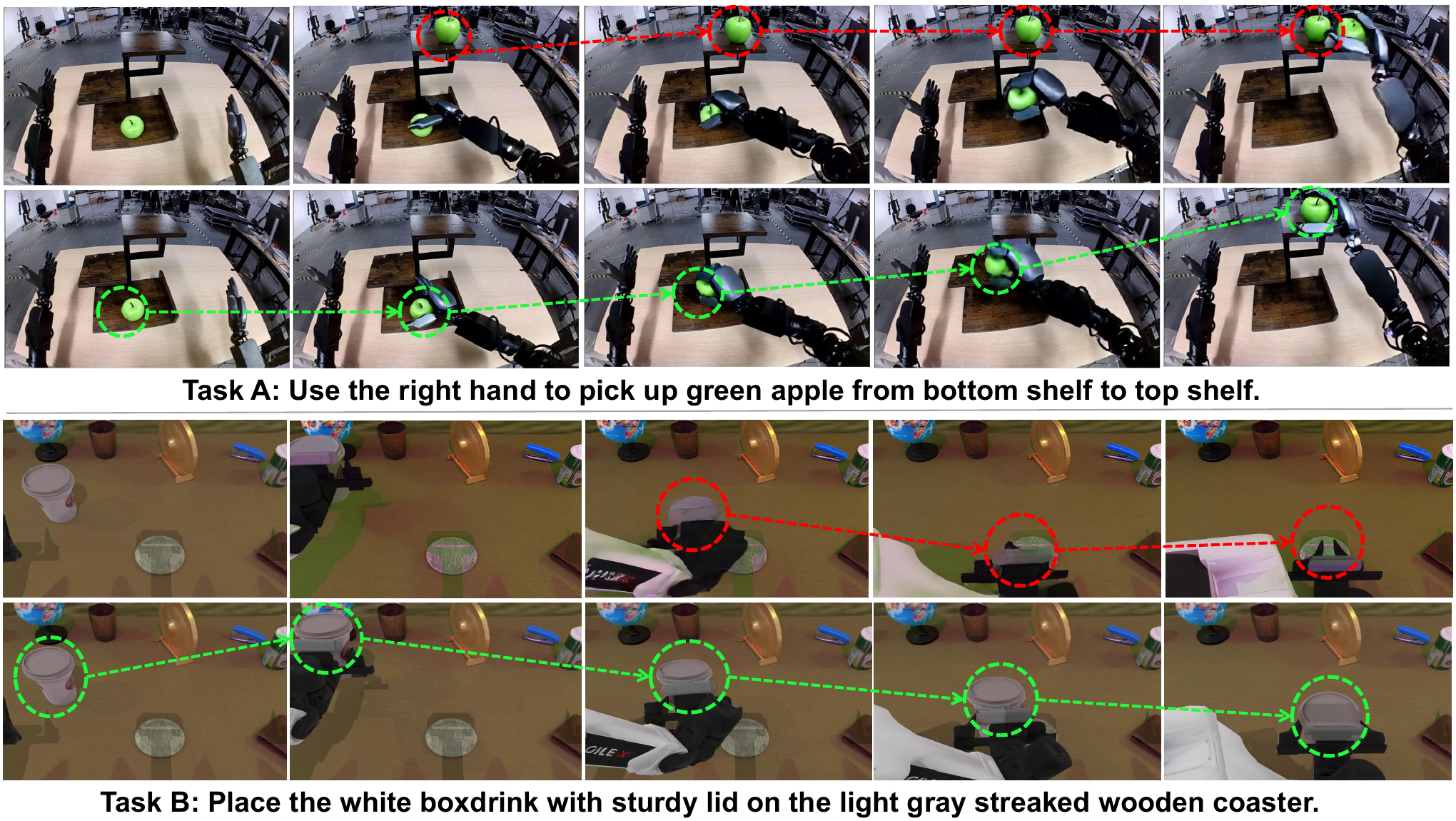}
\caption{
Embodied world models face the challenge of \textbf{Model Laziness}, where instance consistency is violated by spurious duplication or disappearance during interaction.
\textbf{\textcolor{red}{Red}} denotes Standard SFT, which exhibits such failures, while
\textbf{\textcolor{green!50!black}{Green}} denotes \method{}, which better maintains physical consistency throughout the interaction.
Dashed arrows trace target evolution over time.
}

  \label{fig:teaser}
\end{figure}

\begin{abstract}
Embodied world models enable scalable simulation of embodied interactions for robot learning.
However, existing models are prone to \textbf{Model Laziness}, as they focus on visual fidelity at the expense of physical reasoning and lack process-level supervision over the temporal dynamics of manipulated objects.
In this work, we propose \method{}, a physical evolution-supervision framework for physically consistent target evolution.
\method{} consists of two components: Instance-Guided Restoration (IGR) and Temporal Instance Alignment (TIA).
First, IGR promotes instance consistency through restoration supervision.
Second, TIA promotes cross-frame consistency by aligning target instances across adjacent frames.
We further introduce the Model Laziness Rate (MLR), a metric that measures persistent violations of instance consistency in generated trajectories.
Extensive experiments on DreamGenBench, EWMBench, and PBench demonstrate the effectiveness of \method{}, notably achieving an 87.5\% reduction in MLR compared with GigaWorld-0.
On the WorldArena 2.0 Track 1 leaderboard, our model ranks 6th in JEPA Similarity and 17th overall, which further validates the performance of our evolution supervision strategy.
\end{abstract}

\section{Introduction}
% [中文] 引言
\label{sec:intro}
Embodied world models and related policy-learning methods are emerging as scalable predictive engines for embodied intelligence, modeling diverse interaction dynamics to support robot learning, planning, and simulation \citep{yang2024unisim,bruce2024genie,zhou2024robodreamer,jang2025dreamgen,li2025embodiedwmsurvey,du2023unipi,schmidt2024lapo,ye2025lapa,brohan2023rt2,kim2024openvla,black2025pi0,black2025pi05}.
Recent work \citep{huang2025enerverse,huang2025vid2world,guo2025ctrlworld,gigaworld2026one,xiao2025worldmem,pang2025reviwo,dou2026slotmemory,duan2026liveworld,zheng2026memworld,bardhan2026persistent,li2025worldeval,li2026dworldeval,quevedo2025worldgym} has further advanced embodied world modeling and its evaluation through high-fidelity generation, multiview and long-context modeling, action conditioning, persistent memory, and task-oriented evaluation.

Recent embodied world models further strengthen structural faithfulness through explicit 4D RGB--depth--normal prediction, geometry-aware action conditioning, and disentangled dynamics and visual synthesis, with related physical-evaluation benchmarks \citep{zhen2025tesseract,dreamxteam2026dreamxphi,shan2026dvgwm,liu2026geometry4d,li2025worldmodelbench,bansal2025videophy2,meng2025phygenbench,qin2025worldsimbench,shang2026worldarena,motamed2025physicsiq,garrido2025intuitive}.
Nevertheless, stronger visual and structural modeling does not necessarily ensure physically consistent target evolution: a rollout may exhibit target-instance consistency failures, such as duplication or disappearance, and cross-frame consistency failures, such as deformation and identity drift \citep{yuan2026newtongen,foo2026psivg}.
Figure~\ref{fig:teaser} shows a representative instance consistency failure after standard supervised fine-tuning (SFT).
We refer to such instance consistency failures as \textbf{Model Laziness}: frame-level reconstruction can admit a shortcut that satisfies appearance or endpoint cues without maintaining instance consistency throughout the interaction.

\begin{figure*}[!t]
  \centering
  \includegraphics[width=\textwidth]{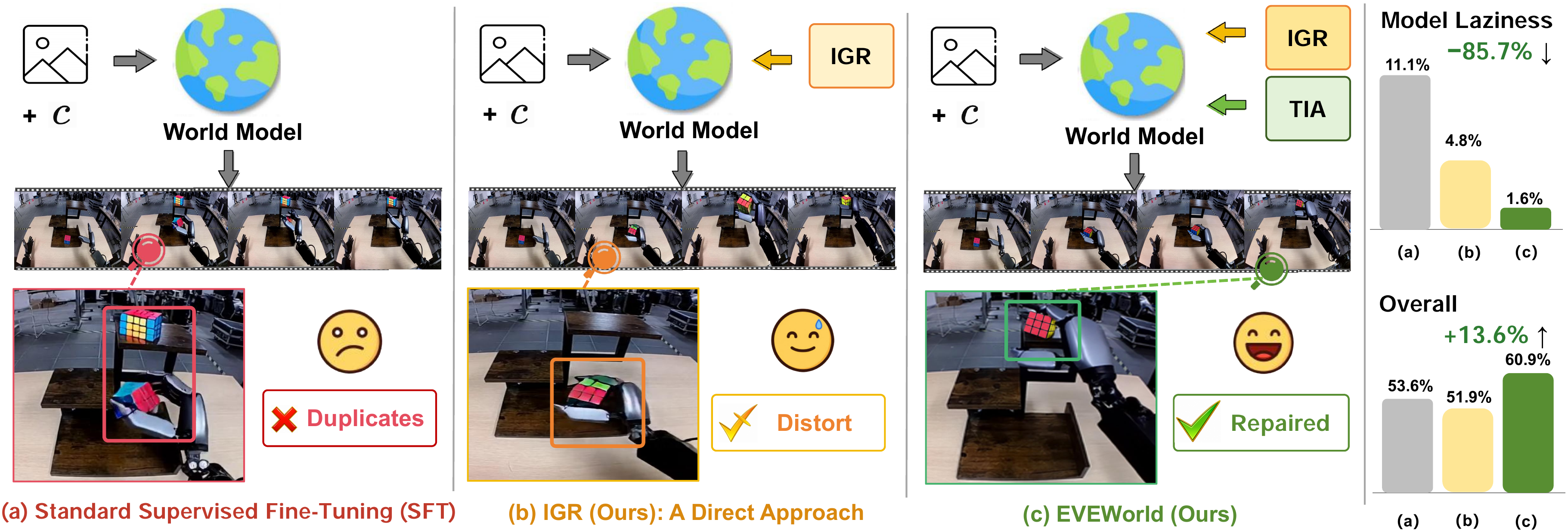}
\caption{
\textbf{Comparison of Standard SFT, IGR, and \method{}.}
\textbf{(a)} Existing video world models \citep{gigaworld2025zero,yang2025cogvideox,wan2025wan,nvidia2025cosmos}, including our Standard SFT baseline, may exhibit instance-consistency failures, such as target duplication at the goal.
\textbf{(b)} Our IGR variant improves instance consistency, but cross-frame consistency failures such as target distortion may remain.
\textbf{(c)} Our full \method{} model combines IGR and TIA to maintain physically consistent target evolution throughout the interaction.
Relative to Standard SFT, \method{} reduces MLR by 85.7\% and improves overall Gemini-IF by 13.6\%.
Additionally, the inputs consist of an initial image and an instruction $c$.
}
\label{fig:overview}
\end{figure*}

In this work, we propose \method{}, an evolution-supervision framework that promotes physically consistent target evolution and mitigates Model Laziness in embodied world models, as shown in Figure~\ref{fig:overview}.
\method{} comprises two complementary components: \textbf{Instance-Guided Restoration (IGR)}, which promotes instance consistency by restoring count-perturbed videos, and \textbf{Temporal Instance Alignment (TIA)}, which promotes cross-frame consistency by aligning target instances across adjacent frames.
To quantify Model Laziness, we introduce the \textbf{Model Laziness Rate (MLR)}, a metric that measures persistent violations of target-instance consistency over a rollout.
Extensive evaluations on DreamGenBench, WorldArena 2.0, EWMBench, and PBench demonstrate the effectiveness of \method{} in promoting physically consistent target evolution.
Notably, on the WorldArena 2.0 Track 1 leaderboard, \method{} achieves an EWMScore-P of 70.17, ranking 17th overall and 6th in JEPA Similarity.
These results further support the effectiveness of evolution supervision under large-scale world-model evaluation.
\section{Related Work}
\label{sec:related}

\paragraph{Video Generation Models for Robotics.}
Large-scale video models provide strong priors for embodied prediction and scalable robot-data generation \citep{blattmann2023svd,openai2024sora,wan2025wan,kong2024hunyuanvideo}, enabling world models for robot learning, planning, and simulation \citep{yang2024unisim,zhou2024robodreamer,jang2025dreamgen,lu2025gwm,gao2025flip,gigaworld2025zero}. Recent work \citep{zhen2025tesseract,dreamxteam2026dreamxphi,shan2026dvgwm,fu2026robomaster} further strengthens structural modeling through richer representations and conditioning: TesserAct jointly predicts RGB, depth, and normals, DreamX-Phi incorporates geometric action conditioning and depth supervision, and DVG-WM disentangles dynamics from visual synthesis, and RoboMaster improves multi-object interaction modeling. In contrast, our \method{} retains the host model's native representation and introduces evolution supervision during post-training to promote physically consistent target evolution.

\paragraph{Physical Reasoning in Embodied World Models.}
Prior works \citep{yang2024fresco,traub2023objectpermanence,jeong2025track4gen,chen2025videoamodal,wu2026consistentinstance,xiao2025worldmem,dou2026slotmemory} improve object-centric and temporally consistent modeling through correspondence, tracking, object permanence, and persistent memory, while other studies \citep{deng2026rethinking,sun2024t2vcompbench,khose2026wfmeval} examine object dynamics, quantity behavior, and broader aspects of physical reasoning from different perspectives. However, despite these advances and evaluations \citep{yang2024fresco,traub2023objectpermanence,jeong2025track4gen,xiao2025worldmem,dou2026slotmemory,deng2026rethinking,sun2024t2vcompbench,khose2026wfmeval}, physically consistent target evolution remains challenging, with manipulated objects exhibiting failures in both instance consistency and cross-frame consistency. In contrast, our \method{} introduces evolution supervision to explicitly promote physically consistent target evolution.

\section{Why Embodied World Models Become Lazy} 
% [中文] 具身世界模型为何会变懒 \label{sec:why}
\subsection{Problem Statement}
% [中文] 问题陈述
\label{sec:problem}

Given an initial observation $I_0$ and an instruction $c$, an embodied video world model predicts future frames $x_{1:T}$. We focus on manipulation scenarios that require instance consistency, where the target-instance count is expected to remain unchanged throughout the interaction.

Let $N_t(I_0,x_t,c)$ denote the number of instances of the target specified by $c$, with $N_0(I_0,c)$ denoting its initial count. We define Model Laziness as
\begin{equation}
\mathrm{ML}(I_0,x_{1:T},c)
=
\mathbb{I}\!\left[
\exists t\in\{1,\ldots,T\}:
N_t(I_0,x_t,c)\neq N_0(I_0,c)
\right].
\label{eq:ml}
\end{equation}
This captures instance consistency failures caused by unsupported target duplication or disappearance. Interactions that intentionally change the target-instance count, such as cutting or assembly, are outside the scope of the current formulation and MLR protocol. Robot-induced occlusion is handled separately through an occlusion-aware check during MLR evaluation.

\subsection{Causes of Model Laziness}
\label{sec:causes}

Standard training emphasizes visual reconstruction while providing limited process-level supervision for physically consistent target evolution \citep{denton2018svg,bengio2015scheduled,jeong2025track4gen}. This leaves two key gaps: instance consistency and cross-frame consistency.

\noindent\textbf{Missing Instance-Consistency Supervision.}
Experimental results show that Standard SFT retains 92\%--99\% of the injected target area across different perturbation settings (Appendix Figure~\ref{fig:mechanism}, Appendix~\ref{app:mechanism_probes}). Because the perturbed region is small relative to the full frame, its reconstruction error can be diluted by the surrounding scene, providing insufficient supervision for recovering the correct target-instance count. IGR (Section~\ref{sec:igr}) addresses this gap by providing a direct restoration target and assigning greater spatial weight to the perturbed region.

\noindent\textbf{Missing Cross-frame Consistency Supervision.}
IGR improves instance consistency, but a correct target-instance count alone does not ensure cross-frame consistency; the target may still become distorted, fade, or shift abruptly across adjacent frames, as shown in Figure~\ref{fig:overview}(b). This stems from the lack of explicit supervision linking the same target across frames, allowing locally plausible predictions to evolve inconsistently over time. TIA (Section~\ref{sec:tia}) addresses this gap by aligning target instances across adjacent frames.

\section{Method}
% [中文] 方法
\label{sec:method}

\paragraph{Framework Overview.}
Given a clean demonstration video $x$ and its instruction $c$, \method{} uses IGR (Section~\ref{sec:igr}) to construct a count-edited video $\tilde{x}$ and trains the model to restore $x$, assigning higher weights to restoration-related regions to promote instance consistency. TIA (Section~\ref{sec:tia}) operates within the video diffusion transformer to align target features across adjacent frames and promote cross-frame consistency. Together, IGR and TIA enable physically consistent target evolution in the final video output. Notably, the demonstration video is used for training supervision; at inference, \method{} takes only the initial image and instruction as input.

\begin{figure*}[!t]
  \centering
  \includegraphics[width=\textwidth]{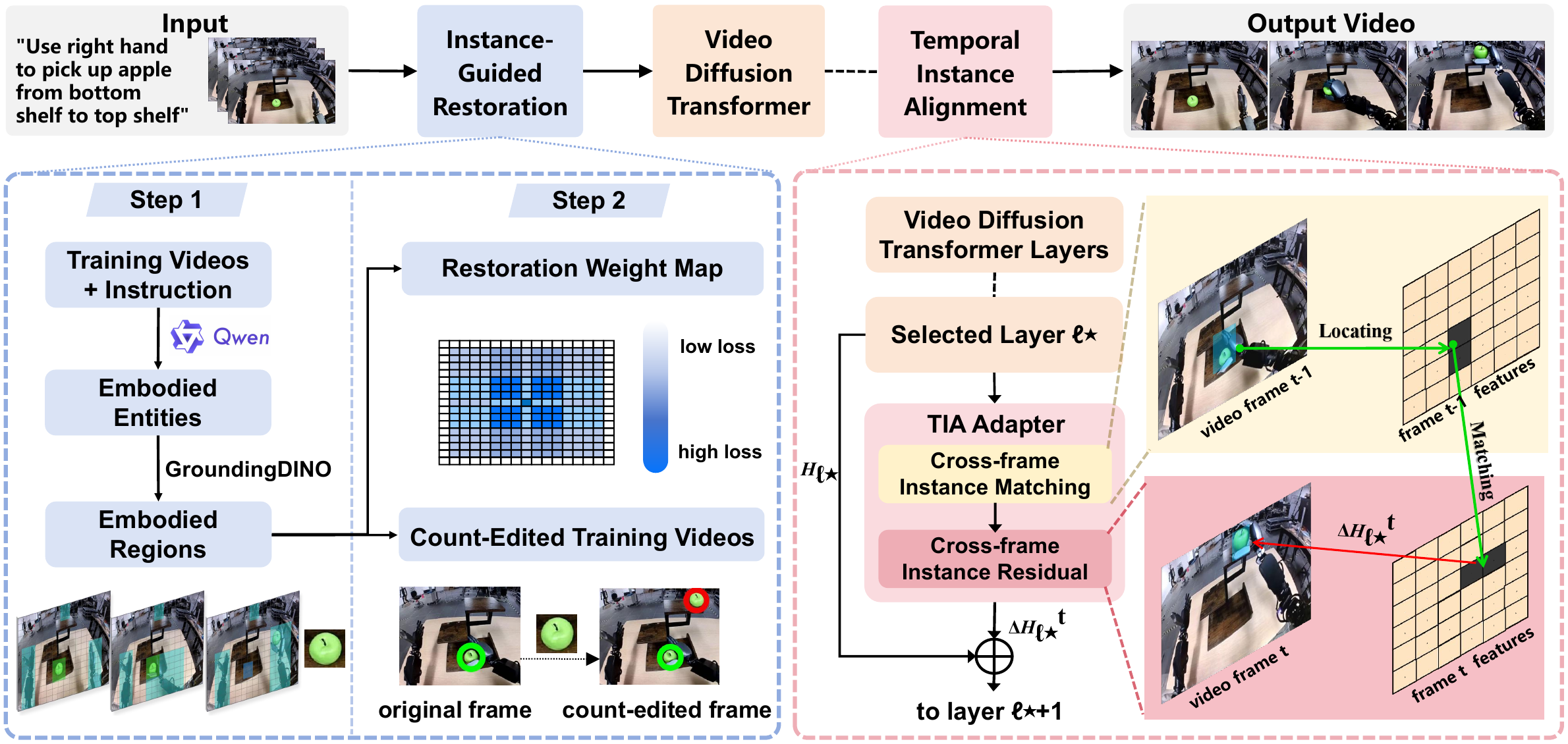}
\caption{
\textbf{Overall architecture of \method{}.}
Given a clean demonstration $x$ and instruction $c$ during training, IGR constructs a count-edited video $\tilde{x}$, which is processed by the video diffusion transformer to restore $x$. Within the transformer, TIA operates at the selected layer $\ell^\star$ to perform cross-frame instance matching and residual refinement, after which the refined features pass through the remaining layers to produce the final output.
}

  \label{fig:method}
\end{figure*}

\subsection{Instance-Guided Restoration}
\label{sec:igr}

To address failures in physically consistent target evolution, we introduce Instance-Guided Restoration (IGR) to promote instance consistency. IGR inserts an extra target instance into clean demonstrations to form count-edited samples, then trains the model to restore the corresponding clean latent representation encoded by a pretrained VAE. A language model and GroundingDINO \citep{liu2024groundingdino} localize the target trajectory to sample a target patch $p$ and valid paste region $\Omega$:

\begin{equation}
\tilde{x}=\mathrm{Paste}(x,p,\Omega),\qquad
z=\mathrm{VAE}(x),\qquad
\tilde{z}=\mathrm{VAE}(\tilde{x}).
\label{eq:paste}
\end{equation}

Training the model to map the count-edited latent $\tilde z$ back to the clean latent $z$ encourages recovery of the correct target-instance count. To prevent this local instance-consistency signal from being diluted by global reconstruction, we assign restoration-related regions three times the reconstruction weight of ordinary regions and normalize the resulting spatial weight map to unit mean. We denote the normalized map by $\hat W$; its exact construction is given in Appendix~\ref{app:igr_details}. Using the EDM-style noise weighting $\lambda(\sigma)$ \citep{karras2022edm}, the restoration objective is

\begin{equation}
\Lpipe=
\mathbb{E}\!\left[
\lambda(\sigma)\frac{1}{CTHW}
\left|
\hat W^{1/2}\odot
\left(D_\theta(\tilde z+\sigma\epsilon,c,\sigma)-z\right)
\right|_2^2
\right].
\label{eq:pipeline}
\end{equation}

Minimizing $\Lpipe$ provides explicit instance-consistency supervision by recovering the correct target-instance state from count-perturbed inputs; details are provided in Appendix~\ref{app:igr_details}.

\subsection{Temporal Instance Alignment}
\label{sec:tia}
IGR improves instance consistency, but a correct target-instance count alone does not guarantee cross-frame consistency. Failures such as identity drift, distortion, fading, or abrupt motion may still occur across adjacent frames. We therefore introduce Temporal Instance Alignment (TIA) to explicitly promote cross-frame consistency by aligning target instances across adjacent frames.

We select the attachment layer using an offline layer-wise matching probe. For each candidate layer $\ell$, the frame-$(t-1)$ target feature retrieves its best match over all token cells of frame $t$, and we select the layer with the lowest endpoint error (EPE):
\begin{equation}
\hat u_t^\ell=
\arg\max_{v}
f_{t-1}^{\ell}(u_{t-1})^\top f_t^\ell(v),
\qquad
\ell^\star=
\arg\min_{\ell\in\mathcal{B}}
\frac{1}{|\mathcal{V}|}
\sum_{t\in\mathcal{V}}
\left\|\hat u_t^\ell-u_t\right\|_2 .
\label{eq:epe}
\end{equation}
where $\mathcal{V}$ contains adjacent-frame pairs with the target visible in both frames and detector confidence above a threshold. At layer $\ell^\star$, TIA has two stages (Figure~\ref{fig:method}, right). \emph{Local instance locating} matches target features within a local neighborhood, while \emph{cross-frame instance transport} converts pre-transport cosine similarities into softmax transport weights, splats projected previous-frame features into frame $t$, normalizes by transport mass, and writes them back through a zero-initialized bounded residual.
\begin{equation}
H_t'
=
H_t^{\ell^\star}
+
\gamma\,\tanh\!\left(
W_{\mathrm{out}}
\big(
\bar H_t-W_{\mathrm{in}}H_t^{\ell^\star}
\big)
\right).
\label{eq:tia_transport_main}
\end{equation}
Here, $\bar H_t$ denotes the transported feature field. GroundingDINO and target locations $u_t$ provide offline alignment supervision during training only; at inference, dense matching and transport operate directly on the model's hidden features and require neither target localization nor an external tracker. The transported features are further supervised to favor the true next-frame target location over nearby candidates:
\begin{equation}
\Lcic=
-\frac{1}{|\mathcal{V}|}
\sum_{t\in\mathcal{V}}
\log
\frac{\exp(s_{t,u_t}/\tau)}
{\sum_{v\in\mathcal{C}_t}\exp(s_{t,v}/\tau)},
\qquad
s_{t,v}=
\bar f_{t-1}^{\ell^\star}(u_{t-1})^\top
\bar f_t^{\ell^\star}(v),
\label{eq:cic}
\end{equation}
where $\bar f^{\ell^\star}$ is the normalized feature after transport. The locating, transport, and alignment components jointly constrain target identity across frames. Further details are provided in Appendix~\ref{app:tia_details}.

\section{Experiments}
% [中文] 实验
\label{sec:experiments}

\subsection{Experimental Setup and Metrics}
\label{sec:dataset}

\paragraph{Benchmarks and Data Splits.}
We use the official DreamGen GR1 fine-tuning split, which contains 92 robot-manipulation videos paired with instructions, and primarily evaluate on DreamGenBench, comprising 126 manipulation tasks across unseen environments, objects, and behaviors \citep{jang2025dreamgen}. We further evaluate cross-domain generalization on WorldArena 1.0, cross-trial robustness on PBench, cross-distribution generalization on AgiBot--EWMBench, and cross-backbone generalization using FlowWAM on WorldArena 2.0 and held-out RoboTwin actions \citep{shang2026worldarena,shang2026worldarena2,bu2025agibot,hu2025ewmbench,mu2025robotwin,nvidia2025pbench,huang2024vbench,liu2024evalcrafter,unterthiner2018fvd,chen2026flowwam}.

\paragraph{Evaluation Metrics.}
\label{sec:mlrmetric}
We evaluate instruction following with Qwen-IF and Gemini-IF \citep{bai2025qwen3vl,geminiteam2024gemini}; WorldArena 1.0/2.0, EWMBench, PBench, and RoboTwin assess complementary aspects of perceptual quality, motion, semantics, consistency, and task fidelity.

Related evaluations \citep{deng2026rethinking,khose2026wfmeval,sun2024t2vcompbench}, including RBench, WFM-Eval, and video numeracy benchmarks, provide indirect signals related to object quantity and persistence. In contrast, we propose the \textbf{Model Laziness Rate (MLR)}, a metric that explicitly measures target-instance consistency at the rollout level using the observed initial count $N_0$ as its reference, without requiring a prompt-specified count or reference future. We exclude under-counts attributable to robot-induced occlusion while retaining all over-counts, and denote the resulting evaluation count by $\widetilde{N}_t(x)$. MLR implements Eq.~\ref{eq:ml} with a two-frame persistence criterion to suppress isolated detection errors. We define
\begin{equation}
\mathrm{MLR}
=
\frac{100}{|\mathcal{D}_{+}|}
\sum_{x\in\mathcal{D}_{+}}
\mathbb{I}\!\left[
\exists t\in\{1,\ldots,T-1\}:
\widetilde{N}_t(x)\neq N_0(x)
\land
\widetilde{N}_{t+1}(x)\neq N_0(x)
\right],
\label{eq:mlr}
\end{equation}
where $\mathcal{D}_{+}=\{x\in\mathcal{D}:N_0(x)>0\}$. Further details are provided in Appendix~\ref{app:mlr}.

\paragraph{Model and Training Setup.}
For the main experiments, we initialize \method{} from the 2B-parameter GigaWorld-0 Video-Pretrain-2B \citep{gigaworld2025zero} and compare against CogVideoX1.5, Wan2.2, and Cosmos-Predict2 as zero-shot baselines \citep{yang2025cogvideox,wan2025wan,nvidia2025cosmospredict2}. Standard SFT, \method{}, and all ablations share the same initialization, training data, and 250-step optimization budget. We fine-tune all transformer parameters on eight NVIDIA H20Z GPUs with an effective batch size of 64 and a learning rate of $4.32 \times 10^{-5}$. For cross-distribution and cross-backbone experiments, we apply the same evolution-supervision method to the corresponding data and backbone. Details are provided in the appendix.

\subsection{Main Results}
% [中文] 主要结果
\label{sec:mainresults}
\begin{table}[!t]
\centering
\caption{
\textbf{Overall comparison on DreamGenBench.}
MLR denotes Model Laziness Rate, and Qwen-IF/Gemini-IF measure instruction-following accuracy.
General video models are evaluated zero-shot, whereas Standard SFT and \method{} are post-trained on DreamGen.
MLR is computed on a shared set of 63 eligible items across all models.
}
\label{tab:dreamgen_overall}
\setlength{\tabcolsep}{2.8pt}
\renewcommand{\arraystretch}{1.0}

\adjustbox{max width=\textwidth}{%
\begin{tabular}{lcccccc>{\columncolor{gray!20}}c}
\toprule
Metric
& CogVideoX
& Wan-TI2V
& Wan-A14B
& Cosmos
& GigaWorld-0
& Standard SFT
& EVEWorld \\
\midrule

MLR (\%) $\downarrow$
& 28.57
& 17.46
& 11.11
& 14.29
& 12.70
& 11.11
& \textbf{1.59} \\

Qwen-IF (\%) $\uparrow$
& 38.89
& 38.89
& 64.29
& 62.70
& 79.37
& 73.81
& \textbf{80.16} \\

Gemini-IF (\%) $\uparrow$
& 5.56
& 10.32
& 15.87
& 24.60
& 60.19
& 53.57
& \textbf{60.85} \\

\bottomrule
\end{tabular}%
}
\end{table}

\paragraph{Overall Results.}
As shown in Table~\ref{tab:dreamgen_overall}, \method{} substantially mitigates Model Laziness while improving instruction following. Compared with Standard SFT, \method{} reduces MLR from 11.11\% to 1.59\% (85.7\% relative reduction), while Qwen-IF increases from 73.81\% to 80.16\% and Gemini-IF from 53.57\% to 60.85\%. Appendix~\ref{app:mlr} (Appendix Figure~\ref{fig:persistence}) further shows that the remaining violations under \method{} are transient, indicating that evolution supervision improves physical consistency without sacrificing instruction following.

\begin{table}[!t]
\centering
\caption{
\textbf{Fine-grained comparison on DreamGenBench.}
GR1-Env, GR1-Object, and GR1-Behavior denote generalization to unseen environments, objects, and behaviors, respectively.
Qwen-IF and Gemini-IF measure instruction-following accuracy.
General video models are evaluated zero-shot, whereas Standard SFT and \method{} share the same DreamGen post-training setting.
}
\label{tab:dreamgen_main}
\setlength{\tabcolsep}{4.5pt}
\renewcommand{\arraystretch}{1.0}

\adjustbox{max width=\textwidth}{%
\begin{tabular}{lccccccc}
\toprule
Method
& Param.
& \multicolumn{2}{c}{GR1-Env IF (\%)}
& \multicolumn{2}{c}{GR1-Object IF (\%)}
& \multicolumn{2}{c}{GR1-Behavior IF (\%)} \\
\cmidrule(lr){3-4}
\cmidrule(lr){5-6}
\cmidrule(lr){7-8}
&
& Qwen $\uparrow$
& Gemini $\uparrow$
& Qwen $\uparrow$
& Gemini $\uparrow$
& Qwen $\uparrow$
& Gemini $\uparrow$ \\
\midrule

\multicolumn{8}{l}{\emph{General video models (zero-shot)}} \\

CogVideoX1.5-5B-I2V
& 5B
& 48.28 & 15.52
& 24.00 & 1.00
& 48.94 & 4.26 \\

Wan2.2-TI2V-5B
& 5B
& 51.72 & 17.24
& 24.00 & 5.00
& 46.81 & 11.70 \\

Wan2.2-I2V-A14B
& A14B
& 58.62 & 20.69
& 66.00 & 13.00
& 65.96 & 15.96 \\

Cosmos-Predict2-2B
& 2B
& 68.97 & 31.03
& 60.00 & 28.00
& 61.70 & 17.02 \\

\midrule
\multicolumn{8}{l}{\emph{GigaWorld-based models}} \\

GigaWorld-0
& 2B
& 79.31 & 54.60
& \textbf{82.00} & \textbf{54.33}
& \textbf{76.60} & \textbf{69.86} \\

Standard SFT
& 2B
& 79.31 & 51.72
& 76.00 & 42.00
& 68.09 & 67.02 \\

\rowcolor{gray!20}
EVEWorld
& 2B
& \textbf{82.76} & \textbf{72.41}
& \textbf{82.00} & 50.33
& \textbf{76.60} & 64.89 \\

\bottomrule
\end{tabular}%
}
\end{table}

\paragraph{Fine-grained Analysis.}
Table~\ref{tab:dreamgen_main} reports instruction-following performance across the Environment, Object, and Behavior splits. Compared with Standard SFT, \method{} improves Qwen-IF on all three splits, with the largest gain on Behavior (68.09\% to 76.60\%), while Gemini-IF improves notably on Environment (51.72\% to 72.41\%) and Object (42.00\% to 50.33\%). Overall, these results show that evolution supervision yields broad gains across different generalization dimensions.

\subsection{Generalization Evaluation}
% [中文] 跨域与跨训练数据的泛化
\label{sec:generalization}
\begin{table}[!t]
\centering
\caption{
\textbf{Zero-shot comparison on WorldArena 1.0.}
All metrics follow the released evaluation protocol. IQ, AQ, DD, Flow, MS, SC, BC, and PC denote Image Quality, Aesthetic Quality, Dynamic Degree, Flow Score, Motion Smoothness, Subject Consistency, Background Consistency, and Photometric Consistency; Overall is their equal-weight mean. MLR is computed on 157 shared eligible items, and bold marks the best GigaWorld-based result.
}
\label{tab:worldarena}
\setlength{\tabcolsep}{3.90pt}\fontsize{10}{10.90}\selectfont
\renewcommand{\arraystretch}{1.0}

\adjustbox{max width=\textwidth}{%
\begin{tabular}{lcccccccccc}
\toprule
Method
& \multicolumn{2}{c}{Appearance}
& \multicolumn{3}{c}{Motion}
& \multicolumn{3}{c}{Consistency}
& \multicolumn{2}{c}{Main Metrics} \\
\cmidrule(lr){2-3}
\cmidrule(lr){4-6}
\cmidrule(lr){7-9}
\cmidrule(lr){10-11}
& IQ $\uparrow$
& AQ $\uparrow$
& DD $\uparrow$
& Flow $\uparrow$
& MS $\uparrow$
& SC $\uparrow$
& BC $\uparrow$
& PC $\uparrow$
& Overall $\uparrow$
& MLR (\%) $\downarrow$ \\
\midrule

\multicolumn{11}{l}{\emph{General I2V models (zero-shot)}} \\

CogVideoX
& 24.51 & 34.80
& 20.71 & 7.24 & 50.78
& 54.96 & 66.35 & 10.20
& 33.69 & 2.55 \\

Wan-TI2V
& 33.48 & 37.90
& 33.00 & 11.38 & 61.86
& 68.88 & 77.17 & 18.19
& 42.73 & 17.20 \\

Wan-A14B
& 51.93 & 43.37
& 42.29 & 8.04 & 64.13
& 67.69 & 75.37 & 33.61
& 48.30 & 29.94 \\

Cosmos
& 57.88 & 38.83
& 50.46 & 16.69 & 71.61
& 64.14 & 73.71 & 21.42
& 49.34 & 8.28 \\

\midrule
\multicolumn{11}{l}{\emph{GigaWorld-based models}} \\

GigaWorld-0
& 58.53 & \textbf{38.79}
& 44.36 & 7.74 & 61.67
& 77.90 & 88.13 & \textbf{48.32}
& 53.18 & 27.39 \\

Standard SFT
& \textbf{60.71} & 38.36
& 48.49 & 7.82 & 64.04
& \textbf{79.34} & 88.67 & 44.17
& 53.95 & 27.39 \\

\rowcolor{gray!20}
EVEWorld
& 59.82 & 37.35
& \textbf{62.90} & \textbf{10.98} & \textbf{67.82}
& 78.72 & \textbf{88.91} & 47.58
& \textbf{56.76} & \textbf{13.38} \\

\bottomrule
\end{tabular}%
}
\end{table}

\paragraph{Zero-Shot Domain Generalization.}
On WorldArena 1.0 \citep{shang2026worldarena}, which includes unseen scenes and robot embodiments, \method{} improves EWMScore-local-8 from 53.95 to 56.76 (+5.21\%) over Standard SFT while reducing MLR from 27.39\% to 13.38\% ($-$51.15\%) (Table~\ref{tab:worldarena}). These results demonstrate improved cross-domain generalization in both motion quality and target-instance consistency.

\begin{table}[!t]
\centering
\caption{
\textbf{Comparison on EWMBench after AgiBot post-training.}
Standard SFT and \method{} use the same AgiBot data and optimization budget. Motion, Semantics, and Overall are the official EWMBench aggregate scores, and DYN, HSD, nDTW, and Scene are component metrics. Complete results are provided in Appendix~\ref{app:ewmbench_full}.
}
\label{tab:agibot_transfer}
\setlength{\tabcolsep}{6.83pt}\fontsize{10}{10.90}\selectfont
\renewcommand{\arraystretch}{1.0}
\adjustbox{max width=\textwidth}{%
\begin{tabular}{lccccccc}
\toprule
Method
& Motion $\uparrow$
& Semantics $\uparrow$
& DYN $\uparrow$
& HSD $\uparrow$
& nDTW $\uparrow$
& Scene $\uparrow$
& Overall $\uparrow$ \\
\midrule
GigaWorld-0
& 50.30
& 2.2497
& 7.19
& 23.29
& 19.83
& \textbf{89.58}
& 3.6486 \\
Standard SFT
& 61.51
& 2.2477
& 14.88
& \textbf{24.35}
& \textbf{22.28}
& 84.38
& 3.7066 \\
\rowcolor{gray!20}
EVEWorld
& \textbf{63.65}
& \textbf{2.2524}
& \textbf{17.55}
& 24.12
& 21.97
& 86.37
& \textbf{3.7525} \\
\bottomrule
\end{tabular}%
}
\end{table}

\paragraph{Generalization Across Training Distributions.}
On EWMBench \citep{hu2025ewmbench}, \method{} trained on the same AgiBot data \citep{bu2025agibot} and optimization budget as Standard SFT improves Motion and Overall by 3.48\% and 1.24\%, respectively (Table~\ref{tab:agibot_transfer}). These results show that the gains of \method{} extend beyond the DreamGen training distribution, particularly in motion modeling.

\begin{table}[!t]
\centering
\caption{
\textbf{Cross-backbone evaluation and ablation on RoboTwin.}
PSNR, SSIM, and LPIPS measure frame-level visual fidelity against aligned RGB ground truth, and Flow-EPE denotes optical-flow endpoint error.
}
\label{tab:flowwam_transfer}
\setlength{\tabcolsep}{6.73pt}\fontsize{10}{10.90}\selectfont
\adjustbox{max width=\textwidth}{%
\begin{tabular}{lccccccc}
\toprule
Variant & IGR & TIA & PSNR (dB) $\uparrow$ & SSIM $\uparrow$ & LPIPS $\downarrow$ & Flow-EPE $\downarrow$ & MLR (\%) $\downarrow$ \\
\midrule
FlowWAM & & & 12.218 & 0.748 & 0.383 & 3.033 & 52.05 \\
Standard SFT & & & 11.687 & 0.735 & 0.414 & 2.828 & 47.89 \\
+ IGR & \checkmark & & 12.544 & 0.770 & 0.380 & 2.487 & 40.03 \\
+ TIA & & \checkmark & \textbf{13.445} & \textbf{0.776} & \textbf{0.333} & \textbf{1.850} & \textbf{22.54} \\
\rowcolor{gray!20}
EVEWorld & \checkmark & \checkmark & 12.765 & 0.769 & 0.365 & 2.207 & 35.21 \\
\bottomrule
\end{tabular}%
}
\end{table}

\paragraph{Cross-Backbone Generalization.}
We replace the GigaWorld image-text-conditioned IT2V backbone with FlowWAM's flow-conditioned WAM architecture (Table~\ref{tab:flowwam_transfer}). On FlowWAM, IGR improves all reported metrics over Standard SFT on held-out RoboTwin actions, but TIA yields consistently larger gains. The full model continues to outperform Standard SFT, suggesting that the contributions of IGR and TIA are backbone-dependent. The resulting FlowWAM-based EVEWorld model, SupervisionWM, achieves an EWMScore-P of 70.17 on the official WorldArena 2.0 Track 1 leaderboard, ranking 17th overall and 6th in JEPA Similarity. These results support the cross-backbone generalization of our evolution-supervision framework, while the optimal component configuration may depend on the model architecture.

\subsection{Ablation Studies and Analysis}
\label{sec:ablation}

\begin{table}[!t]
\centering
\caption{
\textbf{Component ablation on DreamGenBench.}
All variants share the same initialization, training data, and optimization budget. Gemini-IF measures instruction-following accuracy.
}
\label{tab:component_ablation}
\setlength{\tabcolsep}{8.48pt}\fontsize{10}{10.90}\selectfont
\renewcommand{\arraystretch}{1.0}

\adjustbox{max width=\textwidth}{%
\begin{tabular}{lccccccc}
\toprule
Variant
& \multicolumn{2}{c}{Component}
& \multicolumn{4}{c}{Gemini-IF (\%)}
& MLR (\%) $\downarrow$ \\
\cmidrule(lr){2-3}
\cmidrule(lr){4-7}
& IGR
& TIA
& Env $\uparrow$
& Object $\uparrow$
& Behavior $\uparrow$
& Overall $\uparrow$
& \\
\midrule

Standard SFT
& &
& 51.72
& 42.00
& 67.02
& 53.57
& 11.11 \\

IGR only
& \checkmark &
& 49.43
& 36.67
& \textbf{69.50}
& 51.85
& 4.76 \\

TIA only
& & \checkmark
& 55.17
& 42.67
& 68.79
& 55.29
& 7.94 \\

\rowcolor{gray!20}
EVEWorld
& \checkmark & \checkmark
& \textbf{72.41}
& \textbf{50.33}
& 64.89
& \textbf{60.85}
& \textbf{1.59} \\

\bottomrule
\end{tabular}%
}
\end{table}

\paragraph{Component Contributions of IGR and TIA.}
Table~\ref{tab:component_ablation} shows that IGR primarily improves instance consistency, reducing MLR from 11.11\% to 4.76\%; TIA improves cross-frame consistency and Overall Gemini-IF. Combining both yields 1.59\% MLR and a Gemini-IF score of 60.85, demonstrating complementary effects under the GigaWorld-0 image-text-conditioned IT2V architecture.

\begin{table}[!t]
\centering
\caption{
\textbf{Exploration of Alternative Approaches.}
PhysicsLatent uses weakly supervised process tokens; EAG uses energy-guided denoising; LAD-LoRA uses energy-regularized LoRA; Causal Frontier uses causal block-wise generation; ICH-D uses learned duplicate removal; and \method{} combines instance restoration with temporal alignment.
Each route is evaluated against its matched control; for \method{}, the matched control is Standard SFT.
Dyn., Train, Test, Causal, and Repair denote explicit dynamics/evolution modeling, training-time intervention, test-time intervention, causal generation, and learned repair, respectively.
$\Delta$ denotes relative change from the matched control.
Further details are provided in Appendix~\ref{app:failed_alternatives}.
}

\label{tab:failure_compare_main}

% Compact arrows.
\newcommand{\shortup}{%
  \raisebox{0.05ex}{\scalebox{0.9}[0.70]{$\uparrow$}}%
}
\newcommand{\shortdown}{%
  \raisebox{0.05ex}{\scalebox{0.9}[0.70]{$\downarrow$}}%
}

% Fixed-width arrow + value, ensuring identical alignment in every row.
\newcommand{\chg}[2]{%
  \makebox[1.05em][r]{#1}\hspace{0.20em}#2%
}

\setlength{\tabcolsep}{5.40pt}
\renewcommand{\arraystretch}{1.0}

{\small
\adjustbox{max width=\textwidth}{%
\begin{tabular}{
l
ccccc
cc
}

\toprule

Route
& \multicolumn{5}{c}{Design elements}
& \multicolumn{2}{c}{Relative change} \\
\cmidrule(lr){2-6}
\cmidrule(lr){7-8}

& Dyn.
& Train
& Test
& Causal
& Repair
& $\Delta$ MLR (\%) $\downarrow$
& $\Delta$ Gemini-IF (\%) $\uparrow$ \\
\midrule

PhysicsLatent
& \checkmark
& \checkmark
&
&
&
& \chg{\shortdown}{7.0}
& \chg{\shortdown}{11.9} \\

EAG sampling guidance
& \checkmark
&
& \checkmark
&
&
& \chg{\shortup}{150.0}
& \chg{\shortup}{10.0} \\

LAD-LoRA
& \checkmark
& \checkmark
&
&
&
& \chg{}{0.0}
& \chg{\shortdown}{6.1} \\

Causal Frontier
&
& \checkmark
&
& \checkmark
&
& \chg{\shortup}{35.4}
& \chg{\shortdown}{53.8} \\

ICH-D
& \checkmark
& \checkmark
& \checkmark
&
& \checkmark
& \chg{\shortup}{16.6}
& \chg{\shortdown}{30.6} \\

\midrule

\rowcolor{gray!20}
\method{} (ours)
& \checkmark
& \checkmark
&
&
& \checkmark
& \chg{\shortdown}{\textbf{85.7}}
& \chg{\shortup}{\textbf{13.6}} \\

\bottomrule

\end{tabular}%
}
}

\end{table}

\paragraph{Cross-Trial Robustness.}
We further evaluate \method{} on PBench Robot across multiple stochastic inference trials under a matched-budget setting, with evolution supervision as the only methodological difference from SFT$_{250}$. As shown in Table~\ref{tab:pbench_seed}, \method{} improves all six paired trials across 3.8\,s and 5.8\,s rollouts, with mean gains of $+1.54$ and $+1.56$, respectively, demonstrating consistent robustness across inference seeds and rollout horizons.

\begin{table}[!t]
\centering
\caption{
\textbf{Cross-trial robustness on PBench.}
Each SFT$_{250}$--\method{} pair uses the same inference seed, and the three trials use different seeds under otherwise identical settings. Scores denote the overall PBench Robot score, judged by Gemini-3.8-Flash. $\Delta$ denotes the gain over SFT$_{250}$, and Mean $\Delta$ averages the three paired gains.
}
\label{tab:pbench_seed}
\setlength{\tabcolsep}{3.5pt}
\renewcommand{\arraystretch}{1.0}

\adjustbox{max width=\textwidth}{%
\begin{tabular}{
c
ccc
ccc
ccc
c
}
\toprule
\multirow{2}{*}{Horizon}
& \multicolumn{3}{c}{Trial 1}
& \multicolumn{3}{c}{Trial 2}
& \multicolumn{3}{c}{Trial 3}
& \multirow{2}{*}{Mean $\Delta$} \\
\cmidrule(lr){2-4}
\cmidrule(lr){5-7}
\cmidrule(lr){8-10}

& SFT$_{250}$ & \method{} & $\Delta$
& SFT$_{250}$ & \method{} & $\Delta$
& SFT$_{250}$ & \method{} & $\Delta$
& \\
\midrule

3.8\,s
& 57.37
& \textbf{59.27}
& \textbf{$+1.90$}
& 53.40
& \textbf{55.07}
& \textbf{$+1.67$}
& 59.57
& \textbf{60.61}
& \textbf{$+1.04$}
& \textbf{$+1.54$} \\

5.8\,s
& 59.42
& \textbf{60.02}
& \textbf{$+0.60$}
& 60.36
& \textbf{63.02}
& \textbf{$+2.66$}
& 60.79
& \textbf{62.21}
& \textbf{$+1.42$}
& \textbf{$+1.56$} \\

\bottomrule
\end{tabular}%
}
\end{table}

\paragraph{Exploration of Alternative Approaches.}
We explored five alternative approaches to mitigate Model Laziness while improving instruction fidelity: PhysicsLatent, EAG, LAD-LoRA, Causal Frontier, and ICH-D (Table~\ref{tab:failure_compare_main}). However, none achieved both objectives simultaneously. For example, PhysicsLatent reduces Model Laziness at an 11.9\% cost in instruction fidelity, whereas EAG gains 10.0\% in instruction fidelity at a 150.0\% increase in Model Laziness. We share these trade-offs to provide practical insights for future research, with further details in Appendix~\ref{app:failed_alternatives}.

\begin{figure*}[!t]
  \centering
  \includegraphics[width=\textwidth]{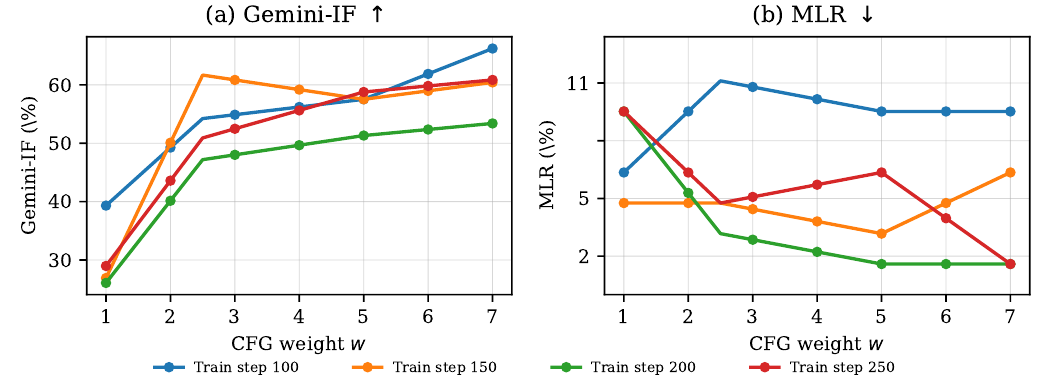}
\caption{
\textbf{CFG sensitivity across training stages on DreamGenBench.}
Gemini-IF and MLR are evaluated under different classifier-free guidance weights $w$, with each curve denoting a training checkpoint.
}
\label{fig:cfg}
\end{figure*}

\paragraph{CFG Sensitivity Analysis.}
Figure~\ref{fig:cfg} shows that Gemini-IF generally increases with higher CFG weights, but MLR varies non-monotonically across guidance weights and training stages. Thus, CFG does not reliably improve instruction fidelity and reduce Model Laziness simultaneously, indicating the importance of explicit supervision for target evolution.

\begin{table}[!t]
\centering
\caption{
\textbf{Sequence-length scaling and inference cost.}
We report latency (p50/p95: median/95th percentile), throughput, memory, and GPU cost across output durations.
}
\label{tab:horizon_compute_main}
\setlength{\tabcolsep}{8.5pt}\fontsize{10}{10.90}\selectfont
\renewcommand{\arraystretch}{1.0}
\adjustbox{max width=\textwidth}{%
\begin{tabular}{ccccccc}
\toprule
\multicolumn{2}{c}{Output} & \multicolumn{2}{c}{Latency (s) $\downarrow$} & Throughput & Memory & Cost \\
\cmidrule(lr){1-2}
\cmidrule(lr){3-4}
Dur. (s) & Frames & p50 & p95 & Frames/min $\uparrow$
& Peak/GPU (GiB) $\downarrow$ & GPU-h/1k $\downarrow$ \\
\midrule
3.8  & 61  & 8.166 & 8.403 & 442.1 & 38.19 & 18.40 \\
5.8  & 93  & 10.932 & 11.352 & 502.2 & 38.33 & 24.69 \\
9.8  & 157 & 18.323 & 19.249 & 504.8 & 38.99 & 41.46 \\
15.8 & 253 & 31.503 & 33.523 & 469.5 & 41.47 & 71.85 \\
19.8 & 317 & 41.737 & 45.007 & 443.3 & 40.70 & 95.34 \\
\bottomrule
\end{tabular}%
}
\end{table}

\paragraph{Sequence-Length Scaling and Inference Cost.}
Table~\ref{tab:horizon_compute_main} shows that increasing output duration from 3.8 to 19.8~s raises p50/p95 latency from 8.17/8.40 to 41.74/45.01~s and GPU-hour cost per 1,000 videos from 18.40 to 95.34, with relatively stable throughput and peak memory. Thus, longer rollouts mainly increase computation with limited additional memory overhead.

\FloatBarrier
\section{Conclusion}
We presented \method{}, a physical evolution supervision framework for embodied world models, and proposed the Model Laziness Rate (MLR), a metric to quantify target-instance consistency failures, which we term Model Laziness. \method{} combines IGR for instance consistency and TIA for cross-frame consistency to promote physically consistent target evolution. Experiments show an 85.7\% MLR reduction on DreamGenBench with improved instruction following, 17th overall and 6th in JEPA Similarity on the WorldArena 2.0 Track 1 leaderboard, and generalization gains across training distributions and model backbones.

\paragraph{Limitations and Future Works.}
The limitations of our current formulation mainly arise from its local, pairwise supervision, which does not explicitly capture long-horizon state dependencies or complex multi-object interactions. Future work will further expand the EVEWorld framework with broader forms of evolution supervision for longer-horizon and relational dynamics.

\clearpage
\subsection*{Reproducibility Statement}
The appendices document the training and evaluation settings. The supplementary material and the public repository provide code, data indices, configurations, evaluation prompts, and item-level results. The repository is available at \url{https://github.com/behappya/EVEWorld}.

\subsection*{AI use statement}

During the preparation of this manuscript, the authors used generative AI tools solely for writing assistance, including language polishing, grammar checking, and improving clarity of presentation. The AI tools were not used for developing the method, conducting experiments, generating results, or making scientific decisions. All AI-assisted edits were reviewed and verified by the authors, who take full responsibility for the final content of this manuscript.

\IfFileExists{iclr2027_conference.bst}{%
  \bibliographystyle{iclr2027_conference}
}{%
  \bibliographystyle{plainnat}
}
\bibliography{eveworld_iclr2027_2}

\clearpage
\appendix
\section*{Appendix Overview}

This supplementary material provides additional details and results for
\method{}, organized as follows:

\begin{itemize}[
    leftmargin=1.4em,
    itemsep=5pt,
    topsep=4pt,
    parsep=0pt
]

\item \hyperref[app:mlr]{\textbf{Appendix A: Model Laziness and MLR}}
    \begin{itemize}[
        label={},
        leftmargin=2.0em,
        itemindent=0pt,
        itemsep=1.5pt,
        topsep=2pt,
        parsep=0pt
    ]
        \item \hyperref[app:mlr_protocol]{\textbf{A.1} MLR: Comparison with Existing Evaluation Metrics}
        \item \hyperref[app:mlr_scope]{\textbf{A.2} Scope and Limitations of MLR}
        \item \hyperref[app:mlr_detector]{\textbf{A.3} Robustness and Sensitivity Analysis of MLR}
        \item \hyperref[app:mlr_cross_model]{\textbf{A.4} Cross-Model Evaluation of MLR}
        \item \hyperref[app:mlr_cross_domain]{\textbf{A.5} Cross-Domain Evaluation of MLR}
        \item \hyperref[sec:vlm_mlr_audit]{\textbf{A.6} Optional VLM Evaluation}
    \end{itemize}

\item \hyperref[app:method_details]{\textbf{Appendix B: Method Details}}
    \begin{itemize}[
        label={},
        leftmargin=2.0em,
        itemindent=0pt,
        itemsep=1.5pt,
        topsep=2pt,
        parsep=0pt
    ]
        \item \hyperref[app:igr_details]{\textbf{B.1} IGR Construction}
        \item \hyperref[app:tia_details]{\textbf{B.2} TIA Alignment}
        \item \hyperref[app:objective]{\textbf{B.3} Joint Training and Inference}
    \end{itemize}

\item \hyperref[app:extended]{\textbf{Appendix C: Additional Experiments and Analysis}}
    \begin{itemize}[
        label={},
        leftmargin=2.0em,
        itemindent=0pt,
        itemsep=1.5pt,
        topsep=2pt,
        parsep=0pt
    ]
        \item \hyperref[app:training_setup]{\textbf{C.1} Implementation Details}
        \item \hyperref[app:igr_variants]{\textbf{C.2} IGR Ablation and Duplicate Restoration}
        \item \hyperref[app:evaluation_stability]{\textbf{C.3} Evaluation Stability}
        \item \hyperref[app:training_sampling_sensitivity]{\textbf{C.4} Training and Long-Horizon Behavior}
    \end{itemize}

\item \hyperref[app:failed_alternatives]{\textbf{Appendix D: Exploration of Alternative Approaches}}
    \begin{itemize}[
        label={},
        leftmargin=2.0em,
        itemindent=0pt,
        itemsep=1.5pt,
        topsep=2pt,
        parsep=0pt
    ]
        \item \hyperref[app:physicslatent]{\textbf{D.1} PhysicsLatent}
        \item \hyperref[app:eag]{\textbf{D.2} EAG}
        \item \hyperref[app:lad_lora]{\textbf{D.3} LAD-LoRA}
        \item \hyperref[app:causal_frontier]{\textbf{D.4} Causal Frontier}
        \item \hyperref[app:ichd]{\textbf{D.5} ICH-D}
    \end{itemize}

\item \hyperref[app:flowwam_benchmarks]{\textbf{Appendix E: Generalization Results}}
    \begin{itemize}[
        label={},
        leftmargin=2.0em,
        itemindent=0pt,
        itemsep=1.5pt,
        topsep=2pt,
        parsep=0pt
    ]
        \item \hyperref[app:ewmbench_full]{\textbf{E.1} Generalization on EWMBench}
        \item \hyperref[app:heldout_action]{\textbf{E.2} FlowWAM Checkpoint Robustness}
    \end{itemize}

\item \hyperref[app:visuals]{\textbf{Appendix F: Qualitative Results}}
    \begin{itemize}[
        label={},
        leftmargin=2.0em,
        itemindent=0pt,
        itemsep=1.5pt,
        topsep=2pt,
        parsep=0pt
    ]
        \item \hyperref[app:qual_laziness]{\textbf{F.1} Instance Consistency}
        \item \hyperref[app:qual_matched]{\textbf{F.2} Cross-Frame Consistency}
        \item \hyperref[app:qual_distortion]{\textbf{F.3} Target Evolution Comparison}
    \end{itemize}

\item \hyperref[app:related_work]{\textbf{Appendix G: Related Work}}

% \item \hyperref[app:implementation]{\textbf{Appendix H: Reproducibility and Limitations}}

\end{itemize}

\section{Model Laziness and MLR}
% [中文] 模型偷懒（Model Laziness）与 MLR
\label{app:mlr}
MLR detects persistent deviations from the initial target-instance count $N_0$, while under-counts supported by robot-induced occlusion are exempted and all over-counts are retained. Appendix~\ref{app:retention_check} tests recovery from an injected count error. Appendix~\ref{app:mlr_protocol} details the evaluation protocol and occlusion check; Appendices~\ref{app:mlr_cross_model}--\ref{app:mlr_cross_domain} cover model, detector, and domain variation.

\subsection{MLR: Comparison with Existing Evaluation Metrics}
\label{app:metric_context}
\label{app:mlr_protocol}

Existing evaluations expose signals related to object persistence and quantity, but differ in both their evaluation anchors and aggregation units.

\noindent\textbf{RBench.}
RBench evaluates spontaneous entity emergence and disappearance as part of Physical-Semantic Plausibility (PSS), together with other physical-semantic violations such as floating, penetration, and non-contact attachment \citep{deng2026rethinking}. After normalization, PSS is combined with Task-Adherence Consistency (TAC) into the Task Completion score:
\begin{equation}
\mathrm{TC}_{\mathrm{RBench}}
=
\frac{\mathrm{PSS}+\mathrm{TAC}}{2}.
\label{eq:rbench_tc}
\end{equation}
Thus, appearance and disappearance failures contribute to a broader assessment of physical plausibility and task correctness rather than forming an independent rollout-level target-instance consistency metric. In contrast, MLR isolates target-instance consistency as a dedicated evaluation unit, making instance-consistency failures directly measurable rather than subsuming them into a broader composite score.

\noindent\textbf{WFM-Eval.}
WFM-Eval explicitly diagnoses object hallucination and temporal inconsistency by comparing a generated video $\hat V$ with a ground-truth reference video $V^\ast$ \citep{khose2026wfmeval}. It distinguishes phantom appearance, phantom disappearance, position drift, and frame jump. Let $n_{\mathrm{pa}}$, $n_{\mathrm{pd}}$, $n_{\mathrm{dr}}$, and $n_{\mathrm{jp}}$ denote the numbers of these detected events over $M$ generated frames and $K$ tracked objects. Its Object Hallucination Rate (OHR) and Harm-weighted Severity Score (HSS) are
\begin{equation}
\mathrm{OHR}_{\mathrm{WFM}}
=
\frac{n_{\mathrm{pa}}+n_{\mathrm{pd}}}{M},
\qquad
\mathrm{HSS}_{\mathrm{WFM}}
=
\min\!\left(
1,
\frac{\sum_{e\in\mathcal{E}} w(e)}{MK}
\right),
\label{eq:wfmeval_metrics}
\end{equation}
where $w(e)$ assigns event-dependent severity weights. These metrics provide fine-grained reference-grounded diagnostics. MLR instead anchors evaluation to the observed initial target state and therefore requires no ground-truth future, enabling target-instance consistency to be evaluated when only the conditioning observation and generated rollout are available.

\noindent\textbf{T2V-CompBench.}
Generative numeracy in T2V-CompBench is prompt-anchored \citep{sun2024t2vcompbench}. GroundingDINO counts each prompted object class in every evaluated frame; a class receives score 1 if the detected quantity matches the quantity specified in the text prompt and 0 otherwise. The frame score averages over object classes, and the video score averages over frames. This rule can be written as
\begin{equation}
S_{\mathrm{num}}
=
\frac{1}{T}
\sum_{t=1}^{T}
\frac{1}{J}
\sum_{j=1}^{J}
\mathbb{I}\!\left[
\hat n_{t,j}
=
n^{\mathrm{prompt}}_{j}
\right],
\label{eq:t2v_numeracy}
\end{equation}
where $\hat n_{t,j}$ is the detected count of object class $j$ at frame $t$, $n^{\mathrm{prompt}}_{j}$ is its prompt-specified quantity, and $J$ is the number of queried object classes. This metric therefore evaluates frame-wise agreement with prompt-specified quantities rather than target-instance consistency with the observed target population throughout an interaction. MLR instead derives the expected target count from the conditioning observation itself, allowing target-instance consistency to be evaluated even when the instruction does not explicitly specify a quantity.

\noindent\textbf{RoboWM-Bench.}
RoboWM-Bench evaluates a different property: embodied executability \citep{jiang2026robowm}. Predicted human-hand or robotic videos are converted into executable actions through human-centric retargeting or a robot-centric inverse dynamics model (IDM), respectively, and the resulting actions are executed in simulation. Evaluation combines step-level checks at semantically meaningful action nodes with final task-level success. If $C_r$ denotes whether key action node $r$ satisfies its required interaction and dynamical constraints and $G$ denotes achievement of the final task objective, its stated task-success criterion can be summarized as
\begin{equation}
S_{\mathrm{task}}
=
\mathbb{I}\!\left[
\left(\bigwedge_{r=1}^{R} C_r\right)
\land G
\right].
\label{eq:robowm_exec}
\end{equation}
This protocol measures whether generated behavior can be translated into successful embodied execution rather than directly measuring target-instance consistency in the generated rollout. MLR provides a complementary model-side diagnostic by directly identifying persistent target-count violations in the generated video, without requiring action decoding or simulator execution.

\paragraph{Positioning and Design Advantages of MLR.}
Taken together, these evaluations address complementary aspects of video and embodied-world-model quality but rely on different evaluation anchors. RBench embeds appearance and disappearance within broader physical and task-level assessment; WFM-Eval evaluates hallucination events relative to a ground-truth future; T2V-CompBench evaluates quantities explicitly specified by the prompt; and RoboWM-Bench assesses downstream embodied execution. MLR is designed for the complementary regime in which the conditioning observation itself defines the expected target count and no reference future is assumed. It therefore provides four properties that are particularly useful for embodied manipulation evaluation: it is \emph{initial-state anchored}, because the observed initial target count $N_0$ is obtained directly from the conditioning state; \emph{reference-free}, because no ground-truth future is required; \emph{rollout-level and persistence-aware}, because sustained violations are distinguished from isolated frame-level fluctuations; and \emph{occlusion-aware}, because robot-supported under-counts are explicitly exempted. Together, these properties make MLR a dedicated diagnostic for persistent target-instance consistency failures in embodied world-model rollouts.

We operationalize MLR as follows. For DreamGenBench, eligibility is determined once from the fixed conditioning frame with GroundingDINO \citep{liu2024groundingdino} and the same target query, yielding the shared set $\mathcal{U}_{63}$ used for every model. GroundingDINO then counts target instances at $T$ uniformly sampled future timestamps in each generated video. Spatially redundant detections are merged; items with no detectable target in the conditioning frame are excluded from the shared set. We denote the eligible set by
\[
\mathcal{D}_{+}
=
\{x\in\mathcal{D}:N_0(x)>0\}.
\]

We first identify deviations from the initial target-instance count $N_0$. Over-counts, i.e., $N_t>N_0$, are always retained as duplication evidence. For under-counts, i.e., $N_t<N_0$, we further check whether the missing target is supported by robot-induced occlusion.

SAM2 \citep{ravi2024sam2} tracks each target instance across the rollout. When the current target mask of an instance is unavailable, we use its most recent reliable mask as the reference target region. GroundingDINO localizes the robot in the current frame, and SAM2 converts the detected robot boxes into an occluder mask $M_t^{\mathrm{robot}}$. For each missing target instance $i$, we compute the target--robot overlap ratio
\begin{equation}
r^{\mathrm{occ}}_{i,t}
=
\frac{
\left|
\widehat{M}^{\mathrm{target}}_{i,t}
\cap
M_t^{\mathrm{robot}}
\right|
}{
\left|
\widehat{M}^{\mathrm{target}}_{i,t}
\right|
},
\label{eq:occ_overlap}
\end{equation}
where $\widehat{M}^{\mathrm{target}}_{i,t}$ denotes the reliable target reference region used for the occlusion check.

Motivated by prior amodal-segmentation criteria
\citep{xiao2021amodal,zeng2022jigsaw,zhan2024amodal} and
target-normalized occlusion reasoning in robotic manipulation
\citep{wada2018instance}, we set $\tau_{\mathrm{occ}}=0.15$ and
treat an under-count as occlusion-supported only if
$r^{\mathrm{occ}}_{i,t}\geq\tau_{\mathrm{occ}}$ for every missing
target instance; Appendix~\ref{sec:mlr_sensitivity} confirms that
the comparison is robust to nearby thresholds. Such under-counts are excluded from violation evidence, whereas all other under-counts and all over-counts are retained. Let $o_t(x)=1$ indicate an occlusion-supported under-count. We define the resulting evaluation count as
\begin{equation}
\widetilde{N}_t(x)
=
\begin{cases}
N_0(x), & o_t(x)=1,\\
N_t(x), & \text{otherwise}.
\end{cases}
\label{eq:mlr_adjusted_count}
\end{equation}

For each eligible video, an MLR event is recorded when the adjusted count deviates from $N_0$ at two consecutive sampled timestamps:
\begin{align}
e(x)
&=
\mathbb{I}\!\left[
\exists t\in\{1,\ldots,T-1\}:
\widetilde{N}_t(x)\neq N_0(x)
\land
\widetilde{N}_{t+1}(x)\neq N_0(x)
\right],
\notag\\
\mathrm{MLR}
&=
\frac{100}{|\mathcal{D}_{+}|}
\sum_{x\in\mathcal{D}_{+}}e(x).
\label{eq:app_mlr}
\end{align}

Unlike frame-wise count agreement, MLR records a violation only when the inconsistency persists at the rollout level. An occlusion-supported under-count resets the persistence criterion, whereas a target that remains missing after the robot occluder moves away again provides disappearance evidence. Requiring two consecutive violating timestamps suppresses isolated detector errors while retaining sustained duplication and disappearance as consistency failures. All detector, query, occlusion-threshold, and persistence settings are fixed across models.

Beyond the scalar MLR value, the same event definition localizes each violation in time. For an MLR-positive rollout, we record the first persistent violation as its onset and analyze how long the count deviation persists. These diagnostics distinguish short detector fluctuations from sustained consistency failures and enable direct inspection of the corresponding generated segment.

\begin{algorithm}[H]
\caption{\textbf{Occlusion-aware Model Laziness Rate evaluation.}
All detector, query, occlusion, and persistence settings are fixed across models.}
\label{alg:mlr_evaluation}
\small
\textbf{Input:} generated set $\mathcal{D}$, target-entity query $q(c)$, fixed detector $G$, sampled timestamp set $\mathcal{T}$.\\
\textbf{Output:} MLR, detector coverage, and per-video event labels.
\begin{enumerate}[leftmargin=1.6em,itemsep=1pt,topsep=2pt]

  \item For each video $x$, run $G$ with $q(c)$ at all sampled timestamps, merge spatially redundant target detections, and obtain $N_t(x)$. If $N_0(x)=0$, mark $x$ as ineligible.

  \item If $N_t(x)>N_0(x)$, retain the deviation as duplication evidence. If $N_t(x)<N_0(x)$, proceed to the occlusion check.

  \item Track target instances with SAM2 and obtain the robot occluder mask in the current frame. For each missing instance, compute $r^{\mathrm{occ}}_{i,t}$ using Equation~\ref{eq:occ_overlap}.

  \item Exempt an under-count only if every missing target instance satisfies $r^{\mathrm{occ}}_{i,t}\geq0.15$. Set $\widetilde{N}_t(x)=N_0(x)$ for such timestamps and $\widetilde{N}_t(x)=N_t(x)$ otherwise.

  \item For every eligible video, set $e(x)=1$ iff $\widetilde{N}_t(x)\neq N_0(x)$ at two consecutive sampled timestamps; otherwise set $e(x)=0$.

  \item Report
  $\mathrm{MLR}=100\sum_{x\in\mathcal{D}_{+}}e(x)/|\mathcal{D}_{+}|$
  and coverage $=100|\mathcal{D}_{+}|/|\mathcal{D}|$, together with event and eligible counts.

\end{enumerate}
\end{algorithm}

\begin{figure}[htbp]
  \centering
  \safeincludegraphics[width=\textwidth]{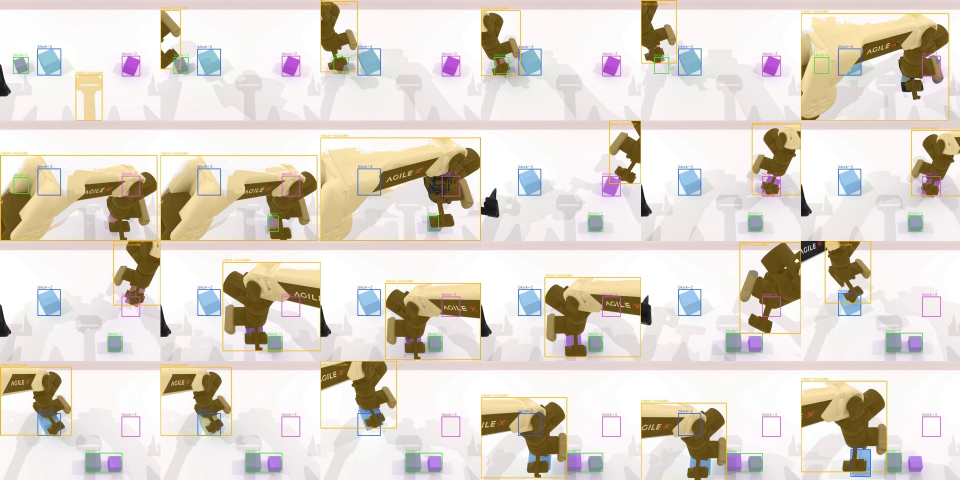}
  \caption{
  \textbf{Example of occlusion-aware MLR evaluation.}
  The rollout shows target detections and robot occlusion across sampled timestamps.
  Under-counts supported by sufficient target--robot overlap are excluded from violation evidence, whereas under-counts that persist without sufficient overlap are retained as target-disappearance evidence.
  }
  \label{fig:mlr_occlusion}
\end{figure}

\paragraph{Occlusion example.}
Figure~\ref{fig:mlr_occlusion} illustrates how the occlusion check distinguishes temporary robot occlusion from target disappearance. During sufficient robot--target overlap, a missing target is excluded from violation evidence. Once the robot moves away, a target that remains missing is again treated as an under-count and contributes to an MLR event if the deviation persists across consecutive sampled timestamps.

\begin{figure}[htbp]
  \centering
  \safeincludegraphics[width=\textwidth]{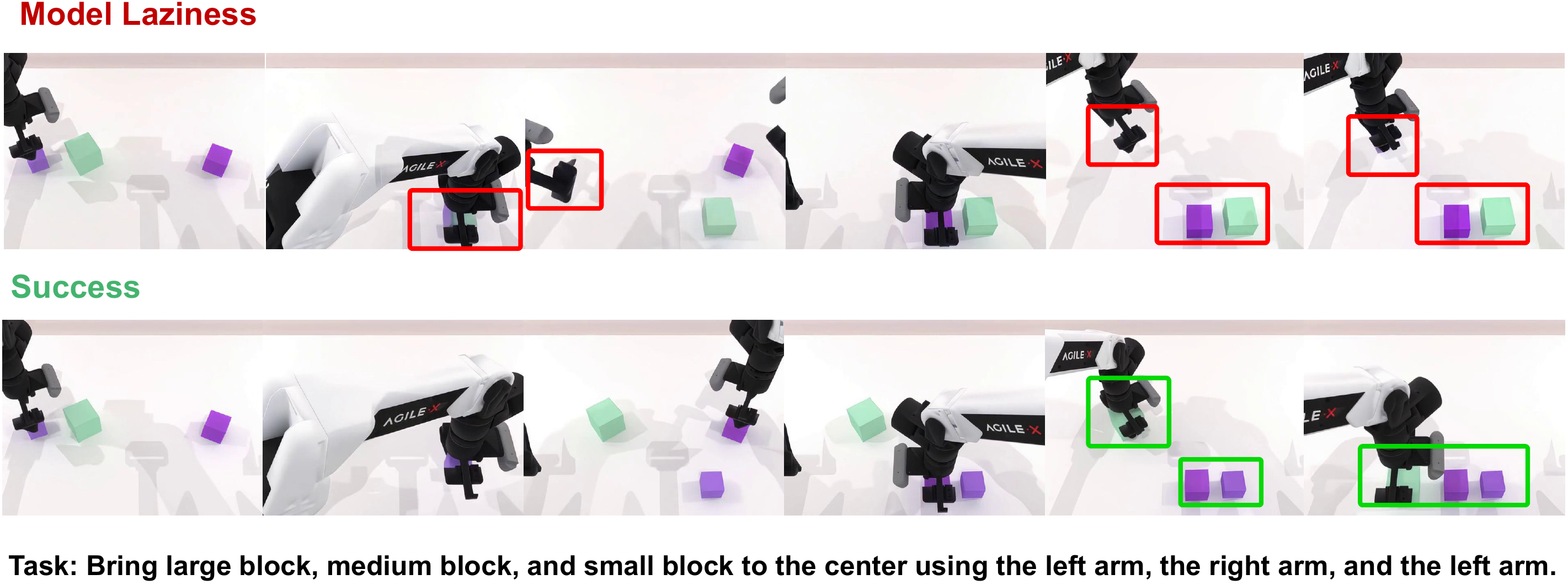}
  \caption{
  \textbf{Examples of MLR counting.}
  \textcolor{red}{Red} marks a persistent count violation retained by MLR.
  \textcolor{green!50!black}{Green} marks an under-count excluded by the occlusion-aware evaluation.
  A disappearance is counted only when under-count evidence persists outside occlusion-supported timestamps.
  }
  \label{fig:mlr_pipeline}
\end{figure}

\subsection{Scope and Limitations of MLR}
\label{app:mlr_scope}

MLR is designed for manipulation scenarios that require instance consistency, including grasping, transport, placement, and rearrangement, where the observed initial target count $N_0$ is expected to remain unchanged throughout the interaction. The protocol does not assume $N_0=1$ and therefore applies to scenes containing multiple instances of the manipulated target.

Interactions that intentionally change the target-instance count, such as cutting, splitting, assembly, or disassembly, fall outside the current formulation because the expected target count can change as part of successful task execution. Evaluating such tasks would require an instruction-conditioned expected count rather than a fixed initial-state reference.

MLR is also deliberately specific to target-instance consistency. Count-preserving failures, such as identity drift, deformation, or other changes that preserve the number of detected target instances, are not captured by MLR. These failures constitute a complementary form of temporal inconsistency and motivate the cross-frame supervision introduced by TIA.

\begin{figure}[htbp]
  \centering
  \safeincludegraphics[width=\textwidth]{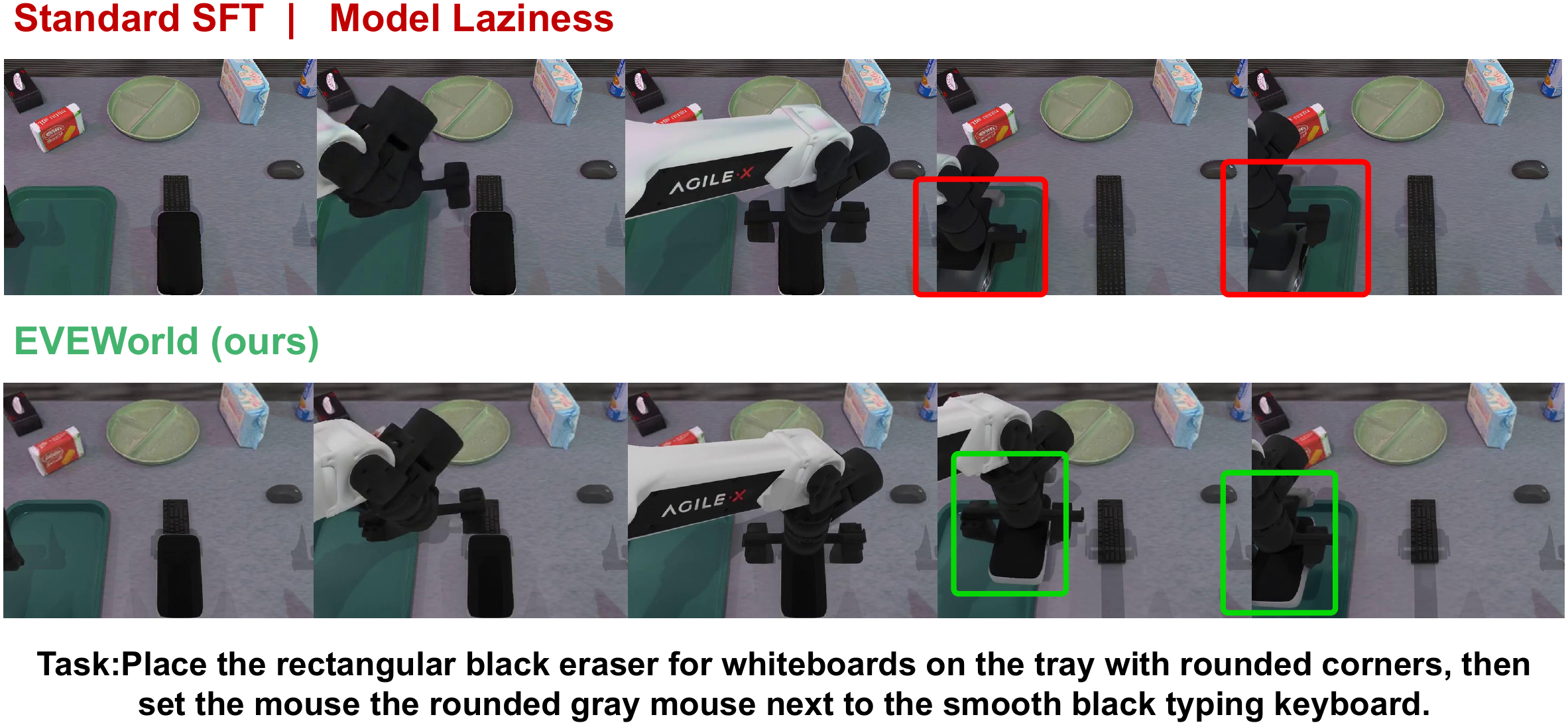}
\caption{
\textbf{Count-preserving cross-frame distortion.}
\textcolor{red}{Red} marks Standard SFT, where the target deforms despite preserving its instance count.
\textcolor{green!50!black}{Green} marks \method{}, which preserves target shape and temporal consistency.
}
  \label{fig:conservation_distortion}
\end{figure}

\subsection{Robustness and Sensitivity Analysis of MLR}
\label{app:mlr_detector}

Because MLR is derived from automatic target counts, we examine its robustness to evaluation-time detection errors, robot-induced occlusion, and the temporal-persistence criterion. The final protocol combines spatial merging, occlusion-aware handling of under-counts, and temporal persistence, with all detector, query, and evaluation settings fixed across models.

\paragraph{Detector and occlusion robustness.}
Spatial merging suppresses redundant detections of the same target, while the occlusion check prevents robot-induced under-counts from being misclassified as target disappearance. Only under-counts are eligible for the occlusion exemption; over-counts are always retained as duplication evidence. An occlusion-supported under-count is treated as non-violating and resets temporal persistence, whereas a target that remains missing after the robot occluder moves away again provides disappearance evidence.

Table~\ref{tab:detector_calibration} contrasts the raw count-deviation rate with the final MLR protocol. The substantial difference between raw detector deviations and the final event rate illustrates the importance of spatial merging, occlusion-aware filtering, and temporal persistence rather than treating every frame-level count mismatch as a physical-consistency failure.

\begin{table}[htbp]
\centering
\caption{
\textbf{Detector calibration for MLR.}
Detector v1 denotes the raw count-deviation rate before spatial merging, occlusion-aware under-count handling, and temporal persistence.
Peak dev. denotes the mean maximum raw count deviation, $\max_t |N_t-N_0|$, over eligible videos.
MLR and Events / elig. are computed using the final occlusion-aware protocol.
}
\label{tab:detector_calibration}
\setlength{\tabcolsep}{5.5pt}
\begin{tabular}{lcccc}
\toprule
Model / config.\ & Detector v1 (\%) & MLR (\%) $\downarrow$ & Events / elig.\ & Peak dev.\ \\
\midrule
GigaWorld-0 (Pretrained) & 62.1 & 12.70 & 8 / 63 & 0.23 \\
Official GR1 Fine-tune & 68.8 & 7.94 & 5 / 63 & 0.14 \\
Standard SFT & 63.2 & 11.11 & 7 / 63 & 0.27 \\
EVEWorld (step 100) & 70.4 & 4.76 & 3 / 63 & 0.17 \\
EVEWorld (seed 6666, step 250) & 66.7 & 3.17 & 2 / 63 & 0.15 \\
EVEWorld (u3, step 300) & 68.0 & 3.17 & 2 / 63 & 0.03 \\
\rowcolor{gray!20}
EVEWorld (step 250, main) & 70.4 & 1.59 & 1 / 63 & 0.11 \\
\bottomrule
\end{tabular}
\end{table}

\paragraph{Temporal persistence.}
For each eligible video, we record whether an adjusted count deviation $\widetilde{N}_t(x)\neq N_0(x)$ persists for at least $k$ consecutive sampled timestamps; $k{=}2$ is the default MLR rule. This persistence requirement suppresses isolated detector fluctuations while retaining sustained target duplication or disappearance.

\begin{figure}[htbp]
  \centering
  \safeincludegraphics[width=\textwidth]{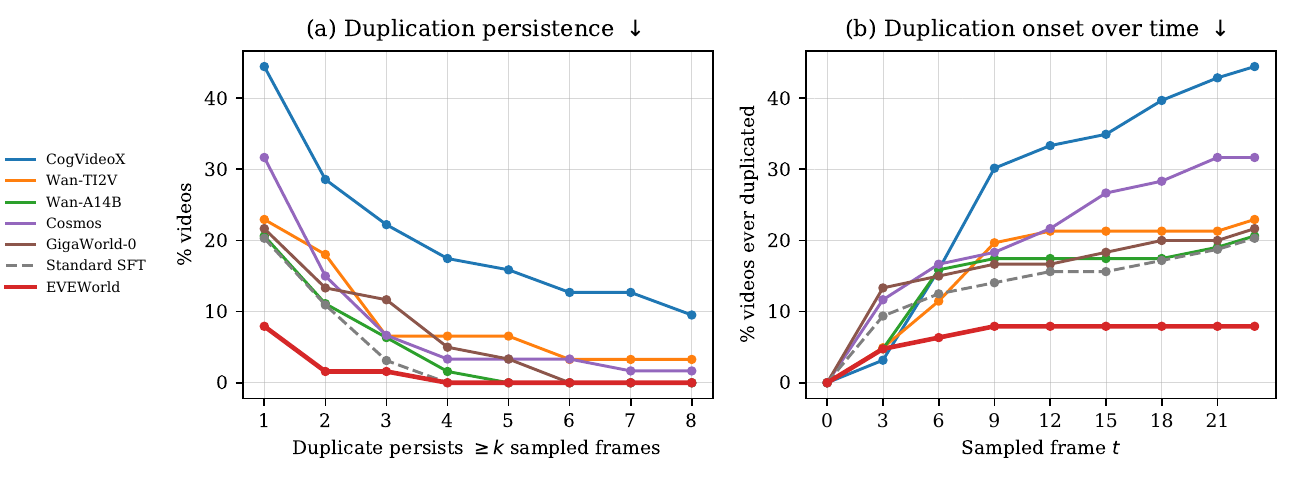}
\caption{
\textbf{Persistence of instance consistency violations across models.}
(a) Fraction of detector-eligible DreamGenBench videos with occlusion-aware count violations lasting at least $k$ consecutive sampled timestamps;
(b) cumulative fraction of videos that have exhibited such a violation by each sampled timestamp.
}
\label{fig:persistence}
\end{figure}

Figure~\ref{fig:persistence} reports the persistence length and cumulative onset of occlusion-aware target-instance consistency violations across models. The sole violation under \method{} is confined to short persistence lengths, whereas violations in the general I2V baselines extend to larger $k$.

\paragraph{Sensitivity to occlusion and persistence thresholds.}
\label{sec:mlr_sensitivity}
The main protocol sets $\tau_{\mathrm{occ}}=0.15$, motivated by prior amodal-segmentation protocols that use approximately 15--17\% occluded-area criteria to identify nontrivial occlusion \citep{xiao2021amodal,zeng2022jigsaw,zhan2024amodal}, and by target-normalized inter-instance occlusion reasoning in robotic manipulation \citep{wada2018instance}.

We recompute MLR on the same shared set $\mathcal{U}_{63}$ with
$\tau_{\mathrm{occ}}\in\{0.10,0.15,0.20,0.25\}$ and
$k\in\{1,2,3\}$, while keeping all generated videos, detections,
tracks, masks, queries, and eligible items fixed.
Across all configurations, \method{} maintains a lower MLR than
Standard SFT, with relative reductions ranging from 70.0\% to
88.9\%.
Under the main setting, MLR decreases from $7/63$ ($11.11\%$) to
$1/63$ ($1.59\%$), corresponding to an 85.7\% relative reduction.

Lower $\tau_{\mathrm{occ}}$ exempts more under-counts as
robot-supported occlusions, whereas larger $\tau_{\mathrm{occ}}$
retains more under-counts as MLR candidates; increasing $k$ removes
short-lived deviations. Neither change reverses the comparison between
Standard SFT and \method{}.

\begin{table}[htbp]
\centering
\caption{
\textbf{Sensitivity of MLR to the occlusion-overlap threshold
$\tau_{\mathrm{occ}}$ and persistence requirement $k$.}
Each cell reports events / 63 (MLR in \%).
All settings use the same eligible set $\mathcal{U}_{63}$.
The main setting, $\tau_{\mathrm{occ}}=0.15$ and $k=2$, is shaded.
}
\label{tab:mlr_rule_sensitivity}

\setlength{\tabcolsep}{5.2pt}
\renewcommand{\arraystretch}{1.08}

\adjustbox{max width=\textwidth}{%
\begin{tabular}{c ccc ccc}
\toprule
& \multicolumn{3}{c}{Standard SFT}
& \multicolumn{3}{c}{EVEWorld} \\
\cmidrule(lr){2-4}
\cmidrule(lr){5-7}

$\tau_{\mathrm{occ}}$
& $k=1$ & $k=2$ & $k=3$
& $k=1$ & $k=2$ & $k=3$ \\
\midrule

0.10
& 11 / 63 (17.46)
& 5 / 63 (7.94)
& 4 / 63 (6.35)
& 2 / 63 (3.17)
& 1 / 63 (1.59)
& 1 / 63 (1.59) \\

0.15
& 14 / 63 (22.22)
& \cellcolor{gray!20}\textbf{7 / 63 (11.11)}
& 5 / 63 (7.94)
& 3 / 63 (4.76)
& \cellcolor{gray!20}\textbf{1 / 63 (1.59)}
& 1 / 63 (1.59) \\

0.20
& 17 / 63 (26.98)
& 9 / 63 (14.29)
& 6 / 63 (9.52)
& 3 / 63 (4.76)
& 1 / 63 (1.59)
& 1 / 63 (1.59) \\

0.25
& 20 / 63 (31.75)
& 10 / 63 (15.87)
& 7 / 63 (11.11)
& 4 / 63 (6.35)
& 3 / 63 (4.76)
& 1 / 63 (1.59) \\

\bottomrule
\end{tabular}
}
\end{table}
\subsection{Cross-Model Evaluation of MLR}
\label{app:mlr_cross_model}

We apply the same occlusion-aware MLR protocol across model families on DreamGenBench. For the controlled cross-model comparison, all event labels are evaluated on the same shared set of 63 detector-eligible items. Table~\ref{tab:mlr_dgb} reports event counts, eligible counts, 95\% Wilson confidence intervals, and coverage.

Across the evaluated model families, MLR ranges from 7.94--28.57\%; CogVideoX1.5, Wan2.2-TI2V, Cosmos-Predict2, and pretrained GigaWorld-0 reach 28.57\%, 17.46\%, 14.29\%, and 12.70\%, respectively. These results show that persistent target-instance consistency failures occur across multiple video-generation and embodied-world-model families rather than being specific to Standard SFT or the GigaWorld backbone. On the same shared evaluation set, \method{} attains an MLR of 1.59\%.

\begin{table}[htbp]
\centering
\caption{
\textbf{Cross-model MLR} on DreamGenBench.
Events / eligible denotes the number of Model Laziness cases over the shared 63-item set.
95\% Wilson CI denotes the confidence interval of MLR, and Coverage denotes the fraction of detector-eligible prompts.
}
\label{tab:mlr_dgb}
\setlength{\tabcolsep}{5.6pt}
\begin{tabular}{lcccc}
\toprule
Model & Events / eligible & MLR (\%) $\downarrow$ & 95\% Wilson CI & Coverage (\%) \\
\midrule
GigaWorld-0 (Pretrained) & 8 / 63 & 12.70 & [6.58, 23.11] & 50.00 \\
Official GR1 Fine-tune & 5 / 63 & 7.94 & [3.44, 17.27] & 50.00 \\
Standard SFT & 7 / 63 & 11.11 & [5.49, 21.20] & 50.00 \\
Wan2.2-TI2V-5B & 11 / 63 & 17.46 & [10.04, 28.62] & 50.00 \\
Wan2.2-I2V-A14B & 7 / 63 & 11.11 & [5.49, 21.20] & 50.00 \\
CogVideoX1.5-5B-I2V & 18 / 63 & 28.57 & [18.90, 40.70] & 50.00 \\
Cosmos-Predict2-2B & 9 / 63 & 14.29 & [7.70, 24.97] & 50.00 \\
\rowcolor{gray!20}
EVEWorld & \textbf{1 / 63} & \textbf{1.59} & \textbf{[0.28, 8.46]} & 50.00 \\
\bottomrule
\end{tabular}
\end{table}

\subsection{Cross-Domain Evaluation of MLR}
\label{app:mlr_cross_domain}

We further evaluate the same occlusion-aware MLR protocol on WorldArena 1.0 \citep{shang2026worldarena} using the officially released evaluation code locally. An instruction-only parser first selects manipulation prompts with an identifiable target (817 of 1{,}000), after which we apply the same initial-frame eligibility rule and MLR protocol used on DreamGenBench.

Table~\ref{tab:mlr_wa} reports the same cross-domain comparison as main-text Table~\ref{tab:worldarena}, on the shared set of 157 detector-eligible videos and with 95\% Wilson confidence intervals for every MLR value. MLR decreases from 27.39\% under Standard SFT to 13.38\% under \method{}, showing that the reduction in persistent target-instance consistency failures extends beyond the DreamGen domain. On this shared set the two baselines coincide at 27.39\% while violating on different videos.

\begin{table}[htbp]
\centering
\caption{
\textbf{Cross-domain MLR} on WorldArena 1.0.
Results use the same occlusion-aware protocol and the same shared set of 157 detector-eligible videos as main-text Table~\ref{tab:worldarena}, with 95\% Wilson confidence intervals.
}
\label{tab:mlr_wa}
\setlength{\tabcolsep}{5.9pt}
\begin{tabular}{lcccc}
\toprule
Model & Events / eligible & MLR (\%) $\downarrow$ & 95\% Wilson CI & Coverage (\%) \\
\midrule
GigaWorld-0 (Pretrained) & 43 / 157 & 27.39 & [21.02, 34.84] & 19.22 \\
Standard SFT & 43 / 157 & 27.39 & [21.02, 34.84] & 19.22 \\
\rowcolor{gray!20}
EVEWorld & \textbf{21 / 157} & \textbf{13.38} & \textbf{[8.92, 19.58]} & 19.22 \\
\bottomrule
\end{tabular}
\end{table}

% [disabled, round 27] The WorldArena per-model MLR table was removed because it
% duplicated the same models and benchmark as main-text Table 3 with different
% values. Original content is kept below for restoration after re-measuring.
% \begin{table}[htbp]
% \centering
% \caption{
% \textbf{Per-model MLR} on WorldArena 1.0.
% Each model is evaluated under the same occlusion-aware MLR protocol.
% }
% \label{tab:mlr_wa_permodel}
% \setlength{\tabcolsep}{18.0pt}
% \begin{tabular}{lcc}
% \toprule
% Model & MLR (\%) $\downarrow$ & 95\% Wilson CI \\
% \midrule
% CogVideoX1.5-5B-I2V & 1.24 & [0.53, 2.88] \\
% Wan2.2-TI2V-5B & 14.42 & [11.02, 18.64] \\
% Wan2.2-I2V-A14B & 24.90 & [19.90, 30.67] \\
% Cosmos-Predict2-2B & 7.24 & [4.83, 10.71] \\
% \bottomrule
% \end{tabular}
% \end{table}
%
% Original paragraph that described the removed table:
% Table~\ref{tab:mlr_wa_permodel} additionally reports MLR for the
% general-purpose video models evaluated on WorldArena. Notably, CogVideoX
% exhibits a low MLR despite having the lowest EWMScore-local-8 in main-text
% Table~\ref{tab:worldarena}, illustrating that MLR captures target-instance
% consistency specifically rather than overall perceptual or task quality.
%
% Previous 182-video values of the retained table above, for reference:
% GigaWorld-0 (Pretrained) & 46 / 182 & 25.27 & [19.52, 32.05] & 22.28
% Standard SFT             & 47 / 182 & 25.82 & [20.01, 32.64] & 22.28
% EVEWorld                 & 23 / 182 & 12.64 & [8.57, 18.25] & 22.28

\subsection{Optional VLM Evaluation}
\label{sec:vlm_mlr_audit}

For settings where a capable multimodal model is available, VLM-based temporal
reasoning can provide an optional review of detector-based MLR events. This review
is auxiliary: it neither replaces nor modifies the frozen detector-based MLR protocol,
nor does it change the MLR labels or values reported in this paper. Because VLMs
differ in visual and temporal reasoning capability, their judgments are treated as
supporting evidence rather than ground truth, with manual review for ambiguous cases.
Our qualitative distortion analysis uses the same framework with additional manual
verification; examples are provided in Appendix~\ref{app:qual_distortion}.

\paragraph{VLM Inputs and Outputs.}
When reviewing the same evaluation set, the eligible set $\mathcal{D}^{+}$,
target query $q(c)$, and initial target count $N_0$ are inherited from the frozen
primary MLR protocol. The VLM can be provided with the conditioning frame,
instruction, target query, $N_0$, frames aligned with the MLR sampling timeline,
and a denser temporal view of the rollout. A JSON schema can constrain the output,
including event type, interval, supporting frames, visibility explanation, and
confidence.

\paragraph{MLR Judgment.}
For each candidate event, the VLM reviews the temporal evidence and returns the
structured judgment above. Duplication and disappearance are mapped back to the
fixed MLR sampling timeline and counted only when they satisfy the original
persistence rule. Benign visibility loss, including supported robot or non-robot
occlusion, temporary out-of-view motion, cropping, or other temporally supported
visibility changes, is flagged as potential false positives of the primary
detector-based MLR protocol and reported separately; they do not alter the primary
MLR values. Distortion is excluded from MLR because the target count is preserved,
but retained as a separate cross-frame consistency failure. Uncertain cases can be
passed to manual review, with repeated judgments or a refutation query optionally
used to check decision stability.

\FloatBarrier

\section{Method Details}
% [中文] 方法细节
\label{app:method_details}
\subsection{IGR Construction}
\begin{figure}[htbp]
  \centering
  \safeincludegraphics[width=\textwidth]{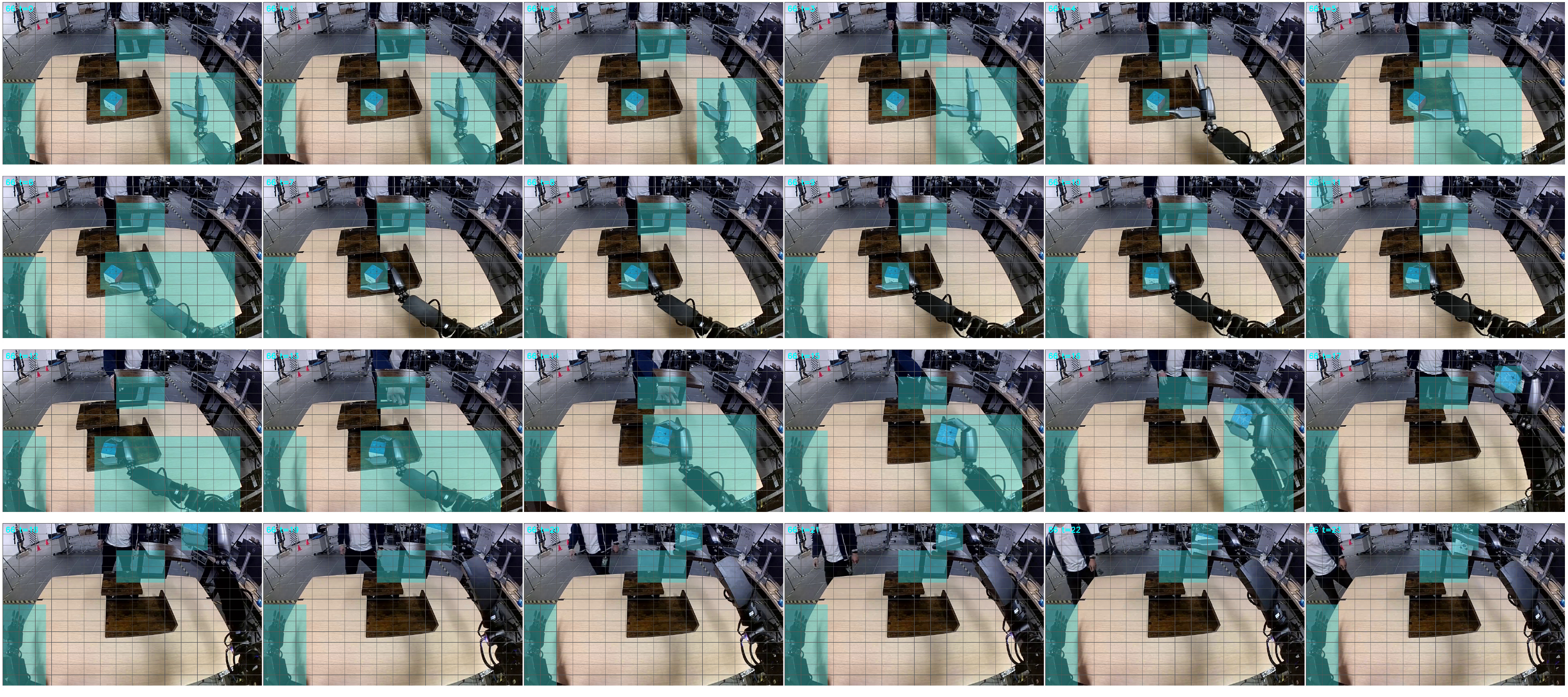}
\caption{
\textbf{Regional weighting in IGR.}
The final v4 restoration region and the actual pasted-instance region receive three times the reconstruction weight of ordinary regions before unit-mean normalization.
}
  \label{fig:igr_pipeline}
\end{figure}

% [中文] IGR 样本构建
\label{app:parsing}
\label{app:igr_details}
IGR constructs a corrupted--clean training pair from each valid target trajectory. An offline language parser extracts the target, source, and destination from the instruction. GroundingDINO \citep{liu2024groundingdino} localizes the target in sampled frames, and temporal association links valid detections into a trajectory. IGR then extracts an instance patch $p$ and pastes one additional target instance \citep{ghiasi2021copypaste} at a location $\Omega$ that stays inside the frame, does not overlap the tracked target, and passes the occupancy and collision gates. If no valid paste region exists, we use the clean video as both input and target. With $x$ the clean video, the corrupted input and the restoration target are
% [中文] IGR 从每条有效目标轨迹构建一个受扰--干净的训练对。离线语言解析器从指令中抽取目标、源位置和目的位置。GroundingDINO \citep{liu2024groundingdino} 在采样帧中定位目标，时间关联将有效检测连接为一条轨迹。随后 IGR 抽取一个实例块 $p$，并在位置 $\Omega$ 粘贴一个额外的目标实例 \citep{ghiasi2021copypaste}，该位置须保持在画面内、不与被跟踪目标重叠，并通过占用和碰撞门控。若不存在有效粘贴区域，则将干净视频同时作为输入和目标。记 $x$ 为干净视频，受扰输入与恢复目标为
\begin{equation}
\tilde{x}=\mathrm{Paste}(x,p,\Omega),\qquad
z=\mathrm{VAE}(x),\qquad
\tilde{z}=\mathrm{VAE}(\tilde{x}).
\label{eq:app_igr_pair}
\end{equation}
The model receives $\tilde z$ and reconstructs $z$. We use the final v4 restoration-region annotation shown in Figure~\ref{fig:igr_pipeline} to define a binary spatial emphasis mask $M_{\mathrm{rest}}$. The mask covers the target trajectory and task-relevant interaction regions around the source, destination, robot arm, and gripper, and additionally includes the actual pasted-instance region. Let
\begin{equation}
M(p)
=
\mathbb{I}\!\left[
p\in
\mathcal{R}_{\mathrm{v4}}
\cup
\mathcal{R}_{\mathrm{paste}}
\right],
\label{eq:igr_mask}
\end{equation}
where $p$ indexes a spatiotemporal latent location. Before normalization, we define
\begin{equation}
W(p)=1+2M(p),
\label{eq:igr_weight_raw}
\end{equation}
so that restoration-related locations receive three times the reconstruction weight of ordinary locations. We then normalize the map to unit mean,
\begin{equation}
\hat W(p)
=
\frac{W(p)}
{\frac{1}{|\Omega_{\mathrm{lat}}|}
\sum_{q\in\Omega_{\mathrm{lat}}}W(q)},
\label{eq:igr_weight_norm}
\end{equation}
where $\Omega_{\mathrm{lat}}$ denotes all spatiotemporal locations in the latent grid. This normalization preserves the $3{:}1$ relative weighting while keeping the global loss scale unchanged.
% [中文] 模型接收 $\tilde z$ 并重建 $z$。由于粘贴实例只占画面的一小部分，$\hat W$ 对其重建误差加权，并归一化为单位均值。

For DreamGen training, we propose copy--paste corruption with probability $p_{\mathrm{dup}}=0.5$. A proposal is applied only when the target trajectory and paste region pass the localization, occupancy, and collision gates; otherwise the sample falls back to the clean input. Consequently, the realized corrupted-sample fraction is lower than the proposal probability and is determined jointly by $p_{\mathrm{dup}}$ and the feasibility gates.
% [中文] 扰动提议概率与可行性门控共同决定实际扰动率。我们针对不同训练语料选择提议概率，在不放宽几何门控的前提下保留足够的有效粘贴区域。
\label{app:filtering}

\begin{algorithm}[H]
\caption{\textbf{Offline construction of an IGR supervision pair.} Invalid or unselected corruptions fall back to the clean training example.}
\label{alg:igr_construction}
\small
\textbf{Input:} clean video $x$, instruction $c$, corruption probability $p_{\mathrm{dup}}$, offline parser $P$, grounder $G$.\\
\textbf{Output:} corrupted input $\tilde{x}$, clean target $x$, and normalized regional weight map $\hat W$.
\begin{enumerate}[leftmargin=1.6em,itemsep=1pt,topsep=2pt]
  \item Parse $(\text{target},\text{source},\text{destination})\leftarrow P(c)$; return the clean example if the parse is invalid.
% [中文] 解析 $(\text{target},\text{source},\text{destination})\leftarrow P(c)$；若解析无效，则返回干净样本。
  \item Ground the target on sampled frames with $G$ and temporally associate valid detections into a trajectory $\mathcal{T}$; return the clean example if the confidence or continuity gate fails.
% [中文] 用 $G$ 在采样帧上定位目标，并将有效检测在时间维上关联为轨迹 $\mathcal{T}$；若置信度或连续性门控未通过，则返回干净样本。
  \item Draw the corruption decision with probability $p_{\mathrm{dup}}$; return the clean example if corruption is not selected, otherwise extract a target patch from a reliable trajectory observation.
% [中文] 以概率 $p_{\mathrm{dup}}$ 抽取扰动决策；若未选中扰动，则返回干净样本，否则从可靠的轨迹观测中抽取一个目标块。
  \item Enumerate paste regions that remain in-frame, do not overlap the tracked target, and satisfy the occupancy and collision gates; return the clean example if this set is empty.
% [中文] 枚举保持在画面内、不与被跟踪目标重叠且满足占用和碰撞门控的粘贴区域；若该集合为空，则返回干净样本。
  \item Sample a valid region $\Omega$, paste the additional instance into a copy of the video to obtain $\tilde{x}$, and retain the original video $x$ as the restoration target.
% [中文] 采样一个有效区域 $\Omega$，将额外实例粘贴到视频副本中得到 $\tilde{x}$，并保留原视频 $x$ 作为恢复目标。
  \item Assign weight 3 to the corrupted interaction region and weight 1 elsewhere, then normalize the resulting map to unit mean to obtain $\hat W$.
% [中文] 由受扰交互区域构建 $\hat W$，并按样本归一化，使空间加权不改变全局损失尺度。
\end{enumerate}
\end{algorithm}

\begin{table}[htbp]
\centering
\caption{
\textbf{Comparison of spatial-weight designs for IGR.}
We vary both the spatial support of restoration supervision and its relative
weighting strength.
Legacy uses the original five-level weighting, while the remaining variants
use binary weighting.
Path+Loc restricts supervision to the target trajectory and source/destination
regions; Interaction additionally covers robot-arm, gripper, and other
manipulation-relevant regions.
Coverage denotes the fraction of spatiotemporal latent locations assigned an
emphasized restoration weight.
MLR and IF are both reported in percentages, where lower MLR and higher IF are
better.
The Development columns compare candidate designs under a shared internal
development protocol and are not directly comparable to the final results
reported in the main text.
}
\label{tab:weightmap_design}
\setlength{\tabcolsep}{4.4pt}
\renewcommand{\arraystretch}{1.06}

\adjustbox{max width=\textwidth}{%
\begin{tabular}{lcccccccc}
\toprule
&
\multicolumn{3}{c}{Spatial Support}
&
\multicolumn{2}{c}{Relative Weight}
&
\multicolumn{1}{c}{Coverage}
&
\multicolumn{2}{c}{Development}
\\
\cmidrule(lr){2-4}
\cmidrule(lr){5-6}
\cmidrule(lr){7-7}
\cmidrule(lr){8-9}

Design
& Path/Loc.
& Arm/Grip.
& Paste
& Base
& Emph.
& Median
& MLR $\downarrow$
& IF $\uparrow$
\\
\midrule

Legacy multi-level
& \checkmark
& \checkmark
& \checkmark
& 0.5
& $2/3/4/6$
& 0.2679
& 2.18
& 60.31
\\

Binary-3
& \checkmark
& \checkmark
& \checkmark
& 1
& 3
& 0.2685
& 2.21
& 60.26
\\

Path+Loc-2x
& \checkmark
&
& \checkmark
& 1
& 2
& 0.2380
& 3.34
& 59.82
\\

Path+Loc-3x
& \checkmark
&
& \checkmark
& 1
& 3
& 0.2147
& 3.08
& 59.96
\\

Interaction-2x
& \checkmark
& \checkmark
& \checkmark
& 1
& 2
& 0.5361
& 2.06
& 60.42
\\

\rowcolor{gray!20}
Interaction-3x
& \checkmark
& \checkmark
& \checkmark
& 1
& 3
& 0.5361
& \textbf{1.94}
& \textbf{60.57}
\\

\bottomrule
\end{tabular}%
}
\end{table}

\paragraph{Spatial-Weight Design Selection.}
We compared several spatial-emphasis designs before fixing the final IGR weighting. First, we tested a multi-level map that assigned different relative weights to the target path, destination, vacated foreground, gripper, distractors, and pasted instance, using levels of $0.5/2/3/4/6$. We also tested a simpler $3\times$ mask restricted to the target path and locations. The latter was more sensitive to incomplete localization because restoration errors are not confined to these two geometric regions: they can also occur near the gripper, along the robot-arm motion corridor, and around the source and destination regions.

We therefore refined a broader binary restoration mask while keeping a single emphasis factor. The v1 annotation covers the arm--object interaction region broadly; v2 removes irrelevant background-person regions and tightens the foreground support; and v4 corrects the remaining missed arm regions identified during annotation inspection. Table~\ref{tab:weightmap_design} reports the resulting coverage statistics. In our development comparisons, the multi-level weighting and the binary $3\times$ design yielded comparable average performance. We therefore adopt the simpler v4 binary weighting in the final model: ordinary locations receive relative weight 1, restoration-related locations receive relative weight 3, and the actual pasted-instance region uses the same relative weight 3. This design avoids excessively large local weights while retaining coverage of the regions in which duplicate-restoration errors can arise.

\subsection{TIA Alignment}
% [中文] TIA 跨帧对齐与层选择
\label{app:tia_details}

TIA consists of local cross-frame alignment and feature transport at a selected backbone layer $\ell^\star$. We first describe the alignment mechanism and then the offline consistency probe used to select $\ell^\star$ across different domains and backbones. Throughout this section, $f^{-}_t=\mathrm{norm}(W_{\mathrm{in}}H_t^{\ell^\star})$ denotes the projected and $\ell_2$-normalized feature used for locating and transport, and $f^{+}_t$ the $\ell_2$-normalized refined feature after transport, written $\bar f^{\ell^\star}$ in the main text. The two stages use different windows: transport scores are computed over the dense window $\mathcal{C}^{\mathrm{tr}}_t$ centered at the same coordinate for every cell, whereas the consistency loss is supervised over the window $\mathcal{C}_t$ centered at the true next-frame location $u_t$.

\paragraph{Cross-frame alignment and feature transport.}
At the selected layer $\ell^\star$, TIA establishes local target consistency between each pair of adjacent frames \citep{tang2023dift,zhang2023twofeatures,oquab2024dinov2,doersch2023tapir,karaev2024cotracker}. A shared linear layer projects features from both frames into a normalized $d$-dimensional space. The feature at the previous target location $u_{t-1}$ then queries the local window $\mathcal{C}^{\mathrm{tr}}_t$ centered at the same coordinate in frame $t$, which yields the pre-transport similarity:
% [中文] 在所选层 $\ell^\star$，TIA 在每一对相邻帧之间建立局部目标对应关系。一个共享线性层将两帧特征投影到归一化的 $d$ 维空间，随后以前一帧目标位置 $u_{t-1}$ 处的特征查询帧 $t$ 中以相同坐标为中心的局部窗口 $\mathcal{C}^{\mathrm{tr}}_t$，得到搬运前的相似度：
\begin{equation}
s^{-}_{t,v}
=
f^{-}_{t-1}(u_{t-1})^\top
f^{-}_t(v).
\label{eq:app_tia_match}
\end{equation}

For feature transport, we apply the local matcher densely. Each cell in frame $t{-}1$ queries a local window centered at the same coordinate in frame $t$; a softmax with temperature $\tau$ converts the similarities into transport weights \citep{oord2018cpc}. The projected frame-$(t{-}1)$ features are splatted into frame $t$ using these weights, normalized per cell by the accumulated mass, and written back through a bounded residual update:
% [中文] 在特征搬运中，我们稠密应用该局部匹配器。帧 $t{-}1$ 的每个单元查询帧 $t$ 中以相同坐标为中心的局部窗口；相似度经过温度为 $\tau$ 的 softmax 后得到搬运权重。投影后的帧 $(t{-}1)$ 特征按权重搬运到帧 $t$，按单元以累积质量归一化，并通过有界残差更新写回：
\begin{equation}
H_t'
=
H_t^{\ell^\star}
+
\gamma\,
\tanh\!\big(
W_{\mathrm{out}}
\big(
\bar H_t-W_{\mathrm{in}}H_t^{\ell^\star}
\big)
\big),
\label{eq:app_tia_transport}
\end{equation}
where $\bar H_t$ is the transported feature field, $\gamma$ controls the residual magnitude, and $W_{\mathrm{out}}$ is a bias-free linear layer initialized to zero. The first frame is left unchanged. The consistency loss is computed on the refined features after this update: we re-normalize $H_t'$ to obtain $f^{+}_t=\mathrm{norm}(H_t')$, recompute the similarities $s^{+}_{t,v}=f^{+}_{t-1}(u_{t-1})^\top f^{+}_t(v)$ over the window $\mathcal{C}_t$ centered at the true next-frame location $u_t$, treat $u_t$ as the positive target and the remaining candidates as negatives, and optimize these correspondences with the temperature-scaled consistency loss $\Lcic$ defined in the main text. The transport weights and the consistency loss share the temperature $\tau$ but are computed in separate passes.
% [中文] 其中 $\bar H_t$ 为搬运后的特征场，$\gamma$ 控制残差幅度，$W_{\mathrm{out}}$ 为无偏置且零初始化的线性层。第一帧保持不变。一致性损失在本次更新之后的精修特征上计算：先对 $H_t'$ 重新归一化得到 $f^{+}_t=\mathrm{norm}(H_t')$，再在以真实下一帧位置 $u_t$ 为中心的窗口 $\mathcal{C}_t$ 内重算相似度 $s^{+}_{t,v}=f^{+}_{t-1}(u_{t-1})^\top f^{+}_t(v)$，以 $u_t$ 为正样本、其余候选为负样本，并通过正文定义的温度缩放一致性损失 $\Lcic$ 进行优化。搬运权重与一致性损失共用温度 $\tau$，但分两次单独计算。

\begin{algorithm}[H]
\caption{\textbf{TIA training-time consistency and feature transport.}}
\label{alg:tia_transport}
\small
\textbf{Input:} adjacent-frame hidden features $(H_{t-1}^{\ell^\star},H_t^{\ell^\star})$, target locations $(u_{t-1},u_t)$, shared down-projection $W_{\mathrm{in}}$, zero-initialized output projection $W_{\mathrm{out}}$, transport window $\mathcal{C}^{\mathrm{tr}}_t$, supervision window $\mathcal{C}_t$ centered at $u_t$, temperature $\tau$, residual scale $\gamma$.\\
\textbf{Output:} updated current-frame features $H_t'$, consistency loss $\Lcic$.
\begin{enumerate}[leftmargin=1.6em,itemsep=1pt,topsep=2pt]
  \item Project both frames with $W_{\mathrm{in}}$ and normalize the resulting features to obtain $f^{-}$. Each cell in frame $t{-}1$ forms a query, and the local window around the corresponding coordinate in frame $t$ forms its candidate set $\mathcal{C}^{\mathrm{tr}}_t$.

  \item Compute the pre-transport similarities $s^{-}_{t,v}$ within each local window and convert them into transport weights using a softmax with temperature $\tau$.

  \item Splat the projected frame-$(t{-}1)$ features into frame $t$ according to these weights, normalize each destination cell by its accumulated mass, and apply the residual update in Equation~\eqref{eq:app_tia_transport}. The first frame is left unchanged.

  \item Normalize the updated features $H_t'$ to obtain $f^{+}$, recompute the similarities $s^{+}_{t,v}$ within the supervision window $\mathcal{C}_t$, treat $u_t$ as the positive target and the remaining candidates as negatives, and compute the temperature-scaled consistency loss $\Lcic$.

  \item Back-propagate $\Lcic$ through the TIA sidecar and the trainable backbone.
\end{enumerate}
\end{algorithm}

\paragraph{Layer selection via consistency probing.}
\phantomsection
\label{app:tia_layer_selection}
We select the TIA attachment layer using an offline consistency probe rather than fixing it heuristically. For each candidate layer, the target feature in the previous frame retrieves its nearest match over the full token grid of the next frame, using the raw $\ell_2$-normalized block features rather than the $W_{\mathrm{in}}$-projected features used inside TIA, and endpoint error (EPE) \citep{teed2020raft} is measured against the localized target position. Because feature reliability can differ across domains and backbone architectures, we additionally report a setting-specific consistency-reliability measure. Table~\ref{tab:tia_layer_probe} summarizes the probe results and the resulting layer choices for GigaWorld-0/DreamGen, AgiBot, and FlowWAM.
% [中文] 我们通过离线一致性探针选择 TIA 的接入层，而不是启发式地固定某一层。对于每个候选层，以前一帧目标特征在下一帧的完整 token 网格中检索最近匹配（使用原始且 $\ell_2$ 归一化的 block 特征，而非 TIA 内部使用的 $W_{\mathrm{in}}$ 投影特征），并相对定位到的目标位置计算端点误差（EPE）。考虑到不同领域和主干架构中的特征可靠性可能不同，我们同时报告与设置对应的一致性可靠度指标。表~\ref{tab:tia_layer_probe} 汇总 GigaWorld-0/DreamGen、AgiBot 和 FlowWAM 上的探针结果及最终层选择。

\begin{table}[htbp]
\centering
\caption{
\textbf{Layer-wise consistency probe} across domains and backbones.
EPE denotes the mean endpoint error of adjacent-frame target matching in token cells.
Reliab. is the setting-specific consistency-reliability score:
object--background separation for DreamGen, static-feature stability for AgiBot,
or moving-pair top-1 hit rate for FlowWAM.
Block 23 is selected for GigaWorld-0/DreamGen and AgiBot, and block 12 is selected for FlowWAM.
}
\label{tab:tia_layer_probe}
\setlength{\tabcolsep}{14.0pt}

\adjustbox{max width=\textwidth}{%
\begin{tabular}{ccccccc}
\toprule
& \multicolumn{2}{c}{GigaWorld-0 / DreamGen}
& \multicolumn{2}{c}{AgiBot}
& \multicolumn{2}{c}{FlowWAM} \\
\cmidrule(lr){2-3}
\cmidrule(lr){4-5}
\cmidrule(lr){6-7}
Block
& EPE $\downarrow$
& Reliab. $\uparrow$
& EPE $\downarrow$
& Reliab. $\uparrow$
& EPE $\downarrow$
& Reliab. $\uparrow$ \\
\midrule
8  & 0.80 & $-$0.09 & 9.24 & 0.72 & 3.50 & 0.172 \\
9  & 0.80 & $-$0.07 & 8.88 & 0.74 & 3.56 & 0.171 \\
10 & 0.93 & $-$0.09 & 8.00 & 0.73 & 3.46 & 0.178 \\
11 & 0.93 & $-$0.09 & 7.52 & 0.74 & 3.45 & 0.176 \\
12 & 0.98 & $-$0.08 & 6.82 & 0.74
& \cellcolor{gray!20}3.42
& \cellcolor{gray!20}0.182 \\
13 & 0.68 & $-$0.07 & 6.56 & 0.75 & 3.46 & 0.177 \\
14 & 0.98 & $-$0.05 & 6.48 & 0.78 & 3.51 & 0.174 \\
15 & 0.98 & $-$0.10 & 6.39 & 0.81 & 3.49 & 0.172 \\
16 & 0.73 & $-$0.06 & 6.44 & 0.81 & 3.59 & 0.165 \\
17 & 0.78 & $-$0.06 & 6.62 & 0.85 & 3.64 & 0.158 \\
18 & 0.98 & $-$0.07 & 6.54 & 0.83 & 3.51 & 0.158 \\
19 & 0.91 & $-$0.07 & 6.64 & 0.83 & 3.56 & 0.155 \\
20 & 0.78 & $-$0.07 & 6.22 & 0.83 & 3.60 & 0.154 \\
21 & 0.75 & $-$0.10 & 6.44 & 0.80 & 3.65 & 0.146 \\
22 & 0.63 & $-$0.16 & 6.98 & 0.83 & 3.69 & 0.135 \\
23 & \cellcolor{gray!20}0.55
& \cellcolor{gray!20}$-$0.10
& \cellcolor{gray!20}5.72
& \cellcolor{gray!20}0.76
& 3.68 & 0.144 \\
24 & 1.08 & $-$0.19 & 9.81 & 0.81 & 3.71 & 0.139 \\
25 & 0.96 & $-$0.20 & 7.53 & 0.71 & 3.89 & 0.126 \\
26 & 1.28 & $-$0.24 & 9.79 & 0.65 & 3.87 & 0.124 \\
\bottomrule
\end{tabular}
}
\end{table}

\FloatBarrier

\subsection{Joint Training and Inference}
% [中文] 训练目标
\label{app:objective}
The full training objective combines the IGR restoration loss with the TIA consistency loss:
% [中文] 完整的训练目标将 IGR 恢复损失与 TIA 对应损失相结合：
\begin{equation}
\mathcal{L}=\Lpipe+\lambda_{\mathrm{TIA}}\,\Lcic .
\label{eq:app_objective}
\end{equation}

For samples with an applied IGR corruption, $\Lpipe$ restores the clean latent from the count-perturbed input; otherwise, it reduces to the base clean-video training objective. The consistency term is computed on the post-transport features. Its weight and optional noise gate are set per backbone in Appendix~\ref{app:training_setup}.
% [中文] 对于实际施加 IGR 扰动的样本，$\Lpipe$ 从数量扰动输入中恢复干净潜变量；否则，它退化为基础的干净视频训练目标。对应损失在搬运后的特征上计算，其权重和可选噪声门控按骨干设置，详见附录~\ref{app:training_setup}。

% [中文] 对应损失在搬运后的特征上计算，其权重和可选噪声门控按骨干设置，详见附录~\ref{app:training_setup}。
\label{app:inference_protocol}
At inference, the model receives the first frame, the instruction, and the benchmark-specific control input; the parser, the grounder, and the corruption generator are training-time only.
% [中文] 推理时，模型接收第一帧、指令和该基准特定的控制输入；解析器、定位器和扰动生成器仅在训练阶段使用。

\FloatBarrier
\section{Additional Experiments and Analysis}
% [中文] 实验设置与补充结果
\label{app:extended}
\label{app:mechanistic_ablation}
\subsection{Implementation Details}
% [中文] 实现细节
\label{app:training_setup}
Within each evaluation arm, \method{} and its control use the same initialization, training set, and optimization budget. The DreamGenBench and EWMBench experiments initialize from GigaWorld-0 Video-Pretrain-2B and update the full transformer with CAME-8bit \citep{luo2023came}, a learning rate of $4.32\times10^{-5}$, and an effective batch size of 64. For DreamGenBench, we train on 92 GR1 videos for 250 steps at $480\times768$ resolution using 93-frame clips at 16 FPS, and evaluate the raw step-250 checkpoint. For EWMBench, we train on 777 AgiBot clips for 50 steps at $480\times640$ resolution with the same frame count and frame rate, and evaluate the raw step-50 checkpoint.
% [中文] 在每个评估组内，\method{} 与其对照使用相同的初始化、训练集和优化预算。DreamGenBench 与 EWMBench 实验从 GigaWorld-0 Video-Pretrain-2B 初始化，并使用 CAME-8bit \citep{luo2023came} 更新全部 Transformer 参数，学习率为 $4.32\times10^{-5}$，有效批量大小为 64。在 DreamGenBench 上，我们在 92 个 GR1 视频上以 $480\times768$ 分辨率、16 FPS 的 93 帧片段训练 250 步，并评估原始的第 250 步检查点。在 EWMBench 上，我们在 777 个 AgiBot 片段上以 $480\times640$ 分辨率、相同帧数与帧率训练 50 步，并评估原始的第 50 步检查点。

For RoboTwin, both models initialize from the released FlowWAM Stage-1 checkpoint and train a rank-32 LoRA \citep{hu2022lora} on 2,250 episodes for 1,128 optimization steps. We use AdamW \citep{loshchilov2019adamw} with a learning rate of $10^{-4}$, weight decay 0.01, and batch size 8. Training clips contain 29 frames at 24 FPS with a resolution of $480\times640$; evaluation uses the final-epoch checkpoint.
% [中文] 在 RoboTwin 上，两个模型都从已发布的 FlowWAM Stage-1 检查点初始化，在 2,250 个片段上训练一个秩为 32 的 LoRA \citep{hu2022lora}，共四个 epoch（1,128 步）。我们使用 AdamW \citep{loshchilov2019adamw}，学习率为 $10^{-4}$，权重衰减为 0.01，批量大小为 8。训练片段为 24 FPS 下的 29 帧，分辨率为 $480\times640$；评估使用最后一个 epoch 的检查点。

In the evaluated instantiations, TIA uses a 64-dimensional projection, $7\times7$ local windows, temperature $\tau=0.07$, and residual scale $\gamma=0.1$. It attaches to block 23 on GigaWorld-0 and AgiBot, and block 12 on FlowWAM. On GigaWorld-0, $\lambda_{\mathrm{TIA}}$ warms from 0 to 0.5 over the first 20 steps and applies at $\sigma\in[0.2,0.5]$; FlowWAM uses $\lambda_{\mathrm{TIA}}=0.1$ without noise gating.
% [中文] 在所评估的实例化中，TIA 使用 64 维投影、$7\times7$ 局部窗口、温度 $\tau=0.07$ 和残差系数 $\gamma=0.1$。DreamGen 使用第 22--23 块，FlowWAM 使用第 12 块。在 GigaWorld-0 上，$\lambda_{\mathrm{TIA}}$ 在前 20 步从 0 增至 0.5，并作用于 $\sigma\in[0.2,0.5]$；FlowWAM 使用 $\lambda_{\mathrm{TIA}}=0.1$，不设噪声门控。

\label{app:inference_setup}
DreamGenBench and WorldArena use the raw step-250 checkpoint, 30 denoising steps, and CFG 7.0 at $480\times768$ resolution with 93 frames at 16 FPS. EWMBench uses the raw step-50 checkpoint, the three benchmark seeds, 30 steps, and CFG 7.0 at $480\times640$ with the same clip length and frame rate. RoboTwin uses the final-epoch checkpoint and CFG 5.0 at $480\times640$, generating the full trajectory at 24 FPS. All settings condition on the first frame and instruction; RoboTwin additionally uses robot-only optical flow. Appendix~\ref{app:training_sampling_sensitivity} studies CFG and denoising choices separately.
% [中文] DreamGenBench 和 WorldArena 使用原始第 250 步检查点、种子 004、30 个去噪步骤和 CFG 7.0，以 $480\times768$ 分辨率生成 16 FPS 的 93 帧视频。EWMBench 使用原始第 50 步检查点、三个基准种子、30 个去噪步骤和 CFG 7.0，以 $480\times640$ 分辨率生成相同帧数与帧率的视频。RoboTwin 使用最后一个 epoch 的检查点、种子 42、40 个去噪步骤和 CFG 5.0，以 $480\times640$ 分辨率生成 24 FPS 的完整轨迹。所有设置均以第一帧和指令为条件，RoboTwin 额外使用仅机器人的光流。附录~\ref{app:training_sampling_sensitivity} 单独研究 CFG 和去噪设置。

\subsection{IGR Ablation and Duplicate Restoration}
% [中文] IGR 消融与复原诊断
\label{app:igr_variants}
\label{app:probes}
\label{app:mechanism_probes}
\label{app:retention_check}

\paragraph{Restoration-objective ablation.}
Table~\ref{tab:igr_internal} isolates the effects of duplicate-restoration supervision and spatial weighting. Removing spatial emphasis from the restoration objective still reduces MLR relative to Standard SFT, but leaves substantially more violations than the complete \method{}. Conversely, applying spatial emphasis without the duplicate-restoration target does not improve instance consistency. IGR alone reduces MLR to 4.76\%, while the complete \method{} further reduces it to 1.59\%. Notably, removing spatial emphasis yields a slightly higher Overall instruction-following score than the complete model (61.11\% vs.\ 60.85\%) but increases MLR from 1.59\% to 6.35\%, supporting the use of the complete objective.
% [中文] 表~\ref{tab:igr_internal} 分离重复实例复原监督与空间加权的作用。去除空间强调后仍可降低 MLR，但明显弱于完整方法；仅施加空间强调而没有重复实例复原目标则无法改善实例一致性。单独使用 IGR 可将 MLR 降至 4.76\%，完整方法进一步降至 1.59\%。值得注意的是，去除空间强调后 Overall 略高，但 MLR 从 1.59\% 上升至 6.35\%，因此我们采用完整目标。

\begin{table}[htbp]
\centering
\caption{
\textbf{IGR restoration controls} on DreamGenBench.
The first two variants isolate spatial weighting and duplicate-restoration supervision.
IGR w/o spatial emphasis removes the spatial weighting, and Spatial emphasis only applies the weighting without the duplicate-restoration target.
Env, Object, Behavior, and Overall report Gemini-IF instruction-following accuracy in percentages.
}
\label{tab:igr_internal}
\setlength{\tabcolsep}{5.5pt}
\begin{tabular}{lccccc}
\toprule
Variant & MLR (\%) $\downarrow$ & Env $\uparrow$ & Object $\uparrow$ & Behavior $\uparrow$ & Overall $\uparrow$ \\
\midrule
IGR w/o spatial emphasis & 6.35 & 56.32 & 44.00 & 82.27 & 61.11 \\
Spatial emphasis only & 14.29 & 34.48 & 34.67 & 62.41 & 44.97 \\
IGR only & 4.76 & 49.43 & 36.67 & 69.50 & 51.85 \\
\rowcolor{gray!20}
\method{} & 1.59 & 72.41 & 50.33 & 64.89 & 60.85 \\
\bottomrule
\end{tabular}
\end{table}

\begin{figure}[htbp]
  \centering
  \safeincludegraphics[width=\textwidth]{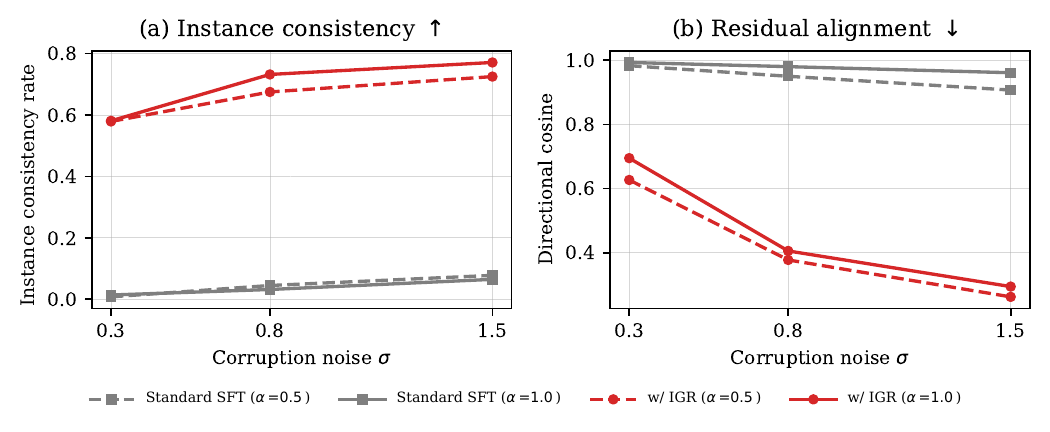}
\caption{
\textbf{Controlled restoration analysis of IGR.}
Injected duplicates probe restoration behavior under different corruption-noise levels $\sigma$ and paste opacities $\alpha$.
(a) Standard SFT largely retains the injected duplicate, whereas IGR substantially improves duplicate removal across corruption strengths.
(b) Residual errors under Standard SFT remain strongly aligned with the injected duplicate, while IGR reduces this alignment after restoration training.
}
\label{fig:mechanism}
\end{figure}

\paragraph{Controlled duplicate-restoration evaluation.}
We further test whether IGR learns to actively correct a corrupted target count rather than merely preserving the input appearance. We construct an independent corruption set by inserting one additional target instance into clean demonstrations and reconstruct the resulting videos under six combinations of paste opacity $\alpha$ and diffusion noise level $\sigma$. Starting from the Standard SFT checkpoint, the model preserves 92--99\% of the injected area, indicating that the spurious instance is largely retained. After 50 IGR fine-tuning steps, retention decreases to 23--42\% across all six settings (Table~\ref{tab:corruption_robustness}). The directional cosine decreases together with retention, showing that the residual reconstruction error becomes not only smaller but also less aligned with the injected duplicate. Figure~\ref{fig:mechanism} summarizes both effects across corruption strengths.
% [中文] 我们进一步通过受控实验检验 IGR 是否能够主动修正被扰乱的目标实例数量，而不仅仅是保持输入外观。我们在独立的受扰集合中向干净演示插入一个额外目标实例，并在六组粘贴不透明度 $\alpha$ 与扩散噪声 $\sigma$ 的组合下进行复原。从 Standard SFT 检查点出发，模型保留了 92--99\% 的注入区域；经过 50 步 IGR 微调后，该比例降至 23--42\%。方向余弦也随之降低，说明剩余误差不仅减小，而且不再高度对齐于注入的重复实例。

\begin{table}[htbp]
\centering
\caption{
\textbf{Robustness of duplicate removal} across corruption strengths.
Retention is the fraction of injected duplicate area preserved by the reconstruction.
Dir. cos. measures the alignment between the residual error and the injected duplicate.
Before and after denote the Standard SFT checkpoint and the model after 50 IGR fine-tuning steps, respectively.
}
\label{tab:corruption_robustness}
\setlength{\tabcolsep}{5.5pt}

\begin{tabular}{cccccc}
\toprule
Opacity $\alpha$
& $\sigma$
& Ret. before $\downarrow$
& Ret. after $\downarrow$
& Rel. reduction (\%) $\uparrow$
& Dir. cos. before / after \\
\midrule
0.5 & 0.3 & 0.992 & 0.421 & 57.5 & 0.983 / 0.627 \\
0.5 & 0.8 & 0.955 & 0.325 & 66.0 & 0.950 / 0.378 \\
0.5 & 1.5 & 0.922 & 0.275 & 70.2 & 0.907 / 0.263 \\
1.0 & 0.3 & 0.986 & 0.419 & 57.5 & 0.993 / 0.695 \\
1.0 & 0.8 & 0.968 & 0.268 & 72.3 & 0.980 / 0.406 \\
1.0 & 1.5 & 0.935 & 0.229 & 75.5 & 0.961 / 0.295 \\
\bottomrule
\end{tabular}
\end{table}

\paragraph{Feature diagnostics.}
Two additional diagnostics use the same feature-probe family as Appendix~\ref{app:tia_details}. Blocks 25--26 best separate changed from unchanged regions, with score gaps of 0.43--0.45, while blocks 10--16 provide the strongest detection of injected duplicates, with AUC values of 0.79--0.80.
% [中文] 另外两项诊断使用与附录~\ref{app:tia_details} 相同的特征探针族。第 25--26 块最能区分已变化与未变化区域，分数差距为 0.43--0.45；第 10--16 块对注入重复实例的检测最强，AUC 为 0.79--0.80。

\subsection{Evaluation Stability}
% [中文] 评估稳定性
\label{app:evaluation_stability}
\label{app:horizon_hardware}

We quantify variability from repeated instruction-following judgments and inference seeds. Table~\ref{tab:dreamgen_gemini_sd} reruns Gemini-IF on fixed videos and reports every repeat underlying each mean; the main ordering remains unchanged. Table~\ref{tab:matched_seed} further compares Standard SFT and \method{} under matched inference seeds, explicitly controlling for sampling stochasticity. The consistent advantage of \method{} indicates that the observed gains are not attributable to favorable seed selection.
% [中文] 我们量化重复指令遵循判定和推理种子带来的变异。表~\ref{tab:dreamgen_gemini_sd} 在固定视频上重复运行 Gemini-IF，并报告每个均值对应的全部重复，主要排序保持不变。表~\ref{tab:matched_seed} 在匹配的推理种子下比较 Standard SFT 与 \method{}；

\begin{table}[htbp]
\centering
\caption{
\textbf{Gemini-IF judge variance} across models on DreamGenBench.
Overall reports mean $\pm$ SD over repeated Gemini-IF evaluations of the same generated videos.
Env, Object, and Behavior report the corresponding mean scores on each split.
Repeats denotes the number of independent Gemini-IF evaluations.
}
\label{tab:dreamgen_gemini_sd}
\setlength{\tabcolsep}{3.3pt}
\begin{tabular}{lccccc}
\toprule
& & \multicolumn{4}{c}{Gemini-IF (\%) $\uparrow$} \\
\cmidrule(lr){3-6}
Model & Repeats & Overall & Env & Object & Behavior \\
\midrule
CogVideoX1.5-5B-I2V \citep{yang2025cogvideox} & 2 & 5.56 $\pm$ 1.12 & 15.52 & 1.00 & 4.26 \\
Wan2.2-TI2V-5B \citep{wan2025wan} & 2 & 10.32 $\pm$ 1.12 & 17.24 & 5.00 & 11.70 \\
Wan2.2-I2V-A14B \citep{wan2025wan} & 2 & 15.87 $\pm$ 3.37 & 20.69 & 13.00 & 15.96 \\
Cosmos-Predict2-2B \citep{nvidia2025cosmos} & 2 & 24.60 $\pm$ 1.12 & 31.03 & 28.00 & 17.02 \\
GigaWorld-0 \citep{gigaworld2025zero} & 6 & 60.19 $\pm$ 2.90 & 54.60 & 54.33 & 69.86 \\
Official GR1 Fine-tune \citep{jang2025dreamgen} & 2 & 63.10 $\pm$ 6.17 & 55.17 & 46.00 & 86.17 \\
Standard SFT & 6 & 53.57 $\pm$ 2.92 & 51.72 & 42.00 & 67.02 \\
\rowcolor{gray!20}
EVEWorld & 6 & 60.85 $\pm$ 1.79 & 72.41 & 50.33 & 64.89 \\
\bottomrule
\end{tabular}
\end{table}

\begin{table}[htbp]
\centering
\caption{
\textbf{Matched inference-seed comparison} of Standard SFT and \method{} on DreamGenBench.
Gemini-IF reports mean $\pm$ SD over repeated judge evaluations at each inference seed.
$\Delta$ denotes the improvement of \method{} over Standard SFT in percentage points.
}
\label{tab:matched_seed}
\setlength{\tabcolsep}{14pt}

\begin{tabular}{lcc>{\columncolor{gray!20}}cc}
\toprule
Seed & Judge repeats & Standard SFT (\%) & \method{} (\%) & $\Delta$ (pp) \\
\midrule
004 & 6 & 53.57 $\pm$ 2.92 & 60.85 $\pm$ 1.79 & $+$7.28 \\
061 & 3 & 57.41 $\pm$ 2.79 & 58.73 $\pm$ 1.37 & $+$1.32 \\
\bottomrule
\end{tabular}
\end{table}

Under the transfer protocol, five complete FlowWAM generations from the same
checkpoint differ by at most 0.68 Local-11 points
(Table~\ref{tab:seed_variance_full}), indicating stable overall performance
across inference seeds.
We further examine per-component variation across five inference seeds on a
50-prompt probe set (Table~\ref{tab:seed_variance_components}).
Motion-related metrics show larger variation than appearance-related metrics,
but the aggregate Local-11 score remains relatively stable.
Local-11 denotes the project-local 11-metric evaluator.
% [中文] 在迁移协议下，同一检查点的五次完整 FlowWAM 生成在 Local-11 上最多相差 0.68 分（表~\ref{tab:seed_variance}，上）。五个探针种子下的运动相关分项比外观分项波动更大（下）。Local-11 表示项目内部的 11 指标评估器。

\begin{table}[htbp]
\centering
\caption{
\textbf{Inference-seed variance} under the full-1000 transfer protocol.
Local-11 is evaluated from the same FlowWAM checkpoint across five inference seeds.
Mean $\pm$ SD reports variation across seeds.
}
\label{tab:seed_variance_full}
\setlength{\tabcolsep}{8.0pt}

\begin{tabular}{lcccccc}
\toprule
Seed & 6 & 12 & 20 & 42 & 70 & Mean $\pm$ SD \\
\midrule
Local-11
& 57.12
& 56.89
& 57.57
& 57.37
& 57.46
& 57.28 $\pm$ 0.28 \\
\bottomrule
\end{tabular}
\end{table}

\begin{table}[htbp]
\centering
\caption{
\textbf{Per-component inference-seed variance} under the transfer protocol.
Component scores are evaluated across five inference seeds on a 50-prompt probe set.
Mean $\pm$ SD reports variation across seeds.
}
\label{tab:seed_variance_components}
\setlength{\tabcolsep}{5.8pt}

\begin{tabular}{lcccccc}
\toprule
Component
& Seed 0
& Seed 17
& Seed 35
& Seed 42
& Seed 70
& Mean $\pm$ SD \\
\midrule
Image Quality
& 58.95 & 59.41 & 59.35 & 59.37 & 57.46
& 58.91 $\pm$ 0.83 \\

Aesthetic
& 44.30 & 44.18 & 44.56 & 44.38 & 44.35
& 44.35 $\pm$ 0.14 \\

Dynamic Degree
& 29.59 & 31.44 & 29.79 & 27.88 & 33.17
& 30.37 $\pm$ 2.01 \\

Flow Score
& 18.65 & 20.57 & 17.97 & 15.67 & 20.11
& 18.59 $\pm$ 1.95 \\

Motion Smoothness
& 68.64 & 68.79 & 68.14 & 68.15 & 71.28
& 69.00 $\pm$ 1.31 \\

Subject Consistency
& 73.84 & 76.67 & 73.22 & 80.08 & 75.56
& 75.87 $\pm$ 2.72 \\

Background Consistency
& 80.25 & 82.59 & 79.25 & 86.17 & 82.36
& 82.12 $\pm$ 2.67 \\

Photometric
& 24.71 & 28.02 & 24.61 & 30.75 & 24.45
& 26.51 $\pm$ 2.80 \\

Interaction
& 54.40 & 56.00 & 57.60 & 58.40 & 58.80
& 57.04 $\pm$ 1.82 \\

Perspectivity
& 87.60 & 86.00 & 89.20 & 88.00 & 85.20
& 87.20 $\pm$ 1.60 \\

Instruction Following
& 47.20 & 47.20 & 50.80 & 53.20 & 54.80
& 50.64 $\pm$ 3.45 \\

Local-11
& 53.47 & 54.63 & 54.05 & 55.64 & 55.23
& 54.60 $\pm$ 0.88 \\
\bottomrule
\end{tabular}
\end{table}

\subsection{Training and Long-Horizon Behavior}
% [中文] 训练与长时域行为
\label{app:training_sampling_sensitivity}

\paragraph{Training dynamics.}
We first examine how instance consistency and instruction following evolve during post-training. 
As training proceeds toward the selected step-250 checkpoint, MLR generally decreases while Qwen-IF improves, indicating that the reduction in Model Laziness is accompanied by better task fidelity rather than obtained at its expense.
% obtained at its expense (Figure~\ref{fig:laziness_step}).
% \begin{figure}[!htbp]
%   \centering
%   \safeincludegraphics[width=\textwidth]{figures/fig_mlr_vs_step.pdf}
% \caption{
% \textbf{Training progress} on DreamGenBench.
% (a) MLR decreases as training proceeds, indicating fewer instance consistency violations.
% (b) Qwen-IF tracks instruction-following performance over the same training process.
% }
% \label{fig:laziness_step}
% \end{figure}

\paragraph{Guidance sensitivity.}
We further evaluate classifier-free guidance (CFG) \citep{ho2022cfg} across multiple post-training checkpoints and guidance weights. Table~\ref{tab:cfg_main} shows that stronger guidance generally raises Gemini-IF, but MLR is substantially less monotonic. For example, increasing $w$ can improve instruction following while leaving MLR unchanged or even increasing it at the same checkpoint. Conversely, the lowest MLR values occur only for particular combinations of training stage and guidance weight rather than systematically at the strongest guidance setting. These results show that CFG primarily controls inference-time instruction fidelity, whereas target-instance consistency remains dependent on the learned post-training behavior.

\begin{table}[htbp]
\centering
\caption{
\textbf{CFG sensitivity} across training stages on DreamGenBench.
Each cell reports Gemini-IF $\uparrow$ / MLR $\downarrow$ in percentages under different CFG weights $w$ and training steps.
$w{=}1.0$ denotes inference without classifier-free guidance.
The shaded column marks the step-250 checkpoint used in the main experiments.
The shaded entry at step 250 and $w{=}7.0$ corresponds to the main EVEWorld configuration reported in main-text Table~\ref{tab:dreamgen_overall}.
}
\label{tab:cfg_main}
\setlength{\tabcolsep}{0.8pt}
\renewcommand{\arraystretch}{1.08}

\resizebox{\linewidth}{!}{%
\begin{tabular}{lcccc>{\columncolor{gray!20}}cc}
\toprule
CFG weight $w$ & \multicolumn{6}{c}{Train step} \\
\cmidrule(lr){2-7}
& 50 & 100 & 150 & 200 & 250 & 300 \\
\midrule

1.0
& 26.08 / 9.52
& 39.32 / 6.35
& 26.91 / 4.76
& 26.08 / 9.52
& 28.98 / 9.52
& 31.87 / 6.35 \\
2.5
& 45.95 / 4.76
& 54.23 / 11.11
& 61.68 / 4.76
& 47.19 / 3.17
& 50.91 / 4.76
& 55.06 / 4.76 \\
5.0
& 50.91 / 7.94
& 57.53 / 9.52
& 57.53 / 3.17
& 51.32 / 1.59
& 58.78 / 6.35
& 48.85 / 4.76 \\
7.0
& 52.15 / 4.76
& 66.23 / 9.52
& 60.43 / 6.35
& 53.40 / 1.59
& 60.85 / 1.59
& 45.95 / 3.17 \\

\bottomrule
\end{tabular}%
}
\end{table}

\paragraph{Long-horizon behavior.}
Finally, we examine whether longer rollouts introduce broader physical and process-level degradation.
On PBench \citep{nvidia2025pbench}, overall physical-QA performance decreases as the output duration grows from 3.8\,s to 19.8\,s, with temporal reasoning degrading most strongly (Table~\ref{tab:pbench_horizon}).
Averaged over all five durations, Standard SFT improves all four PBench dimensions over the pretrained backbone.
The advantage becomes more pronounced at longer horizons: over durations of at least 9.8\,s, it improves Domain, Phys, and Time by 1.65, 1.95, and 4.24 points on average, respectively.
These results indicate that post-training improves broader physical reasoning even as long-horizon generation remains increasingly challenging.

\begin{table}[htbp]
\centering
\caption{
\textbf{PBench physical-QA accuracy} across output durations.
Domain reports the overall PBench Robot score, judged by Qwen3.6-VL.
Phys, Space, and Time report accuracy on physical, spatial, and temporal reasoning.
Mean averages the five durations, and Mean ($\geq$9.8\,s) averages the three longest horizons.
}
\label{tab:pbench_horizon}
\setlength{\tabcolsep}{8.6pt}
\adjustbox{max width=\textwidth}{%
\begin{tabular}{lcccccccc}
\toprule
& \multicolumn{4}{c}{GigaWorld-0 (Pretrained)} & \multicolumn{4}{c}{Standard SFT} \\
\cmidrule(lr){2-5}
\cmidrule(lr){6-9}
Dur. (s) & Domain & Phys. & Space & Time & Domain & Phys. & Space & Time \\
\midrule
3.8 & 82.43 & 85.71 & 84.24 & 82.18 & 80.98 & 86.69 & 83.33 & 79.64 \\
5.8 & 79.87 & 84.42 & 82.12 & 74.18 & 79.89 & 85.71 & 83.94 & 74.91 \\
9.8 & 74.32 & 83.77 & 76.06 & 65.09 & 78.90 & 87.99 & 79.70 & 74.55 \\
15.8 & 67.54 & 79.22 & 73.03 & 49.82 & 68.12 & 79.55 & 70.61 & 53.82 \\
19.8 & 66.97 & 76.95 & 74.55 & 50.18 & 66.76 & 78.25 & 73.94 & 49.45 \\
\midrule
Mean & 74.23 & 82.01 & 78.00 & 64.29 & \textbf{74.93} & \textbf{83.64} & \textbf{78.30} & \textbf{66.47} \\
Mean ($\geq$9.8\,s) & 69.61 & 79.98 & 74.55 & 55.03 & \textbf{71.26} & \textbf{81.93} & \textbf{74.75} & \textbf{59.27} \\
\bottomrule
\end{tabular}
}
\end{table}

Longer horizons also expose failures that are not captured by target-instance count alone.
On 92 matched prompts, both GigaWorld-0 variants exhibit substantially higher shortcut severity, shortcut frequency, and missing-stage frequency at 7.8\,s than at 3.8\,s (Table~\ref{tab:cross_model_laziness}).
At the shared 3.8\,s horizon, Wan2.2-TI2V-5B exhibits still larger process errors than either GigaWorld-0 variant.
These diagnostics complement MLR by measuring broader failures of process completeness rather than instance consistency alone.

\begin{table}[htbp]
\centering
\caption{
\textbf{Long-horizon process evaluation} across models and output durations.
Qwen-based evaluation is performed on 92 matched prompts.
Severity measures shortcut severity.
Any shortcut and Missing stage report the fractions of generations containing a shortcut or a missing stage.
}
\label{tab:cross_model_laziness}
\setlength{\tabcolsep}{9.7pt}
\begin{tabular}{lcccc}
\toprule
Model / horizon & Severity $\downarrow$ & Any shortcut $\downarrow$ & Missing stage $\downarrow$ & $n$ \\
\midrule
GigaWorld-0 SFT, 3.8s & 0.94 & 0.32 & 0.30 & 92 \\
GigaWorld-0 SFT, 7.8s & 1.79 & 0.63 & 0.60 & 92 \\
GigaWorld-0 Pretrained, 3.8s & 1.17 & 0.39 & 0.38 & 92 \\
GigaWorld-0 Pretrained, 7.8s & 1.97 & 0.67 & 0.66 & 92 \\
Wan2.2-TI2V-5B, 3.8s & \textbf{2.30} & \textbf{0.82} & \textbf{0.72} & 92 \\
\bottomrule
\end{tabular}
\end{table}

\section{Exploration of Alternative Approaches}
% [中文] 备选设计
\label{app:failed_alternatives}
\label{app:negative_controls}

We evaluate five alternative designs spanning training to inference:
PhysicsLatent, EAG, LAD-LoRA, Causal Frontier, and ICH-D.
Because their training budgets and generation protocols differ, each route is compared with its own control using the same prompts, conditioning frames, and inference seeds. Table~\ref{tab:failed_routes} reports MLR and Gemini-IF, with 95\% confidence intervals computed by paired bootstrap over the matched pairs.

No alternative jointly reduces MLR and improves instruction following.
PhysicsLatent slightly reduces MLR but lowers Gemini-IF, EAG improves Gemini-IF but increases MLR, and the remaining routes show no joint gain.

\begin{table}[htbp]
\centering
\caption{
\textbf{Comparison with alternative designs.}
Each route is evaluated against its matched control using the same prompts, conditioning frames, and inference seeds.
Panel (a) reports MLR for each route and its matched control.
Panel (b) reports Gemini-IF as mean $\pm$ SD over three judge repeats.
Rel.\ $\Delta$ denotes the relative change from the matched control, and 95\% CIs are computed by paired bootstrap. MLR values are reported to one decimal place; Rel.\ $\Delta$ and the CIs are computed from unrounded values.
}
\label{tab:failed_routes}

\vspace{2pt}
\textbf{(a) Model Laziness Rate (MLR)}\\[2pt]

{\setlength{\tabcolsep}{7.0pt}
\begin{tabular}{llccccc}
\toprule
Route
& Matched control
& \multicolumn{2}{c}{MLR (\%) $\downarrow$}
& Rel.\ $\Delta$
& 95\% CI \\
\cmidrule(lr){3-4}
& & Control & Route & & \\
\midrule
PhysicsLatent
& Pretrain 5.8s
& 17.4
& 16.2
& $-$7.0
& [$-$51.5, $+$69.1] \\

EAG
& weight 0
& 15.4
& 38.5
& $+$150.0
& [0.0, $+$500.0] \\

LAD-LoRA
& LoRA control
& 3.8
& 3.8
& 0.0
& [0.0, 0.0] \\

Causal Frontier
& control50
& 30.8
& 41.7
& $+$35.4
& [$-$42.9, $+$300.0] \\

ICH-D
& Pretrain (4 seeds)
& 11.2
& 13.1
& $+$16.6
& [$-$19.4, $+$70.8] \\
\bottomrule
\end{tabular}}

\vspace{6pt}
\textbf{(b) Gemini-IF}\\[2pt]

{\setlength{\tabcolsep}{8.0pt}
\begin{tabular}{lcccc}
\toprule
Route
& Control
& Route
& Rel.\ $\Delta$
& 95\% CI \\
\midrule
PhysicsLatent
& 45.65 $\pm$ 3.92
& 40.22 $\pm$ 3.26
& $-$11.9
& [$-$29.8, $+$7.9] \\

EAG
& 41.67 $\pm$ 3.61
& 45.83 $\pm$ 9.55
& $+$10.0
& [$-$15.8, $+$54.5] \\

LAD-LoRA
& 51.04 $\pm$ 1.80
& 47.92 $\pm$ 4.77
& $-$6.1
& [$-$24.1, $+$13.3] \\

Causal Frontier
& 27.08 $\pm$ 3.61
& 12.50 $\pm$ 6.25
& $-$53.8
& [$-$100.0, $+$85.7] \\

ICH-D
& 45.83 $\pm$ 1.55
& 31.79 $\pm$ 0.47
& $-$30.6
& [$-$40.1, $-$20.2] \\
\bottomrule
\end{tabular}}
\end{table}

\subsection{PhysicsLatent}
\label{app:physicslatent}
PhysicsLatent augments the conditioning sequence with eight learned process tokens. The tokens are computed from the pooled reference-frame latent and prompt embedding and are read by every cross-attention layer. Weak-label heads predict the target's start and goal boxes, contact events, a coarse 2D trajectory, and the manipulation phase, using pseudo-labels parsed from the instruction together with color and motion heuristics. Against the pretrained 5.8\,s control (92 matched pairs), MLR moves from 17.4\% to 16.2\% while Gemini-IF drops from 45.65 to 40.22, leaving no joint gain.
% [中文] PhysicsLatent 用八个可学习的过程 token 扩充条件序列。这些 token 由池化后的参考帧 latent 和提示词嵌入计算得到，并被每个交叉注意力层读取。弱标签头预测目标的起点框与终点框、接触事件、一条粗略的 2D 轨迹以及操作阶段，所用伪标签由指令解析并结合颜色与运动启发式得到。与预训练 5.8\,s 对照（92 个匹配对）相比，MLR 从 17.4\% 变为 16.2\%，而 Gemini-IF 从 45.65 降至 40.22，未形成联合增益。

\begin{algorithm}[H]
\caption{\textbf{PhysicsLatent training-sample construction.} Pseudo-labels come from instruction parsing and color/motion heuristics; low-confidence labels are masked out.}
\label{alg:physlatent}
\small
\textbf{Input:} training video $x$, instruction $c$, frozen VAE and text encoder, process-token encoder $E_\psi$, weak-label heads $\{h_k\}$.\\
\textbf{Output:} process-token-augmented input and auxiliary training targets.
\begin{enumerate}[leftmargin=1.6em,itemsep=1pt,topsep=2pt]
  \item Encode the reference frame with the VAE and pool it to a $15{\times}24$ grid; embed the instruction with the text encoder.
% [中文] 用 VAE 编码参考帧并池化为 $15{\times}24$ 网格；用文本编码器嵌入指令。
  \item Produce $K{=}8$ process tokens (1024-dimensional) from both with $E_\psi$ and append them after the text tokens; every cross-attention layer reads them, and the two CFG branches share the same tokens.
% [中文] 用 $E_\psi$ 从两者生成 $K{=}8$ 个过程 token（1024 维），并附加在文本 token 之后；每个交叉注意力层都会读取它们，两个 CFG 分支共享同一组 token。
  \item Predict the weak labels from the process tokens: start and goal boxes, contact events, an 8-keyframe 2D trajectory, the manipulation phase, and completion flags.
% [中文] 从过程 token 预测弱标签：起点框与终点框、接触事件、8 关键帧的 2D 轨迹、操作阶段以及完成标志。
  \item Obtain pseudo-labels by parsing the instruction (target, source, destination) and applying color- and motion-based heuristics on the frames; mask out low-confidence labels.
% [中文] 通过解析指令（目标、源位置、目的位置）并在帧上应用基于颜色和运动的启发式获得伪标签；掩码掉低置信度标签。
  \item Train with $\mathcal{L}_{\mathrm{video}}+\textstyle\sum_k w_k\mathcal{L}_k$, where each $w_k$ is small ($0.03$--$0.08$) so the auxiliary terms do not dominate reconstruction.
% [中文] 使用 $\mathcal{L}_{\mathrm{video}}+\textstyle\sum_k w_k\mathcal{L}_k$ 训练，其中每个 $w_k$ 都很小（$0.03$--$0.08$），使辅助项不主导重建。
\end{enumerate}
\end{algorithm}

\subsection{EAG}
\label{app:eag}
EAG modifies sampling without retraining the video model. After each CFG update, a frozen latent-action model \citep{schmidt2024lapo,ye2025lapa} assigns a transition energy $E(\hat z_0)$ to the predicted clean latent $\hat z_0$ using a soft top-3 aggregation of per-transition reconstruction errors. The sampler then takes one normalized step along the negative energy gradient:
% [中文] EAG 在不重新训练视频模型的情况下修改采样过程。每次 CFG 更新后，一个冻结的 latent-action 模型 \citep{schmidt2024lapo,ye2025lapa} 通过对各转移的重建误差做 soft top-3 聚合，为预测的干净 latent $\hat z_0$ 赋予转移能量 $E(\hat z_0)$。随后采样器沿能量负梯度方向走一步归一化步长：
\begin{equation}
\hat z_0 \leftarrow \hat z_0 \;-\; w(\sigma)\,\|\hat z_0\|\,\frac{\nabla_{\hat z_0} E(\hat z_0)}{\|\nabla_{\hat z_0} E(\hat z_0)\|},
\qquad
w(\sigma)=\frac{w_0}{1+\sigma},
\label{eq:eag_update}
\end{equation}
where $w_0$ is the guidance weight and $w_0{=}0$ recovers the unmodified sampler. Against the weight-0 control, Gemini-IF rises from 41.67 to 45.83, while MLR rises from 15.4\% to 38.5\% (16 prompt--seed pairs, 13 detector-eligible). This small comparison is reported as a development result.
% [中文] 其中 $w_0$ 为引导权重，$w_0{=}0$ 时恢复为未修改的采样器。与权重为 0 的对照相比，Gemini-IF 从 41.67 升至 45.83，MLR 则从 15.4\% 升至 38.5\%（16 个提示词--种子对，13 个检测器符合条件）。这一小规模比较作为开发阶段结果报告。

\begin{algorithm}[H]
\caption{\textbf{EAG-guided denoising step.} Only the sampler changes; the model and the energy model stay frozen.}
\label{alg:eag}
\small
\textbf{Input:} denoiser $D_\theta$, frozen transition-energy model $E$, scheduler, guidance weight $w_0$.\\
\textbf{Output:} the guided denoising update at the current step.
\begin{enumerate}[leftmargin=1.6em,itemsep=1pt,topsep=2pt]
  \item Run the usual CFG update and form the clean-latent prediction $\hat z_0$ from the model output.
% [中文] 执行常规的 CFG 更新，并由模型输出形成干净 latent 预测 $\hat z_0$。
  \item Score $\hat z_0$ with the frozen transition energy $E(\hat z_0)$.
% [中文] 用冻结的转移能量 $E(\hat z_0)$ 为 $\hat z_0$ 打分。
  \item Take one normalized step down the energy (Eq.~\eqref{eq:eag_update}); the step length scales with $\|\hat z_0\|$, so the update is bounded.
% [中文] 沿能量下降方向走一步归一化步长（Eq.~\eqref{eq:eag_update}）；步长随 $\|\hat z_0\|$ 缩放，因此更新是有界的。
  \item Convert the change back into the model output space, $\Delta_{\mathrm{noise}}=\Delta \hat z_0 / c_{\mathrm{out}}(\sigma)$, and let the scheduler proceed as usual.
% [中文] 将该变化转换回模型输出空间，$\Delta_{\mathrm{noise}}=\Delta \hat z_0 / c_{\mathrm{out}}(\sigma)$，然后让调度器照常进行。
\end{enumerate}
\end{algorithm}

\subsection{LAD-LoRA}
\label{app:lad_lora}
LAD-LoRA moves the transition-energy signal from sampling to parameter updates. We train a rank-64 attention LoRA \citep{hu2022lora}, with the backbone frozen, by adding the energy of the model's denoised output as a regularizer:
% [中文] LAD-LoRA 将转移能量信号从采样阶段转移到参数更新中。在主干冻结的情况下，我们通过把模型去噪输出的能量作为正则项，训练一个秩为 64 的注意力 LoRA \citep{hu2022lora}：
\begin{equation}
\mathcal{L}=\mathcal{L}_{\mathrm{EDM}}+\frac{\tilde w}{1+\sigma}\,E(\hat z_{0,\theta}),
\label{eq:ladlora}
\end{equation}
where $\tilde w$ is recomputed at every step so that the energy gradient stays at a fixed fraction of the denoising gradient on the shared prediction $\hat z_0$ (target ratio 0.1, capped at 1.0). Against a plain LoRA control under the same protocol (32 matched pairs), LAD-LoRA has the same MLR (3.8\%) and reduces Gemini-IF from 51.04 to 47.92.
% [中文] 其中 $\tilde w$ 在每一步重新计算，使能量梯度在共享预测 $\hat z_0$ 上保持为去噪梯度的固定比例（目标比例 0.1，上限 1.0）。在相同协议下与普通 LoRA 对照比较时（32 个匹配对），LAD-LoRA 的 MLR 相同（3.8\%），Gemini-IF 则从 51.04 降至 47.92。

\begin{algorithm}[H]
\caption{\textbf{LAD-LoRA training step.} The backbone is frozen; only the rank-64 attention LoRA receives gradients.}
\label{alg:ladlora}
\small
\textbf{Input:} frozen backbone with LoRA adapters, frozen transition-energy model $E$, target gradient ratio $w_{\mathrm{lad}}{=}0.1$.\\
\textbf{Output:} LoRA parameter update.
\begin{enumerate}[leftmargin=1.6em,itemsep=1pt,topsep=2pt]
  \item Denoise the batch and form the model's own clean prediction $\hat z_{0,\theta}$.
% [中文] 对该批次去噪，并形成模型自身的干净预测 $\hat z_{0,\theta}$。
  \item Score $\hat z_{0,\theta}$ with the frozen energy $E$ (the same soft top-3 transition energy as EAG).
% [中文] 用冻结的能量 $E$ 为 $\hat z_{0,\theta}$ 打分（与 EAG 相同的 soft top-3 转移能量）。
  \item Measure the gradient norms of the denoising loss and of $E$ on the shared $\hat z_0$, set the energy weight to hit the target gradient ratio, and gate by $1/(1+\sigma)$.
% [中文] 在共享的 $\hat z_0$ 上测量去噪损失和 $E$ 的梯度范数，将能量权重设为达到目标梯度比例的值，并乘以 $1/(1+\sigma)$ 门控。
  \item Back-propagate Eq.~\eqref{eq:ladlora} into the LoRA parameters only.
% [中文] 仅将 Eq.~\eqref{eq:ladlora} 反向传播到 LoRA 参数。
\end{enumerate}
\end{algorithm}

\subsection{Causal Frontier}
\label{app:causal_frontier}
Causal Frontier generates a video autoregressively in blocks of four latent frames. At each step, the model sees only the clean latents of previously committed blocks, which a near-zero noise level and a mask channel mark as history. The denoising loss applies only to the active block. Generation commits each block irreversibly before denoising the next. At the current budget, prediction errors compound over the rollout \citep{bengio2015scheduled,chen2024diffusionforcing,huang2025selfforcing}: against control50 (16 matched pairs), MLR rises from 30.8\% to 41.7\% and Gemini-IF falls from 27.08 to 12.50.
% [中文] Causal Frontier 以四个 latent 帧为一块，自回归地生成视频。每一步中，模型只能看到已提交块的干净 latent，这些块以接近零的噪声水平和一个掩码通道标记为历史。去噪损失只作用于当前活动块。生成过程中，每个块在去噪下一块之前被不可逆地提交。在当前预算下，预测误差沿 rollout 不断累积 \citep{bengio2015scheduled,chen2024diffusionforcing,huang2025selfforcing}：与 control50（16 个匹配对）相比，MLR 从 30.8\% 升至 41.7\%，Gemini-IF 从 27.08 降至 12.50。

\begin{algorithm}[H]
\caption{\textbf{Causal Frontier training and generation.} Future blocks never enter the input; generation commits each block irreversibly.}
\label{alg:causal_frontier}
\small
\textbf{Input:} video of 24 latent frames split into blocks of 4; the first latent frame is the condition.
\begin{enumerate}[leftmargin=1.6em,itemsep=1pt,topsep=2pt]
  \item[] \textbf{Training.} \emph{(i)} Sample one frontier uniformly per step. \emph{(ii)} Feed the committed prefix as clean, detached latents with a near-zero noise level ($\sigma_h{=}10^{-4}$) and a mask channel set to 1. \emph{(iii)} Corrupt the active block at noise $\sigma$ and concatenate prefix and active block as the input; future blocks are excluded. \emph{(iv)} Apply the denoising loss on the active block only, normalized by the number of active tokens.
% [中文] \item[\textbf{训练。}] \emph{(i)} 每步均匀采样一个 frontier。\emph{(ii)} 将已提交前缀作为干净、detach 的 latent 输入，噪声水平接近零（$\sigma_h{=}10^{-4}$），掩码通道置为 1。\emph{(iii)} 以噪声 $\sigma$ 破坏当前活动块，并将前缀与活动块拼接作为输入；未来块被排除在外。\emph{(iv)} 仅在活动块上应用去噪损失，并按活动 token 数归一化。
  \item[] \textbf{Generation.} \emph{(i)} Denoise the next block (30 steps, CFG 7.0) conditioned on the clean prefix. \emph{(ii)} Commit the block: detach and append it to the prefix. \emph{(iii)} Repeat until all blocks are generated; there is no overlap or refinement pass.
% [中文] \item[\textbf{生成。}] \emph{(i)} 以干净前缀为条件对下一块去噪（30 步，CFG 7.0）。\emph{(ii)} 提交该块：detach 并追加到前缀中。\emph{(iii)} 重复直至所有块生成完毕；没有重叠或修正过程。
\end{enumerate}
\end{algorithm}

\subsection{ICH-D}
\label{app:ichd}
ICH-D applies a learned deletion residual during inference. A frozen linear probe estimates the duplicate probability of each latent cell from novelty and consistency features (16-dimensional statistics collected from several Transformer blocks). A gated residual, whose output projection is initialized to zero and trained before inference, then removes the flagged content from the block-22 representation:
% [中文] ICH-D 在推理阶段应用一个可学习的删除残差。一个冻结的线性探针根据新颖度和对应关系特征（从若干 Transformer 块收集的 16 维统计量）估计每个 latent 单元的重复概率。随后，一个带门控、零初始化的残差从第 22 块的表示中移除被标记的内容：
\begin{equation}
M=\mathrm{sigmoid}\!\big(\beta^\top\phi(h)\big),
\qquad
h_{22}\leftarrow h_{22}-\mathrm{sigmoid}(g)\cdot M\odot W_{\mathrm{out}}(h_{22}),
\label{eq:ichd}
\end{equation}
after which the denoising loop proceeds normally. Over 368 matched pairs against the pretrained model (four seeds), MLR rises from 11.2\% to 13.1\% and Gemini-IF falls from 45.83 to 31.79; the deletion path introduces its own errors.
% [中文] 之后去噪循环照常进行。在与预训练模型的 368 个匹配对（四个种子）上，MLR 从 11.2\% 升至 13.1\%，Gemini-IF 从 45.83 降至 31.79；删除路径引入了自身的错误。

\begin{algorithm}[H]
\caption{\textbf{ICH-D inference-time deletion.} The probe $\beta$ is frozen after offline fitting; $W_{\mathrm{out}}$ is initialized to zero, trained beforehand, and frozen at inference.}
\label{alg:ichd}
\small
\textbf{Input:} frozen per-cell probe $\beta$, trained gate $g$, and trained output projection $W_{\mathrm{out}}$.\\
\textbf{Output:} corrected block-22 representation at each denoising step.
\begin{enumerate}[leftmargin=1.6em,itemsep=1pt,topsep=2pt]
  \item At the block-22 output of every denoising forward pass (both CFG branches), compute the per-cell novelty and consistency features $\phi$.
% [中文] 在每次去噪前向传播（两个 CFG 分支）的第 22 块输出处，计算逐单元的新颖度和对应关系特征 $\phi$。
  \item Score each cell with the frozen probe, $M=\mathrm{sigmoid}(\beta^\top\phi)$, giving a duplication-probability map over the latent grid.
% [中文] 用冻结探针为每个单元打分，$M=\mathrm{sigmoid}(\beta^\top\phi)$，得到 latent 网格上的重复概率图。
  \item Erase the flagged content inside the representation via Eq.~\eqref{eq:ichd}.
% [中文] 通过 Eq.~\eqref{eq:ichd} 在表示内部擦除被标记的内容。
  \item Continue the denoising loop normally; there is no re-noising or repair pass.
% [中文] 照常继续去噪循环；没有重新加噪或修复过程。
\end{enumerate}
\end{algorithm}

Across the matched comparisons, the five alternatives trade off MLR and instruction following, motivating direct restoration and consistency supervision in the backbone.
% [中文] 在匹配比较中，五种备选方案均表现出 MLR 与指令遵循之间的权衡，因而采用主干内的直接恢复与对应监督。

\FloatBarrier
\section{Generalization Results}
% [中文] 泛化结果
\label{app:flowwam_benchmarks}
\label{app:cross_backbone}

\subsection{Generalization on EWMBench}
% [中文] 完整的 EWMBench 结果
\label{app:ewmbench_full}

Table~\ref{tab:agibot_transfer_full} reports all EWMBench components and the
official Motion, Semantics, and Overall scores.
Relative to Standard SFT, \method{} improves Motion by 3.48\% and Overall by
1.24\%, with a 17.94\% gain in DYN.
Other components remain close to the matched baseline, with Standard SFT
higher on BLEU, Diversity, HSD, and nDTW, and GigaWorld-0 higher on CLIP and
Scene.
Logic is identical across all three models.
The gains of \method{} are concentrated in motion-related and aggregate scores
rather than uniformly across all EWMBench components.
% [中文] 表~\ref{tab:agibot_transfer_full} 报告 EWMBench 的全部八个分量及官方 Motion、Semantics 和 Overall 聚合。相对 Standard SFT，\method{} 将 Motion 从 0.62 提高到 0.64，将 Overall 从 3.71 提高到 3.75，并取得最高的 DYN 和 Semantics。Standard SFT 在 BLEU、Diversity、HSD 和 nDTW 上更高，GigaWorld-0 在 CLIP 和 Scene 上更高。三个模型的 Logic 相同。

\begin{table}[htbp]
\centering
\caption{
\textbf{Complete EWMBench results.}
Panel (a) reports semantic and scene components.
Panel (b) reports motion components and the official Motion, Semantics, and Overall scores.
Scores are rescaled for readability: component and Motion scores are multiplied by 100, and Semantics and Overall by 10.
}
\label{tab:agibot_transfer_full}
\renewcommand{\arraystretch}{1.02}

\textbf{(a) Semantic and scene components}\\[2pt]

{\setlength{\tabcolsep}{8.4pt}
\begin{tabular}{lccccc}
\toprule
Method
& BLEU $\uparrow$
& CLIP $\uparrow$
& Diversity $\uparrow$
& Logic $\uparrow$
& Scene $\uparrow$ \\
\midrule
GigaWorld-0
& 33.12
& \textbf{89.09}
& 2.77
& \textbf{100.00}
& \textbf{89.58} \\

Standard SFT
& \textbf{34.44}
& 86.86
& \textbf{3.47}
& \textbf{100.00}
& 84.38 \\

\rowcolor{gray!20}
\method{}
& 34.23
& 87.74
& 3.26
& \textbf{100.00}
& 86.37 \\
\bottomrule
\end{tabular}}

\vspace{6pt}

\textbf{(b) Motion components and aggregate scores}\\[2pt]

{\setlength{\tabcolsep}{8.4pt}
\begin{tabular}{lcccccc}
\toprule
Method
& DYN $\uparrow$
& HSD $\uparrow$
& nDTW $\uparrow$
& Motion $\uparrow$
& Semantics $\uparrow$
& Overall $\uparrow$ \\
\midrule
GigaWorld-0
& 7.19
& 23.29
& 19.83
& 50.30
& 22.50
& 36.49 \\

Standard SFT
& 14.88
& \textbf{24.35}
& \textbf{22.28}
& 61.51
& 22.48
& 37.07 \\

\rowcolor{gray!20}
\method{}
& \textbf{17.55}
& 24.12
& 21.97
& \textbf{63.65}
& \textbf{22.52}
& \textbf{37.53} \\
\bottomrule
\end{tabular}}
\end{table}

\subsection{FlowWAM Checkpoint Robustness}
% [中文] FlowWAM 跨检查点结果
\label{app:heldout_action}

We evaluate six FlowWAM checkpoints on a fixed RoboTwin slice consisting of
episode 45 from each of 50 tasks \citep{mu2025robotwin}.
\method{} improves PSNR by 2.20--6.22\,dB and increases SSIM at every
checkpoint (Table~\ref{tab:flowwam_action_fidelity}), showing consistent
reference-fidelity gains across checkpoints.
% [中文] 我们在固定的 RoboTwin 切片上测试 FlowWAM 六个检查点的保真度 \citep{mu2025robotwin}，即 50 个任务各取第 45 个 episode。\method{} 在每个检查点上都提高 PSNR 和 SSIM（表~\ref{tab:flowwam_action_fidelity}）。主比较使用单独选择的检查点和全部 250 个留出 episode。Core-8 是同一评估器八个无参考指标的均值，与基于参考的 PSNR 和 SSIM 一并报告。

\begin{table}[htbp]
\centering
\caption{
\textbf{FlowWAM fidelity across checkpoints} on a fixed RoboTwin slice.
Each checkpoint is evaluated on episode 45 from each of 50 tasks.
Core-8 denotes the mean of eight no-reference metrics from the same evaluator.
$\Delta$PSNR reports the gain of \method{} over the control.
}
\label{tab:flowwam_action_fidelity}
\setlength{\tabcolsep}{4.5pt}

\begin{tabular}{lccccccc}
\toprule
& \multicolumn{3}{c}{Control}
& \multicolumn{3}{c}{\method{}} \\
\cmidrule(lr){2-4}
\cmidrule(lr){5-7}
Epoch
& PSNR (dB) $\uparrow$
& SSIM $\uparrow$
& Core-8 $\uparrow$
& PSNR (dB) $\uparrow$
& SSIM $\uparrow$
& Core-8 $\uparrow$
& $\Delta$PSNR (dB) \\
\midrule
0 & 16.79 & 0.807 & 47.8 & 18.99 & 0.857 & 43.3 & +2.20 \\
1 & 15.38 & 0.785 & 47.0 & 20.23 & 0.871 & 48.0 & +4.85 \\
2 & 14.92 & 0.757 & 50.5 & 21.14 & 0.885 & 45.9 & +6.22 \\
3 & 15.76 & 0.785 & 50.1 & 18.44 & 0.839 & 48.7 & +2.68 \\
4 & 15.02 & 0.770 & 48.2 & 18.02 & 0.834 & 47.9 & +3.00 \\
5 & 15.18 & 0.767 & 48.5 & 18.45 & 0.843 & 47.8 & +3.27 \\
\bottomrule
\end{tabular}
\end{table}

\FloatBarrier
% Tightened float spacing: the qualitative plates (Figures 7-16) are collected
% in one block after the appendix text, so the small separation keeps the plate
% pages densely packed.
\begingroup
\setlength{\textfloatsep}{2pt}
\setlength{\abovecaptionskip}{0pt}
\section{Qualitative Results}
% [中文] 定性结果
\label{app:visuals}

\subsection{Instance Consistency}
% [中文] 实例一致性
\label{app:qual_laziness}

Figures~\ref{fig:consistency_cases} and \ref{fig:consistency_cases_more} provide qualitative examples of target-instance consistency during manipulation. Standard SFT may introduce surplus or missing target instances over the rollout, whereas \method{} better maintains target-instance consistency throughout the interaction.
% [中文] 图~\ref{fig:consistency_cases} 和图~\ref{fig:consistency_cases_more} 给出操作过程中目标实例一致性的定性示例。Standard SFT 在 rollout 中可能出现多余或缺失的目标实例，而完整方法在交互过程中更好地保持目标实例一致性。

\subsection{Cross-Frame Consistency}
% [中文] 跨帧一致性
\label{app:qual_matched}

Figures~\ref{fig:qual_dup1} and~\ref{fig:qual_dup2} compare Standard SFT, IGR-only, and the full \method{} at matched stages of the same manipulation. Standard SFT exhibits target duplication during the rollout, and IGR-only can still retain such cross-frame inconsistencies despite directly supervising instance restoration. With TIA added, the full \method{} maintains a single coherent target across the corresponding interaction stages. These examples illustrate the complementary roles of IGR and TIA in preserving target consistency over time.
% [中文] 图~\ref{fig:qual_dup1} 与图~\ref{fig:qual_dup2} 在同一操作任务的匹配阶段对比 Standard SFT、仅 IGR 与完整方法。Standard SFT 在 rollout 过程中出现目标复制；仅 IGR 尽管直接监督实例还原，仍可能残留这类跨帧不一致。加入 TIA 后，完整方法在相应交互阶段始终维持单一一致的目标。这两个例子说明 IGR 与 TIA 在保持目标跨帧一致性上的互补作用。

\subsection{Target Evolution Comparison}
% [中文] 保数畸变：数量不变但目标被损坏，主 MLR 看不到
\label{app:qual_distortion}

MLR intentionally measures changes in target-instance count and therefore does not capture failures that preserve $N_0$ while corrupting the target itself. For qualitative analysis, we assess such distortion separately using the same VLM-assisted temporal review framework described in Appendix~\ref{sec:vlm_mlr_audit}, supplemented by manual review for a subset of cases. A distortion is marked when the target count remains unchanged but the target undergoes an \emph{unsupported} change in shape, appearance, or identity that is inconsistent with its preceding temporal state. Figures~\ref{fig:audit570_sft} to~\ref{fig:audit637_ours} show two representative tasks using the same 24 timestamps for all models. In both examples, Standard SFT loses the target and thus produces an MLR count violation, whereas IGR-only preserves the count but progressively distorts the target. \method{} preserves both the target count and its cross-frame consistency.
% [中文] MLR 有意只测目标实例数量的变化，因此看不见「数量保持不变、但目标本身被损坏」的失败。定性分析上，这类畸变复用附录~\ref{sec:vlm_mlr_audit} 的同一套 VLM 时间审核框架单独评估，部分案例另加人工复核。判定口径是目标数量不变、但目标发生了与前一状态不一致的「无支撑」形变、外观变化或身份漂移。图~\ref{fig:audit570_sft} 至图~\ref{fig:audit637_ours} 给出两个代表性任务，所有模型使用同一组 24 个时间戳。两个例子里 Standard SFT 都把目标弄丢，属于主 MLR 本就能抓到的数量违规；IGR-only 保住了数量，但目标逐步畸变；完整方法两者都保住。

\begin{figure}[htbp]
  \centering
  \safeincludegraphics[width=\textwidth]{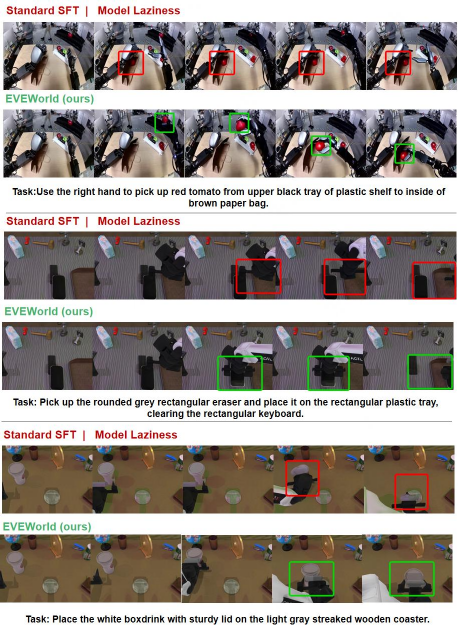}
\caption{
\textbf{Target-instance consistency under Model Laziness.}
\textcolor{red}{Red} marks Standard SFT, where Model Laziness violates target-instance consistency during manipulation.
\textcolor{green!50!black}{Green} marks \method{}, which maintains target-instance consistency throughout the interaction.
}
  \label{fig:consistency_cases}
\end{figure}

\begin{figure}[htbp]
  \centering
  \safeincludegraphics[width=\textwidth]{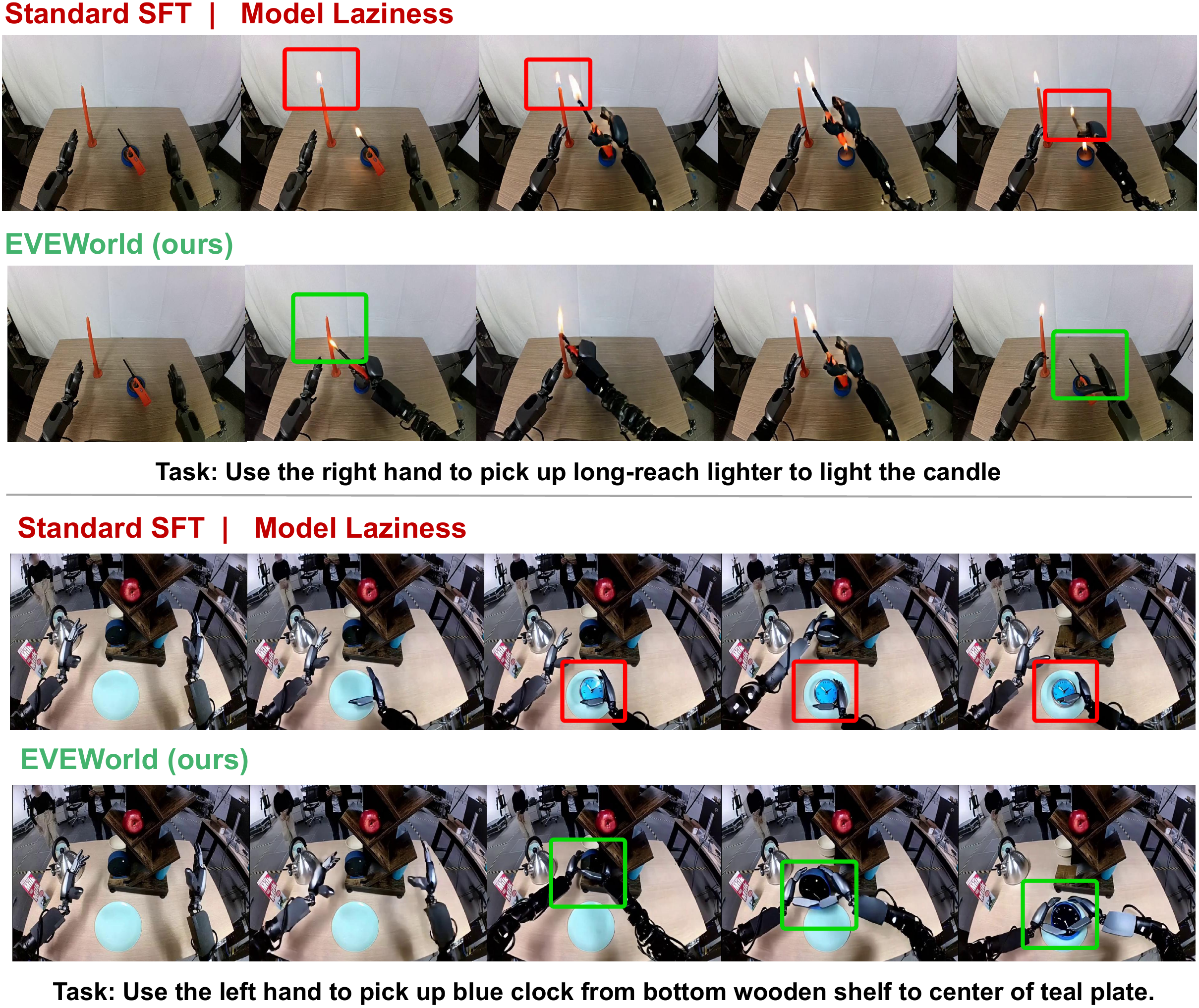}

  \vspace{6pt}
  \safeincludegraphics[width=\textwidth]{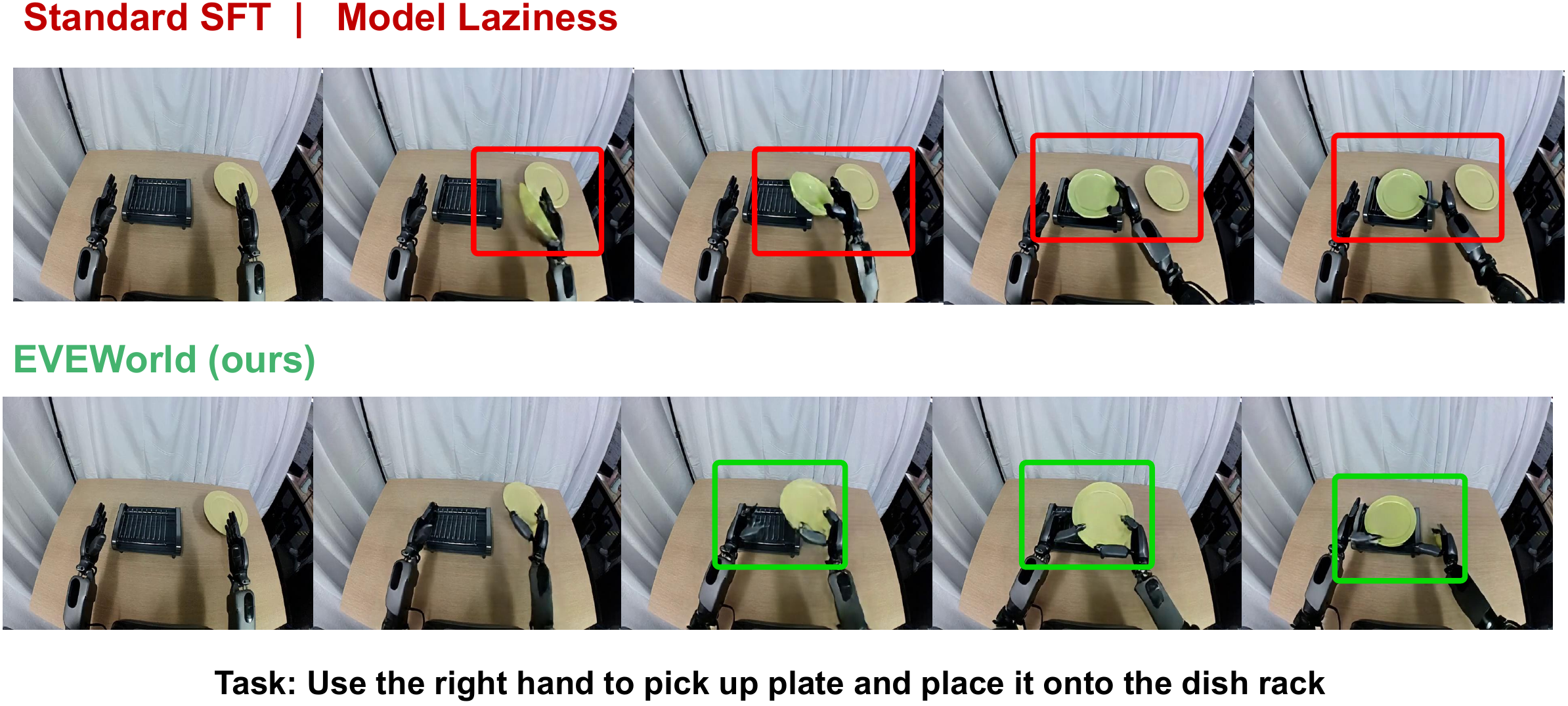}
\caption{
\textbf{Additional Model Laziness cases.}
\textcolor{red}{Red} marks Standard SFT, where target-instance consistency is violated during manipulation.
\textcolor{green!50!black}{Green} marks \method{}, which maintains target-instance consistency throughout the interaction.
}
  \label{fig:consistency_cases_more}
\end{figure}

\begin{figure}[htbp]
  \centering
  \safeincludegraphics[width=\textwidth]{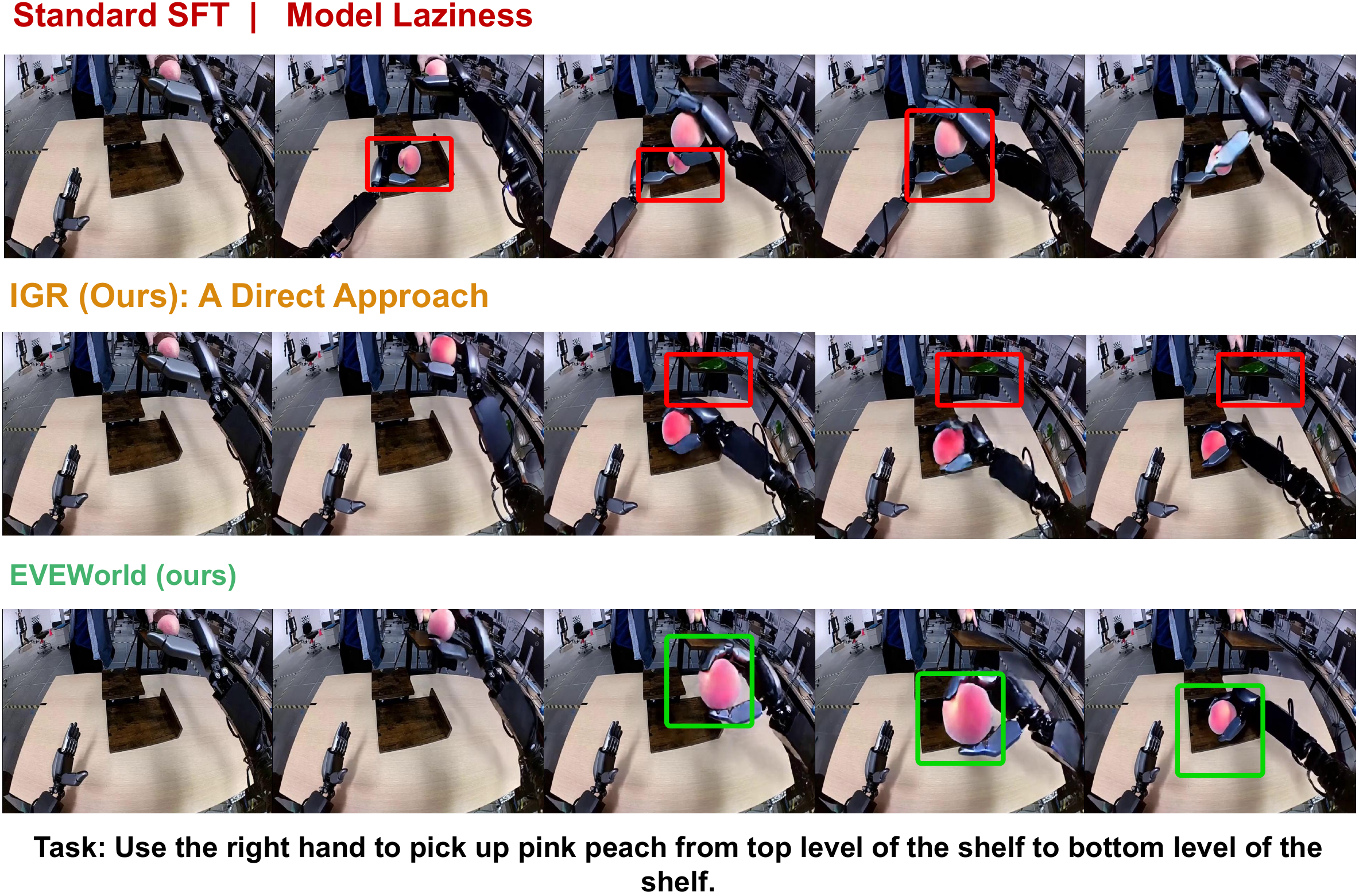}
\caption{
\textbf{Cross-frame target consistency.}
Standard SFT and IGR-only show target duplication, appearance drift, or
deformation (\textcolor{red}{red}), while the full \method{} preserves target
identity and appearance across matched interaction stages
(\textcolor{green!50!black}{green}).
}
  \label{fig:qual_dup1}
\end{figure}

\begin{figure}[htbp]
  \centering
  \safeincludegraphics[width=\textwidth]{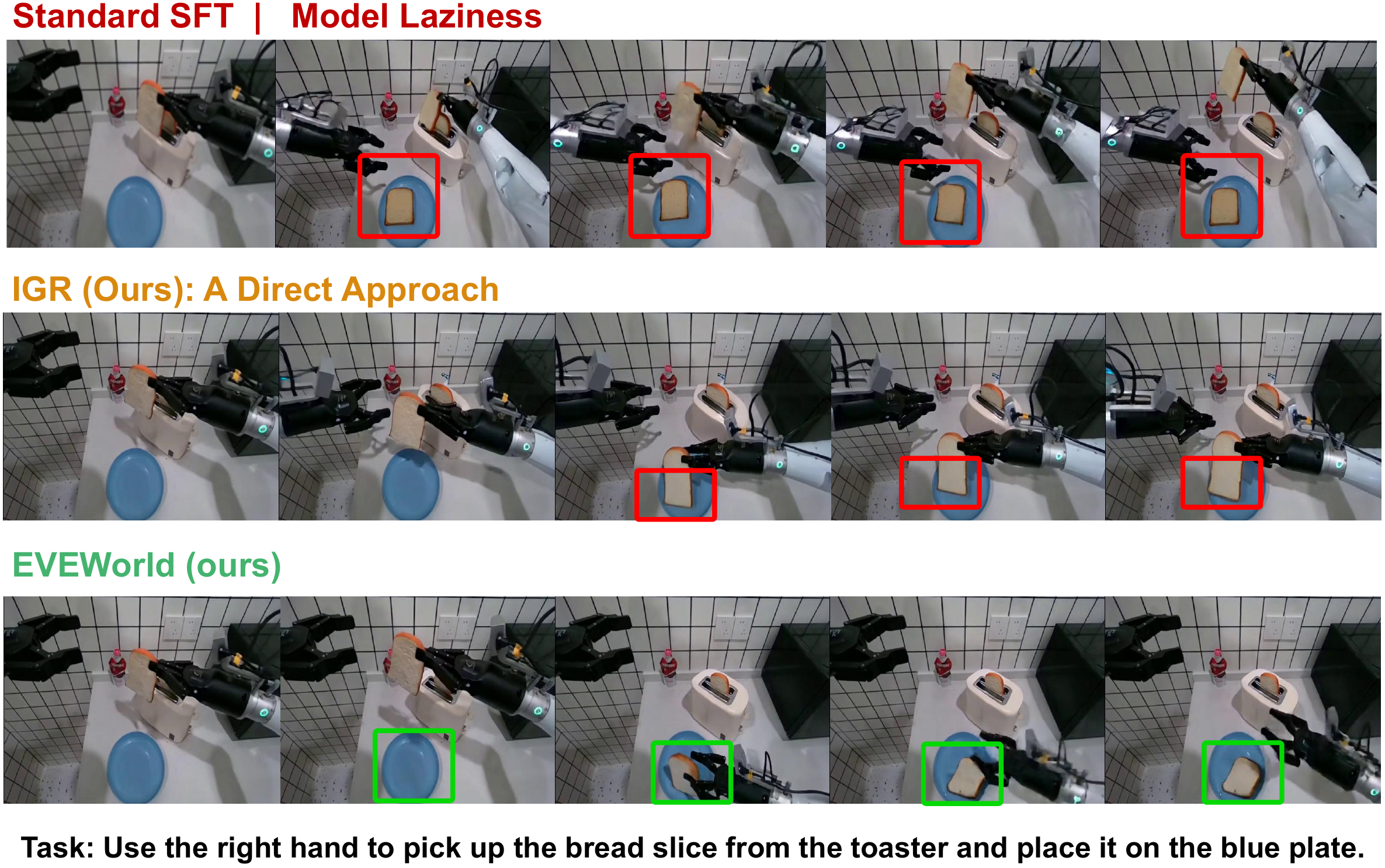}
\caption{
\textbf{Cross-frame target consistency.}
Standard SFT and IGR-only show target duplication, appearance drift, or
deformation (\textcolor{red}{red}), while the full \method{} preserves target
identity and appearance across matched interaction stages
(\textcolor{green!50!black}{green}).
}
  \label{fig:qual_dup2}
\end{figure}

\begin{figure}[htbp]
  \centering
  \safeincludegraphics[width=\textwidth]{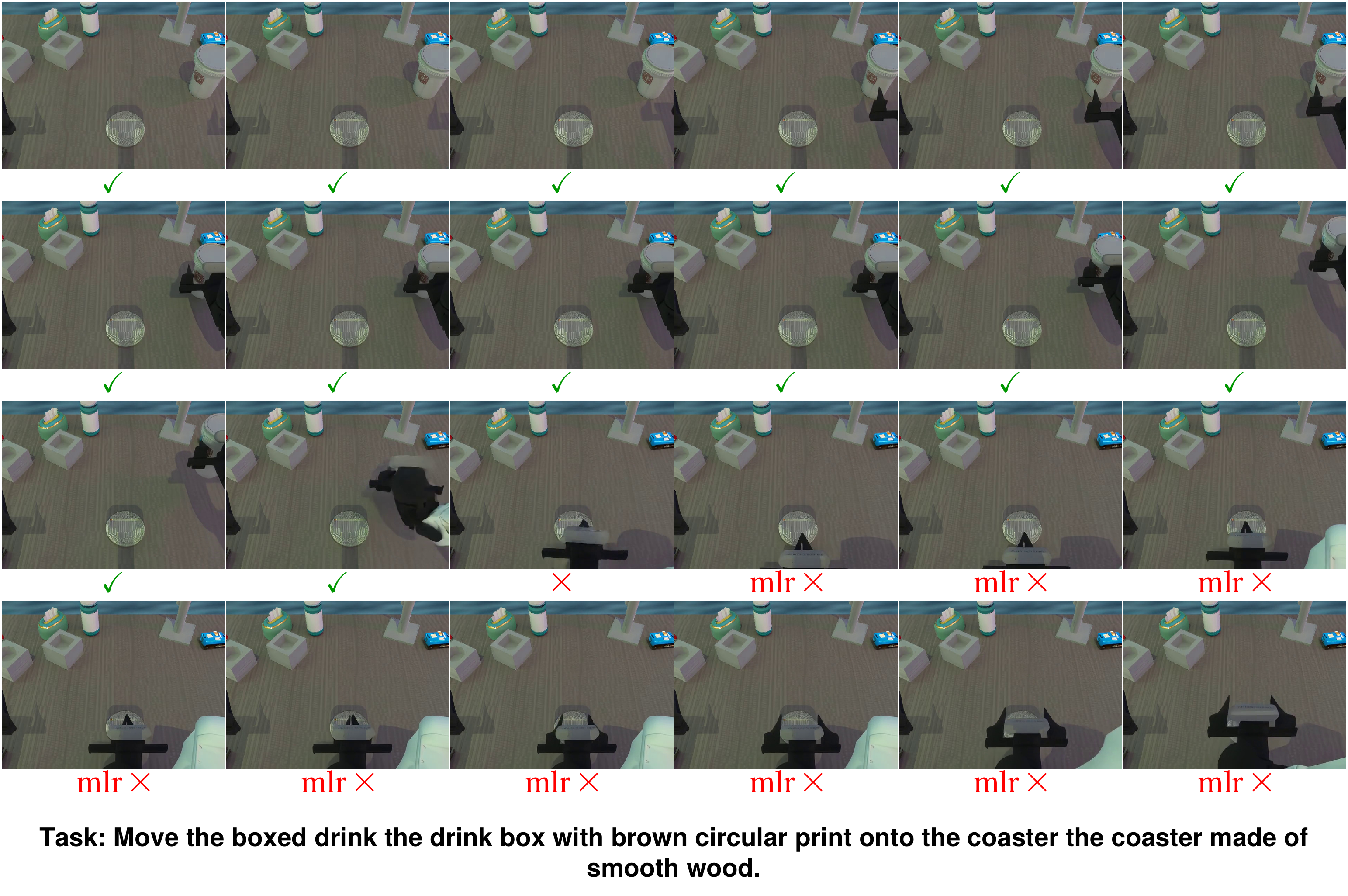}
\caption{
\textbf{Instance inconsistency.}
The 24 fixed MLR timestamps are shown in temporal order.
\textcolor{green!50!black}{Check marks} denote count-consistent states, while
\textcolor{red}{crosses} denote count deviations. The first cross marks the
onset; an MLR event is counted only after two consecutive violating timestamps
to suppress isolated detection errors.
}
  \label{fig:audit570_sft}
\end{figure}

\begin{figure}[htbp]
  \centering
  \safeincludegraphics[width=\textwidth]{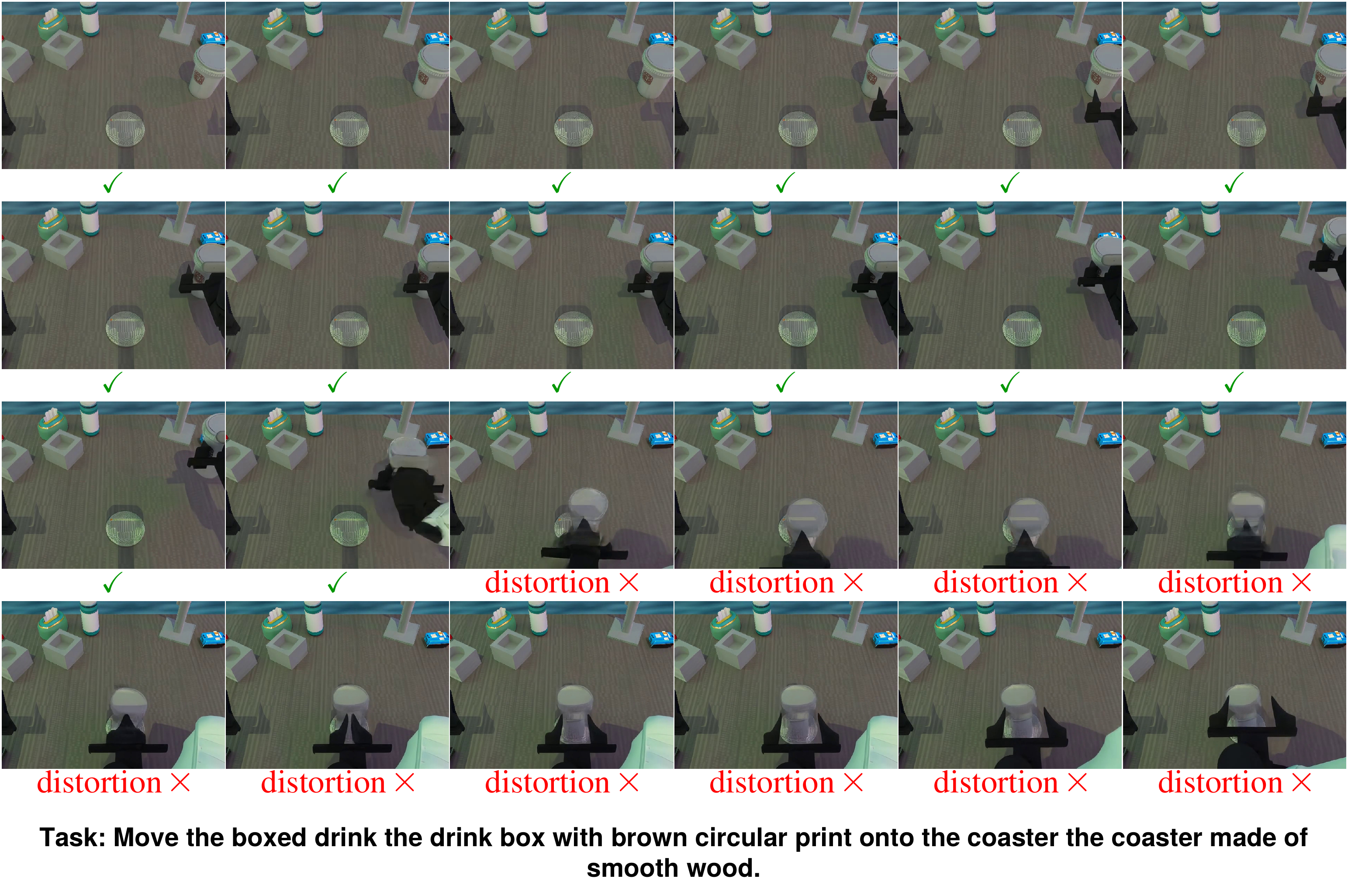}
\caption{
\textbf{Cross-frame inconsistency.}
The 24 fixed timestamps are shown in temporal order.
\textcolor{green!50!black}{Check marks} denote consistent target states, while
\textcolor{red}{crosses} mark persistent appearance, geometry, or identity
changes. The target count is preserved, so these failures are recorded
separately from MLR.
}
  \label{fig:audit570_igr}
\end{figure}

\begin{figure}[htbp]
  \centering
  \safeincludegraphics[width=\textwidth]{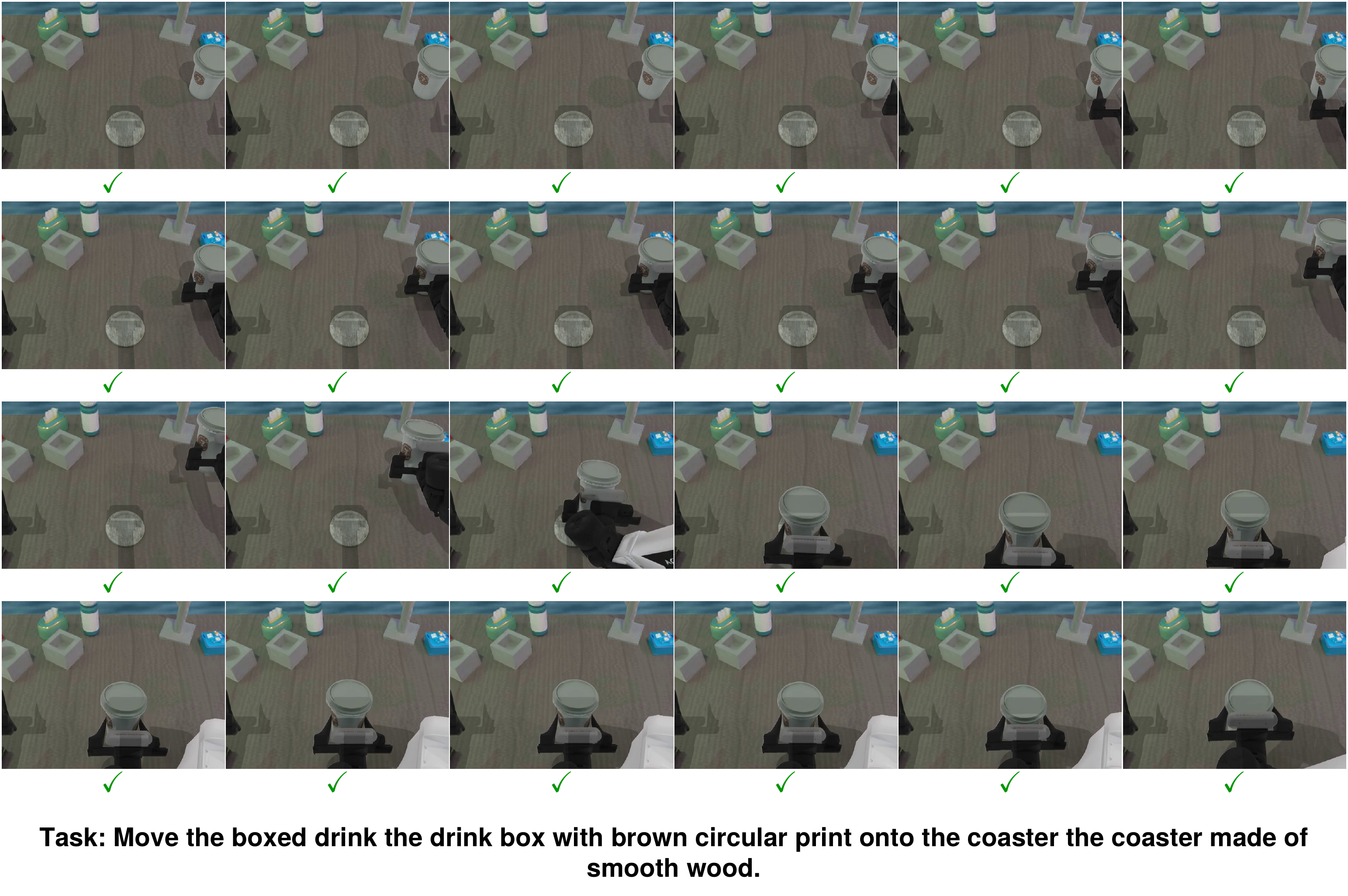}
\caption{
\textbf{Physical consistency.}
The 24 fixed timestamps are shown in temporal order.
\textcolor{green!50!black}{Check marks} indicate consistent target states
throughout the rollout. The target preserves both its instance count and
cross-frame appearance, with no persistent count violation or distortion.
}
  \label{fig:audit570_ours}
\end{figure}

\begin{figure}[htbp]
  \centering
  \safeincludegraphics[width=\textwidth]{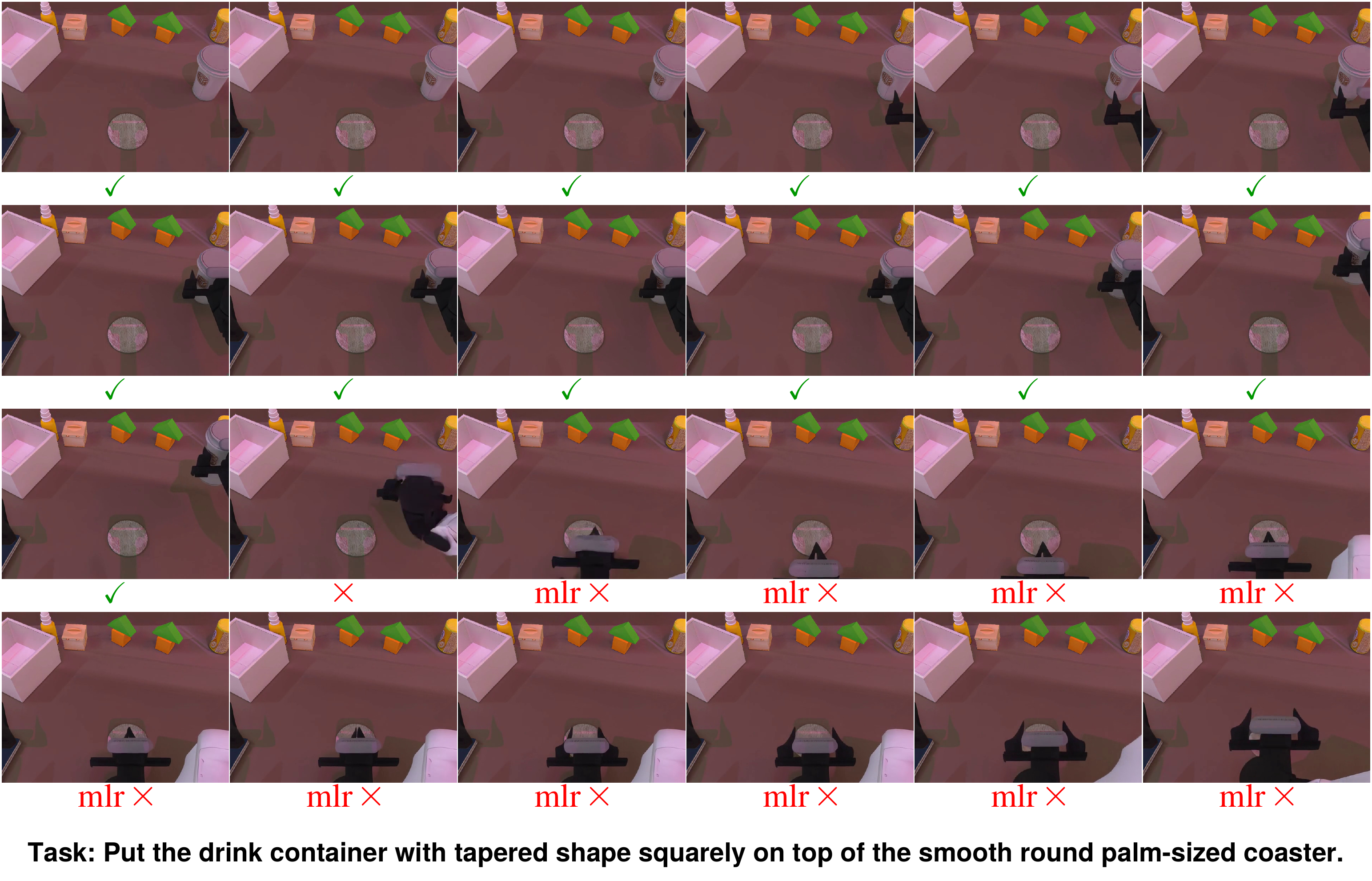}
\caption{
\textbf{Instance inconsistency.}
The 24 fixed MLR timestamps are shown in temporal order.
\textcolor{green!50!black}{Check marks} denote count-consistent states, while
\textcolor{red}{crosses} denote count deviations. The first cross marks the
onset; an MLR event is counted only after two consecutive violating timestamps
to suppress isolated detection errors.
}
  \label{fig:audit637_sft}
\end{figure}

\begin{figure}[htbp]
  \centering
  \safeincludegraphics[width=\textwidth]{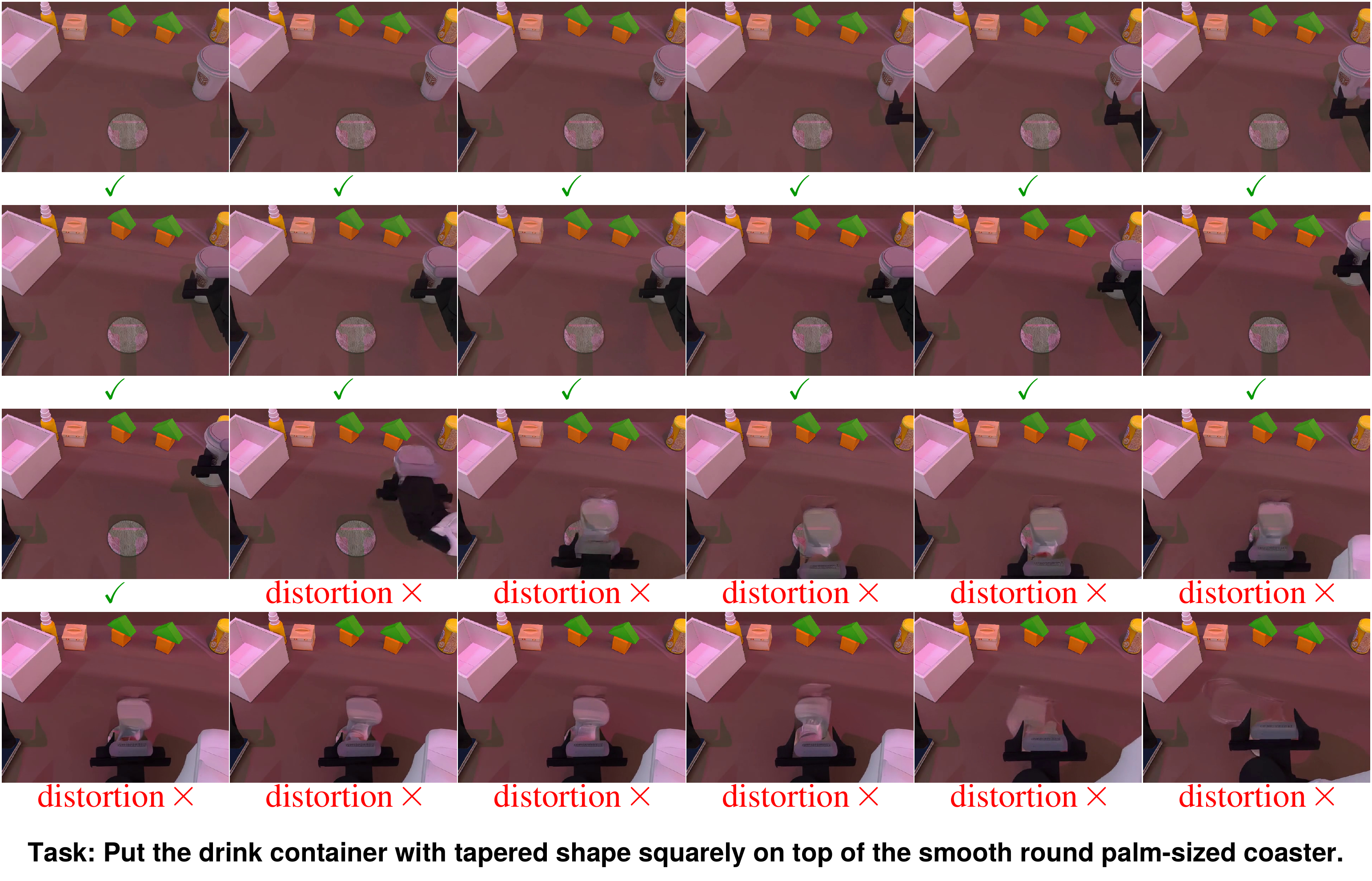}
\caption{
\textbf{Cross-frame inconsistency.}
The 24 fixed timestamps are shown in temporal order.
\textcolor{green!50!black}{Check marks} denote consistent target states, while
\textcolor{red}{crosses} mark persistent appearance, geometry, or identity
changes. The target count is preserved, so these failures are recorded
separately from MLR.
}
  \label{fig:audit637_igr}
\end{figure}

\begin{figure}[htbp]
  \centering
  \safeincludegraphics[width=\textwidth]{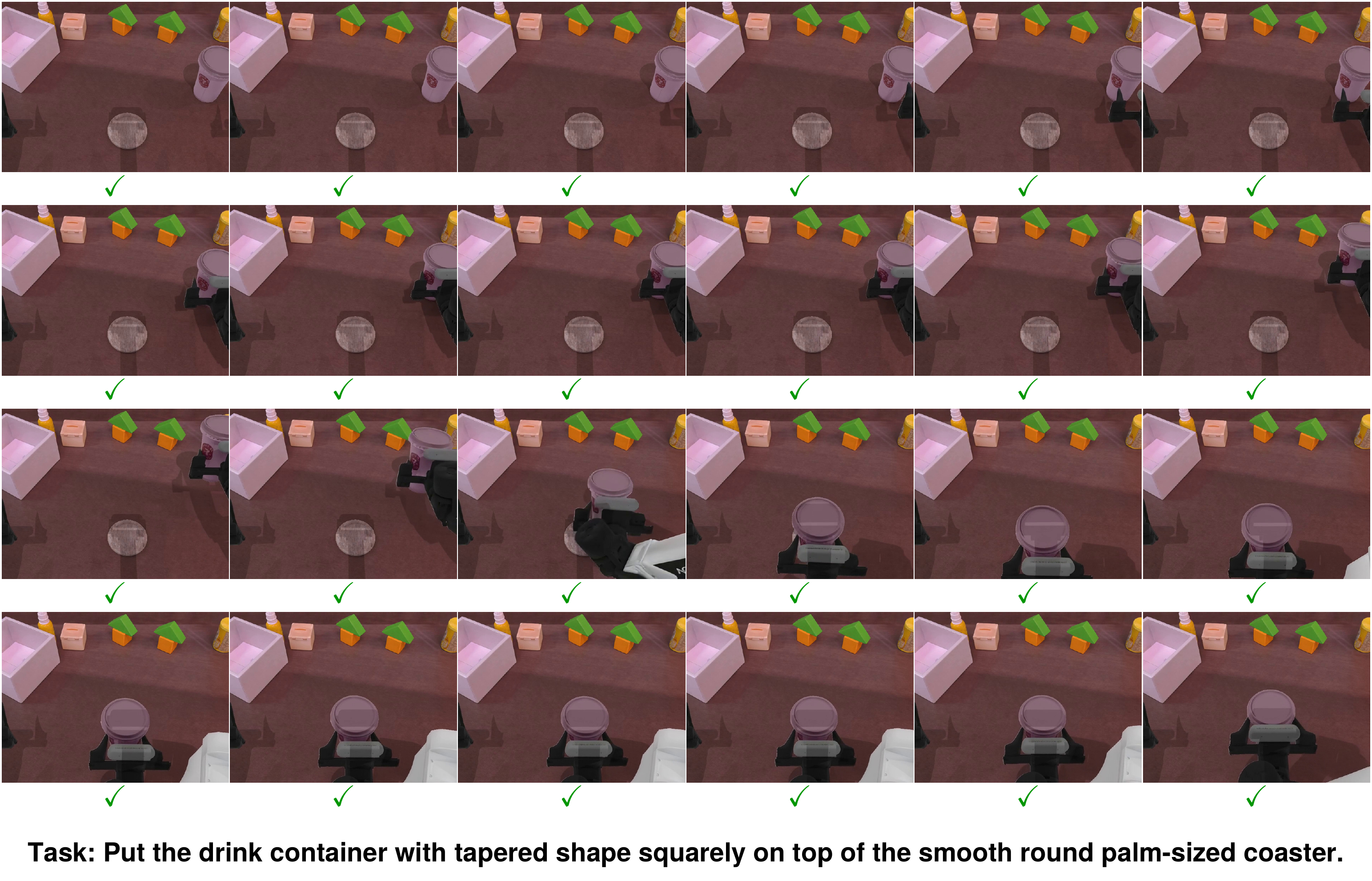}
\caption{
\textbf{Physical consistency.}
The 24 fixed timestamps are shown in temporal order.
\textcolor{green!50!black}{Check marks} indicate consistent target states
throughout the rollout. The target preserves both its instance count and
cross-frame appearance, with no persistent count violation or distortion.
}
  \label{fig:audit637_ours}
\end{figure}
\endgroup

\FloatBarrier

\raggedbottom
\section{Related Work}
\label{app:related_work}

We group related work into structured embodied world models and physical alignment, object editing, and approaches based on correspondence, tracking, or memory.

\paragraph{Structured embodied world models and physical alignment.}
\begin{itemize}[leftmargin=1.5em,topsep=2pt,itemsep=2pt,parsep=0pt]
  \item \textbf{4D scene representations.} TesserAct extends video generation to jointly predict RGB, depth, and surface normals for 4D scene reconstruction \citep{zhen2025tesseract}. While its richer outputs improve geometric modeling, \method{} retains the host model's native representation and adds post-training restoration supervision for target-instance consistency.

  \item \textbf{Disentangled dynamics.} DVG-WM separates a low-resolution dynamics preview from high-fidelity video refinement and evaluates whether the robot acts on the intended object \citep{shan2026dvgwm}. Correct target selection, however, does not ensure that the target remains a single consistent instance throughout the interaction; IGR explicitly trains recovery from count-perturbed demonstrations.

  \item \textbf{Object-aware training.} DreamX-Phi combines action conditioning and depth prediction with target-mask-weighted reconstruction and relational supervision from a frozen V-JEPA teacher \citep{dreamxteam2026dreamxphi}. TIA adds layer-selected local matching and feature transport for the manipulated target across adjacent frames, while IGR supplies an explicit restoration target when instance consistency is violated.

  \item \textbf{Physics-preference learning.} ABot-PhysWorld uses vision-language discriminators to select preferences for physically plausible interaction videos, including contact behavior \citep{chen2026abotphysworld}. Its video-level preference signal does not explicitly pair a duplicated target with a clean future; IGR uses that pairing to directly supervise instance-consistent target evolution.
\end{itemize}

\paragraph{Object editing and corrective supervision.}
\begin{itemize}[leftmargin=1.5em,topsep=2pt,itemsep=2pt,parsep=0pt]
  \item \textbf{Mask-free video editing.} LoVoRA uses learnable localization to add or remove objects in a supplied video from text, without an external mask at inference \citep{xiao2025lovora}. IGR instead uses object insertion as a training perturbation; \method{} predicts the future from an initial image and instruction without an edit request identifying a spurious target.

  \item \textbf{Effect-aware removal.} ROSE removes a specified object and its effects on the scene, including shadows and reflections, from an existing video \citep{miao2025rose}. IGR uses the inserted target for a different purpose: restoring the clean manipulation demonstration during post-training, so the world model learns instance consistency before generating future interactions.
\end{itemize}

\paragraph{Correspondence, tracking, and persistent memory.}
\begin{itemize}[leftmargin=1.5em,topsep=2pt,itemsep=2pt,parsep=0pt]
  \item \textbf{Joint generation and tracking.} Track4Gen jointly learns video generation and point tracking to reduce appearance drift \citep{jeong2025track4gen}. TIA instead selects a layer using target-matching error and locally matches and transports the manipulated target's features between adjacent frames, directly supervising its cross-frame consistency during generation.

  \item \textbf{Geometry-distilled correspondence.} GEM-4D distills dense 4D correspondence from a geometry model into the video backbone, improving geometric consistency without an auxiliary branch at inference \citep{zhou2026gem4d}. TIA focuses its matching and transport on the manipulated target, while IGR adds restoration supervision for instance errors that geometric correspondence alone does not explicitly correct.

  \item \textbf{Scene-revisitation memory.} WorldMem stores observations indexed by poses and timestamps to preserve spatial and temporal consistency when a simulated agent revisits a scene \citep{xiao2025worldmem}. Our evolution supervision instead addresses errors in how the manipulated target changes during an interaction, including unsupported duplication and cross-frame drift.

  \item \textbf{Object-centric memory.} SlotMemory organizes key-value features around reusable object slots, improving long-video entity persistence and showing qualitative object-count stability \citep{dou2026slotmemory}. IGR complements such persistence by explicitly restoring count-perturbed manipulation videos to supervise instance consistency, rather than relying on stored entity features alone.

  \item \textbf{Manipulation history retrieval.} Mem-World uses a 4D surfel memory and action-conditioned history retrieval to maintain scene information under wrist-camera motion and occlusion \citep{zheng2026memworld}. This historical context complements our direct supervision of the manipulated target: IGR promotes instance consistency, and TIA promotes cross-frame consistency during video prediction.
\end{itemize}

\FloatBarrier

\end{document}